\documentclass{article}

\usepackage{arxiv}

\usepackage[utf8]{inputenc} % allow utf-8 input
\usepackage[T1]{fontenc}    % use 8-bit T1 fonts
\usepackage{hyperref}       % hyperlinks
\usepackage{url}            % simple URL typesetting
\usepackage{booktabs}       % professional-quality tables
\usepackage{amsfonts}       % blackboard math symbols
\usepackage{nicefrac}       % compact symbols for 1/2, etc.
\usepackage{microtype}      % microtypography
\usepackage{lipsum}		% Can be removed after putting your text content
\usepackage{graphicx}
\usepackage{natbib}
\usepackage{doi}
\usepackage{amssymb,amsmath,amsthm}
\usepackage{color}
\usepackage[toc,page]{appendix}
\newtheorem{theorem}{Theorem}

\theoremstyle{definition}
\newtheorem{definition}{Definition}[section]

\title{Non-Negative Matrix Factorization with Scale Data Structure Preservation}

\author{ \href{https://orcid.org/0000-0002-9065-9591}{Rachid Hedjam}\\
	Department of Computer Science\\
	Sultan Qaboos University\\
	Muscat, Oman \\
	Synchromedia laboratory for telecommunication and telepresence\\
	École de technologie supérieure,Montreal,  Canada\\
	\texttt{rachid.hedjam@squ.edu.om} \\
	\And
	{Abdelhamid Abdesselam} \\
Department of Computer Science\\
Sultan Qaboos University\\
Muscat, Oman \\
	\texttt{ahamid@squ.edu.om} \\
	 \AND
	  Rahiche Abderrahmane, Mohamed Cheriet \\
	 Synchromedia laboratory for telecommunication and telepresence\\
	 École de technologie supérieure \\
	 Montreal \\
	 Canada\\
	 \texttt{abderrahmane.rahiche.1@ens.etsmtl.ca, mohamed.cheriet@etsmtl.ca} 
}

\renewcommand{\shorttitle}{\textit{arXiv} Template}

\hypersetup{
pdftitle={A template for the arxiv style},
pdfsubject={q-bio.NC, q-bio.QM},
pdfauthor={David S.~Hippocampus, Elias D.~Striatum},
pdfkeywords={First keyword, Second keyword, More},
}

\begin{document}
\maketitle

\begin{abstract}
{The model described} in this paper belongs to the family of non-negative matrix factorization methods designed for data representation and dimension reduction. In addition to preserving the data positivity property, it aims also to preserve the {structure of data} during matrix factorization. The idea is to add, to the NMF cost function, a penalty term to impose a scale relationship between the pairwise similarity matrices of the original and transformed data points. The solution of the new model involves deriving a new parametrized update scheme for the coefficient matrix, which makes it possible to improve the quality of reduced data when used for clustering and classification. The proposed clustering algorithm is compared to some existing NMF-based {algorithms} and to some manifold learning-based algorithms {when applied to} some real-life datasets. The {obtained results} show the effectiveness of the proposed algorithm.
\end{abstract}

% keywords can be removed
\keywords{NMF \and low-rank matrix factorization\and dimensionality reduction\and feature reduction.}

	\section{Introduction}
	\label{sec:intro}
	Low-rank matrix factorization (MF) is a hot topic in many research problems such as feature extraction and dimensionality reduction \cite{vidal2005generalized}, subspace segmentation \cite{liu2010robust}, data clustering \cite{favaro2011closed}, image processing and computer vision \cite{peng2012rasl} to mention a few. The key idea behind MF is that there is a latent data structure embedded in the high dimensional observed data which, once discovered, provides better capacity for learning. Formally,  MF techniques aim to decompose an observed high-dimensional data matrix into its constitute lower-dimensional factorizing matrices (in general two). One of the factorizing matrices represents the lower-dimensional space and the other one represents the spread of latent data in that space. MF has been widely used as a unified technique for dimensionality reduction, clustering, and matrix completion. There are several variants of MF in the literature including basic MF (BMF), {non}-negative MF (NMF) and Orthogonal NMF (ONMF). BMF are those described using traditional matrix decomposition such as principal component analysis (PCA), vector quantization (VQ) and singular value decomposition (SVD). While these techniques provide successful frameworks for simplifying data, they may produce factorizing matrices with negative values, making them difficult to interpret and explain {as many real problems are not negative}. Imposing a data positivity constraint on factorizing matrices is {therefore} more natural. To meet this requirement, NMF has been proposed \cite{lee2001algorithms}. Basically, NMF decomposes a non-negative data matrix $\textbf{X}$ into (usually) two lower rank no-negative matrices $\textbf{W}$ and $\textbf{H}$ (i.e., $\textbf{X}\approx \textbf{WH}$), where the columns of $\textbf{W}$ compose the new basis system, and the columns of $\textbf{H}$ are the coordinates of the transformed data {in the basis space}. Thus, $\textbf{W}$ is called the basis matrix and $\textbf{H}$ is called the coefficient matrix. NMF has the ability to decompose data into representative and meaningful non-negative parts, and this coincides with some psychological studies that prove that the human brain represents objects as parts-of-whole \cite{saavedra2013relative}. This very useful property makes NMF an outstanding technique to solve several real-world problems including classification and clustering \cite{tripodi2016context}, pattern representation and  recognition \cite{WeixiangPRL13}, feature extraction and dimensionality reduction \cite{wen2014local}, face recognition \cite{HaiqingPRL10}, bioinformatics \cite{stravzar2016orthogonal}, and document analysis \cite{shahnaz2006document}. Besides the advantage of discovering parts, NMF can find a learning rate larger than common gradient-based methods due to the multiplicative rule used to update the factorizing matrices \cite{lee1999learning, lee2001algorithms}. In addition to the data positivity constraint, orthogonality constraint can be also imposed on one of the factorizing matrices (or on both of them) to adapt NMF to data clustering problems. Orthogonality plays a key role in the development of another variant of NMF called Orthogonal NMF algorithms (ONMF). It has been shown that there is a relationship between ONMF and some other clustering algorithms such as k-means and spectral clustering \cite{ding2006orthogonal, ding2005spectral, li2013nonnegative}.
	Despite their success and popularity, traditional NMF techniques do not exploit the geometric structure of the data and are limited to working better with data assumed to be sampled from an Euclidean space. However, in many real cases, data is sampled from probability distribution that has support on or near to sub-manifold of the ambient space \cite{cai2010graph}. In order to detect the underlying lower dimensional manifold structure, many manifold learning methods have been published including ISOMAP \cite{tenenbaum2000global}, Locally Linear Embeding (LLE) \cite{roweis2000nonlinear}, Laplacian Eigenmaps \cite{belkin2002laplacian} and Neighborhood Preserving Embedding (NPE) \cite{he2005neighborhood}. The common assumption of these methods is that if two points are close to each other in the original space, the corresponding points in the new reduced space {should be} also close to each other. In other words, the objective of these methods is to preserve the proximity or closeness of the data in the reduced space. This is refereed to as  \textit{local invariance assumption} \cite{belkin2002laplacian, he2004locality}.
	Motivated by manifold learning, NMF has been extended to take into account data neighborhood structure and local invariance during calculating factorizing matrices \cite{cai2010graph, kuang2012symmetric, li2013structure, wei2014neighborhood,long2014graph, ahmed2021neighborhood}. \cite{cai2010graph} proposed  a method called Graph regularization NMF (GNMF) that constructs a nearest neighborhood graph on a scatter of data points to compute a pairwise Euclidean distance of the data in the reduced space and adds it as a penalty term to the NMF cost function. The penalty term, which involves the coefficient matrix, is added to preserve the pairwise closeness of the data points in the reduced space. \cite{kuang2012symmetric} proposed the symmetric NMF method (SymmNMF) which factorizes the similarity matrix of the observed data and calculates a lower rank symmetric  matrix intended for data clustering problems. \cite{li2013structure} proposed the SPNMF (structure preserving NMF) method in which three kinds of constraints are added to NMF, namely local affinity, distant repulsion, and embedding basis redundancy elimination. The goal of SPNMF is to solve the dimensionality reduction problem while preserving the local data affinity. SPNMF is more related to GNMF because both use the nearest neighbor graph to represent the geometric information of the data distribution. The difference between them is that SPNMF, unlike GNMF, imposes a constraint on distant data points from a repulsion perspective. \cite{wei2014neighborhood} proposed the NPNMF (neighborhood preserving NMF) method for dimensionality reduction problem on manifold. In order to preserve the local geometric structure of data, NPNMF imposes a constraint that each data point is represented as a linear combination of its neighbors. \cite{long2014graph} proposed the GDNMF (graph regularized discriminative NMF) method. This method adds two constraints to NMF, the intrinsic geometric data structure using the Laplacian graph and the discriminative information using supervised data labels. The two constraints are primarily used to learn a projection matrix that maps data from the original high dimensional space to a lower dimensional space. 
	The common characteristic of almost all the methods mentioned above, except SymmNMF, is that they share the same strategy of capturing the neighboring structure using Euclidean distance. Although they have been shown to be effective, they are sensitive to nearest neighbor and regularization parameters setting. As for the SymmNMF, the major disadvantage is that it ignores the feature information and only focuses on the pairwise similarity information, resulting in computational disadvantage. To overcome this limitation, \cite{ahmed2021neighborhood} proposed the Neighborhood Structure Assisted NMF (NS-NMF) method, which includes both feature information and geometric data structure information in a unified model designed specifically to address the anomaly detection problem. The new aspect of NS-NMF is that it imposes on the NMF a constraint on the geometric data structure modeled by a minimum spanning tree calculated from the neighborhood similarity graph. The minimum spanning tree favors sparsness in the computed similarity matrix, thus making the computation efficient.\\
	In this article, we propose a new NMF-based method more related to \cite{cai2010graph, kuang2012symmetric, ahmed2021neighborhood} with a difference in the way the pairwise similarity matrix is added to the NMF cost function. We call {this method} Data Structure Preservation NMF {(DSP-NMF)}. The previous methods integrate, in the penalty term, the similarity data matrix after having been calculated from the Euclidean distance between the nodes (data points) of the {corresponding} graph. {In our work, the similarity matrix is estimated} by $\textbf{X}^\top\textbf{X}$. The derivation of the proposed optimization model indeed leads to different updating rules of $\textbf{H}$ and $\textbf{W}$ compared to other methods.  
	Moreover, the proposed method preserves a scale relationship between $\textbf{X}^\top\textbf{X}$ and $\textbf{H}^\top\textbf{H}$. Therefore, it aims to factorize $\textbf{X}$ to $\textbf{WH}$ with the {penalty} $\textbf{X}^\top\textbf{X} \approx \lambda \textbf{H}^\top\textbf{H}$. The parameter $\lambda$ is a scalar that controls the scale between the two matrices. {This constraint imposes} that the pairwise relationship of the transformed data ($\textbf{H}^\top\textbf{H}$) is preserved as a scale form of the pairwise relationship of the original data ($\textbf{X}^\top\textbf{X}$).  {This property is useful in two respects, first, the local invariance is preserved with a scale after matrix factorization, and second, it implicitly involves the diagonalization of the pairwise similarity matrix composed of basis vectors (i.g., $\textbf{W}^\top \textbf{W} = \lambda \textbf{I}$, as described by Theorem 2. {$\textbf{I}$ stands for identity matrix}). Consequently, the calculated basis vectors are more decorrelated leading to less redundancy of the latent data discovered. Therefore, the orthogonality of the matrix \textbf{W} is implicitly imposed in our model compared to the Orthogonal NMF model \cite{ding2006orthogonal}, which explicitly imposes it as a constraint for data clustering purposes. From the optimization point of view the model derived by the Orthogonal NMF \cite{ding2006orthogonal} and that of the proposed model lead to totally different updating rules of the factorizing matrices. On the other hand, the proposed model is intended for preserving the local invariance property, similarly to GNMF \cite{cai2010graph}, Symmetric NMF \cite{kuang2012symmetric} and Neighborhood structured NMF (NS-NMF) \cite{ahmed2021neighborhood} and not necessary for clustering purposes like Orthogonal NMF \cite{ding2006orthogonal}.}
	The proposed method is inspired by the work \cite{HEDJAM2021107814}, in which a penalty term is added to NMF to preserve a scale relationship between the distribution of data and that of the computed clusters centroids. Although the objective function (Eq. 12) in \cite{HEDJAM2021107814} looks like the one in this paper (Eq. \ref{eq-sn-our}), there is a difference between them: i) in \cite{HEDJAM2021107814} the penalty term of Eq. (12) includes the basis matrix \textbf{W}, while in this work it includes the coefficient matrix \textbf{H}; ii) the optimization model in \cite{HEDJAM2021107814} implicitly implies the orthogonality of the matrix \textbf{H}, so the model is more suited to clustering problems; while in this work, it implicitly implies the diagonalization of the matrix \textbf{W}, so the model is more suited to dimension reduction and data representation in the same way as \cite{cai2010graph, kuang2012symmetric, ahmed2021neighborhood}; iii) the derivation of the two models leads to different updating rules for the factorization matrices \textbf{W} and \textbf{H}. To summarize, the contribution of this paper is twofold: 
	\begin{enumerate}
		\item Adding a new penalty term to the cost function of NMF, which {implicitly imposes the diagonalization on the similarity matrix composed from the basis vectors of $\textbf{W}$ (see Theorem 2)}.
		\item Incorporating a new parameter (i.e., $\lambda$) in the model to allow a scale relationship between the pairwise similarity matrices of the original and transformed data points.
	\end{enumerate}    
	The {remainder} of the paper consists of the following: Section \ref{seq-nmf} summarizes the NMF principle and describes some related NMF-based methods. Section \ref{sec-method} presents the proposed method. Experimental results are reported in Section \ref{seq-results}. Finally the discussion and conclusion are presented in Section
	
	\section{Some related works}
	\label{seq-nmf}
	\subsection{Basic NMF}
	{Formaly, let $\textbf{X}\in \mathbb{R}^{m\times n}$ be a matrix of $n$ columns representing the nonegative samples and $m$ rows representing their features,} and $r$ (lower rank) {is a positive integer} $<\{m,n\}$. NMF aims to find non-negative matrices $\textbf{W}\in\mathbb{R}^{m\times r}$ and $\textbf{H}\in\mathbb{R}^{r\times n}$ that {minimize} the following cost-function:
	
	\begin{equation}
	f(\textbf{W}, \textbf{H}) = \frac{1}{2}\|\textbf{X}-\textbf{W}\textbf{H}\|_F^2,
	\label{eq-nmf}
	\end{equation}
	where $\|_.\|_F^2 $ represents the Frobenius norm. The model in Eq. (\ref{eq-nmf}) can be also formulated as an optimization problem of the form:
	
	\begin{equation}
	\min_{\text{W},\text{H}>0}  \| \textbf{X}-\textbf{W}\textbf{H} \|_F^2 = \min_{\text{W},\text{H}>0}\sum_{i,j}(\textbf{X}-\textbf{W}\textbf{H} )_{ij}^2.
	\label{eq-optim1}
	\end{equation}
	
	Using multiplicative updates for non-negative optimization system proposed by Lee and Seung \cite{lee2001algorithms}, $\textbf{H}$ and $\textbf{W}$ are {updated} by:
	
	\begin{equation}
	\textbf{H}^{(t+1)}\leftarrow \textbf{H}^{(t)} \odot \frac{\textbf{W}^{(t)^\top}\textbf{X}}{\textbf{W}^{(t)^\top}\textbf{W}^{(t)}\textbf{H}^{(t)}},
	\label{eq-h0}
	\end{equation}
	
	\begin{equation}
	\textbf{W}^{(t+1)}\leftarrow \textbf{W}^{(t)} \odot \frac{\textbf{X}\textbf{H}^{(t+1)^\top}}{\textbf{W}^{(t)}\textbf{H}^{(t+1)}\textbf{H}^{(t+1)^\top}},
	\label{eq-i0}
	\end{equation}
	{where} $\odot$ stands for the element-wise matrix product, and $\frac{A}{B}$ stands for the element-wise matrix division. $\textbf{H}^{(0)}$ and $\textbf{W}^{(0)}$ are set to random values 
	and the updates are repeated until \textbf{W} and \textbf{H} become stable.
	
	\subsection{Graph regularized NMF (GNMF)}
	The GNMF method \cite{cai2010graph} adds, to the cost function of NMF, a penalty term based on a similarity graph to take into account the neighborhood structure between the data points. {A} parameter $\lambda$ is used to control the contribution of the neighborhood structure information. The resulting model is therefore suitable for clustering on a manifold. GNMF minimizes the following  objective function:
	
	\begin{equation}
	\min_{\text{W},\text{H}>0}\|\textbf{X}-\textbf{WH}\|^2_F + \lambda \text{Tr}(\textbf{HLH}^\top),
	\label{eq-gnmf}
	\end{equation}
	where $\textbf{L}$ is called the Laplacian matrix computed as $\textbf{L}=\textbf{D}-\textbf{W}$, and $\textbf{D}$ is a diagonal matrix with $\textbf{D}_{jj}=\sum_l\textbf{W}_{jl}$. The derivation of this optimization problem leads to the following updating rules:
	
	\begin{equation}
	\textbf{W}^{(t+1)}\leftarrow \textbf{W}^{(t)} \odot \frac{\textbf{X}\textbf{H}^{(t)^\top}}{\textbf{W}^{(t)}\textbf{H}^{(t)}\textbf{H}^{(t)^\top}},
	\label{eq-i}
	\end{equation}
	
	\begin{equation}
	\textbf{H}^{(t+1)}\leftarrow \textbf{H}^{(t)} \odot \frac{\textbf{X}^\top\textbf{W}^{(t+1)}+\lambda \textbf{W}^{(t+1)}\textbf{H}^{(t)^\top}}{\textbf{H}^{(t)^\top}\textbf{W}^{(t+1)^\top}\textbf{W}^{(t+1)}+\lambda \textbf{D} \textbf{H}^{(t)^\top}}.
	\label{eq-h}
	\end{equation}
	
	\subsection{Symmetric NMF (SymmNMF)}
	SymmNMF \cite{kuang2012symmetric} looks for the solution $\textbf{H}$ that minimizes the following objective function:
	
	\begin{equation}
	\min_{\text{H}>0} \|\textbf{A}-\textbf{H}^\top\textbf{H}\|_F^2,
	\end{equation}
	where $\textbf{A}\in \mathbb{R}^{n\times n}$ (with $\textbf{A}^\top = \textbf{A}$) is the pairwise similarity matrix, i.e., $(\textbf{A})_{ij}$ is the similarity value between the $i^{th}$ and $j^{th}$ node in the similarity graph. The multiplicative rule applied to SymmNMF is as follows:
	
	\begin{equation}
	\textbf{H}^{(t+1)}\leftarrow \textbf{H}^{(t)} \odot \frac{\textbf{H}^{(t)}\textbf{X}}{\textbf{H}^{(t)}\textbf{H}^{(t)^\top}\textbf{H}^{(t)}}.
	\label{eq-symmNMF}
	\end{equation}
	
	\subsection{Neighborhood structured NMF (NS-NMF)}
	NS-NMF \cite{ahmed2021neighborhood} aims to minimize the following objective function:
	
	\begin{equation}
	\min_{\text{W},\text{H}>0}\|\textbf{S}-\textbf{H}^\top\textbf{H}\|_F^2 + \alpha\|\textbf{X}-\textbf{WH}\|_F^2+\gamma(\|\textbf{W}\|_F^2+\|\textbf{H}\|_F^2),
	\label{eq-sn-nmf}
	\end{equation}
	
	where $\textbf{S}$ is the minimum spanning tree based neighborhood similarity matrix. The parameter $\alpha$ is used to control the balance between the two cost terms, while the terms $\|\textbf{W}\|_F^2$ and $\|\textbf{H}\|_F^2$ are used for regularizing the objective function against over-fitting, and $\gamma$ is the regularization parameter controlling te extent of over-fitting. The updating rules of the factorizing matrices are given by the gradient descent technique as shown in Eqs. (14, 15) of the reference \cite{kuang2012symmetric}.

	\section{Proposed method}
	\label{sec-method}
	
\begin{figure}[!hb]
		\centering
		\begin{tabular}{ccc}
			\includegraphics[scale=0.35]{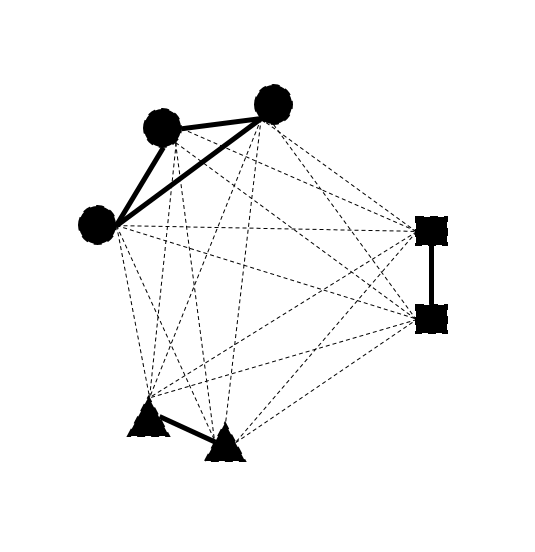}&
			\includegraphics[scale=0.35]{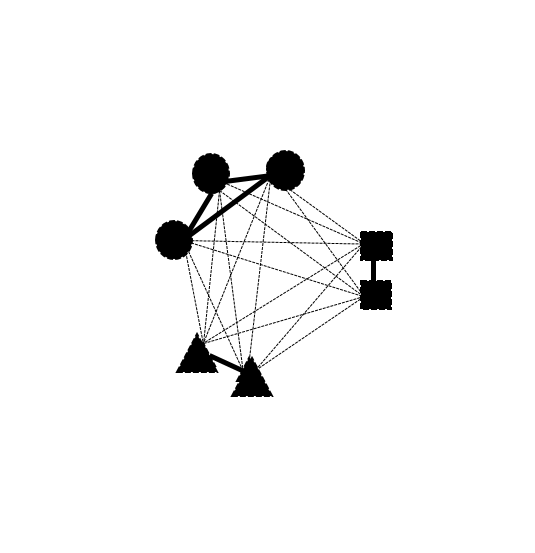}&
			\includegraphics[scale=0.35]{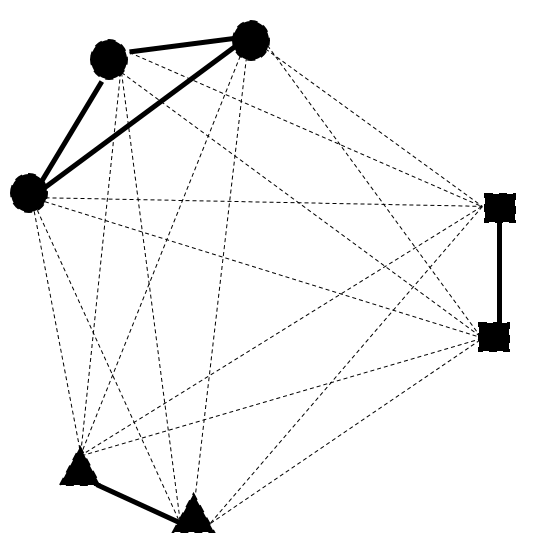}\\
			(a) & (b) & (c)
		\end{tabular}
		\caption{The structure relationship between original data and transformed data. (a) input data structure, (b, c) transformed data structure, with $\lambda > 1$ and $\lambda < 1$ respectively. Strong similarities between data are represent by thick edges, while weak similarity are represented by dotted edges.}
		\label{fig-strcturepreserve}
	\end{figure}

	Let's {note} that $\textbf{X}^\top\textbf{X}$ {represents} the pairwise similarity data matrix. It is in fact a dot-product weighting matrix, i.e., $(\textbf{X}^\top\textbf{X})_{ij}=\textbf{x}_i^\top \textbf{x}_j$, where $\textbf{x}_i$ and $\textbf{x}_j$ are the vectors of the $i^{th}$ and $j^{th}$ data points respectively. If {the columns of $\textbf{X}$ are unit vectors}, the dot {product is} equivalent to the cosine similarity of the two vectors {$\textbf{x}_i$ and $\textbf{x}_j$}. Similarly, { $\textbf{H}^\top\textbf{H}$ represents} the pairwise similarity of the transformed data matrix (coefficient matrix). As mentioned in the introduction, we aim to add, to the cost function of NMF, a penalty term that enforces the factorization model to preserve a scale relationship between the structure of the input data points (described by $\textbf{X}^\top\textbf{X}$) and that of the transformed data points (described by $\textbf{H}^\top\textbf{H}$). More formally, we want to impose $\textbf{X}^\top\textbf{X} \approx \lambda \textbf{H}^\top\textbf{H}$ during calculating $\textbf{W}$ and $\textbf{H}$. The parameter $\lambda$ is a strictly positive scalar that controls the scale between the two data similarity matrices. {In fact, the scale $\lambda$} is used to fine-tune the {relationship and give} the freedom to the algorithm to find the best match between $\textbf{X}^\top\textbf{X}$ and $\lambda\textbf{H}^\top\textbf{H}$. Actually, since the values and dimension of $\textbf{X}$ and $\textbf{H}$ are different, the match between $\textbf{X}^\top\textbf{X}$ and $\textbf{H}^\top\textbf{H}$ is not guaranteed. This is why $ \lambda $ is used to adjust the correspondence between the two terms.  {This can} be illustrated in Figure \ref{fig-strcturepreserve}. Besides, {the proposed penalty term} is useful because it implies the diagonalization of the pairwise similarity matrix of the basis vectors (i.e.,  $\textbf{W}^\top\textbf{W}=\lambda \textbf{I}$) as indicated in Theorem 2 below. The diagonalization of  $\textbf{W}^\top\textbf{W}$ means that {the correlation between the new basis vectors is very low}, which guarantees a good separation of the data when they are projected on them.
	
	Therefore, the proposed objective function to minimize is as follows:
	
	\begin{equation}
	\mathcal{O} = \|\textbf{X}-\textbf{WH}\|_F^2+ \|\textbf{X}^\top \textbf{X}-\lambda \textbf{H}^\top\textbf{H}\|_F^2,
	\label{eq-sn-our}
	\end{equation}
	which can be rewritten as:
	
	\begin{align}
	\mathcal{O} &=Tr\Big [\big (\textbf{X}-\textbf{WH}\big)^\top \big(\textbf{X}-\textbf{WH}\big)\Big] \notag\\
	& +Tr\Big[\big(\textbf{X}^\top \textbf{X}-\lambda \textbf{H}^\top \textbf{H}\big)^\top \big(\textbf{X}^\top \textbf{X}-\lambda \textbf{H}^\top \textbf{H} \big)\Big] \notag \\
	&=Tr(\textbf{X}^\top \textbf{X})-2Tr(\textbf{X}^\top \textbf{WH})+Tr(\textbf{H}^\top \textbf{W}^\top \textbf{WH}) +\notag\\
	&\quad~Tr(\textbf{X}^\top \textbf{X X}^\top \textbf{X})-2\lambda Tr(\textbf{X}^\top X \textbf{H}^\top \textbf{H})+\lambda^2Tr(\textbf{H}^\top \textbf{HH}^\top \textbf{H}). 
	\end{align}
	
	By introducing two variable matrices $\Lambda_w$ and $\Lambda_h$ (of the same dimension as {$\textbf{W}^\top$ and $\textbf{H}^\top$} respectively) used as the Lagrangian multiplier {to} constrain $\textbf{W}\geq 0$ and $\textbf{H}\geq 0$ respectively, the model above can be transformed in the following Lagrangian form:
	\begin{align}
	J &= Tr(\textbf{X}^\top \textbf{X})-2Tr(\textbf{X}^\top \textbf{WH})+Tr(\textbf{H}^\top \textbf{W}^\top \textbf{WH}) +\notag\\
	&\quad~Tr(\textbf{X}^\top \textbf{X X}^\top \textbf{X})-2\lambda Tr(\textbf{X}^\top X \textbf{H}^\top \textbf{H})+\lambda^2Tr(\textbf{H}^\top \textbf{HH}^\top \textbf{H})- \notag\\
	&\quad~ \Big [Tr(\Lambda_w \textbf{W}^\top)+Tr(\Lambda_h \textbf{H}^\top)\Big ],
	\label{eq-lagrage}
	\end{align}
	Therefore, the optimization of the model (\ref{eq-lagrage}) leads to derive the following multiplicative update rules:
	
	\begin{equation}
	\textbf{H}^{(t+1)} = \textbf{H}^{(t)}\odot\frac{\textbf{W}^{(t)^\top} \textbf{X}+2\lambda \textbf{H}^{(t)}\textbf{X}^\top \textbf{X} }{\textbf{W}^{(t)\top} \textbf{W}^{(t)}\textbf{H}^{(t)} + \lambda^2\textbf{H}^{(t)}\textbf{H}^{(t)\top} \textbf{H}^{(t)}   },
	\label{eq-ourh}
	\end{equation}
	
	\begin{equation}
	\textbf{W}^{(t+1)}= \textbf{W}^{(t)}\odot\frac{ \textbf{XH}^{(t+1)^\top} }{ \textbf{W}^{(t)}\textbf{H}^{(t+1)} \textbf{H}^{(t+1)\top}   }. 
	\label{eq-ourw}
	\end{equation}
	
	For the derivation of Eqs. (\ref{eq-ourh}) and (\ref{eq-ourw}), the reader is referred to Appendix A. With regard to the updating rules, the following theorem can be given:
	
	\begin{theorem}
		The objective function $\mathcal{O}$ in Eq. (\ref{eq-sn-our}) is non-increasing under the updating rules in Eqs. (\ref{eq-ourh}) and (\ref{eq-ourw}). 
	\end{theorem}
	To prove the the theorem above, we followed approximately the same procedure as in \cite{cai2010graph}. It is described in Appendix B.\\
	
	For more optimization stability, in practice some researchers require that the Euclidean length of the column vectors in matrix $\textbf{W}$ is 1 \cite{shahnaz2006document,cai2010graph}. the matrix \textbf{H} will be adjusted accordingly so that \textbf{WH} does not change. this can be achieved by:
	
	\begin{equation}
	w_{ir} \leftarrow \frac{w_{ir}}{\sqrt{\sum_iw^2_{ir}}}, \quad h_{rj}\leftarrow h_{rj}\sqrt{\sum_iw^2_{ir}}
	\label{eq-norm}
	\end{equation}
	
	Similarly to \cite{cai2010graph}, our algorithm also adopts the following normalization strategy: Once the multiplicative updating process converges, we normalize \textbf{W} and \textbf{H} by applying Eq. (\ref{eq-norm}). Moreover, to balance the effect of various features, {we apply min-max normalization of the data matrix \textbf{X}} before starting the factorization process. It's worth {noting} that we {did not used} a balancing parameter (factor) between the first and second terms of Eq. (\ref{eq-sn-our}). {Because}, by adding a balancing parameter (e.g., $f$ ) to the second term in Eq. (\ref{eq-sn-our}) {leads} to a factor multiplied by $\lambda$ (ie., $f.\lambda$) in the numerator and denominator of Eq. (\ref{eq-ourh}). Therefore, we think that $\lambda$ can be used also as parameter to control somehow the relative importance of the two terms.  
	
	\begin{theorem}
		Let $\textbf{X}\in \mathbb{R}^{m\times n}$, $\textbf{W}\in \mathbb{R}^{m\times r}$ and $\textbf{H}\in \mathbb{R}^{r\times n}$, such that $\textbf{X}=\textbf{WH}$. If we assume that $\textbf{W}$ has a left inverse, then the two terms $\textbf{X}^\top\textbf{X} \approx \lambda \textbf{H}^\top\textbf{H}$ and  $\textbf{W}^\top\textbf{W} \approx \lambda\textbf{I}_r$ are equivalent ($\lambda$ is a positive scalar).
		\label{theorem2}
	\end{theorem}
	The proof of the Theorem \ref{theorem2} is given in the Appendix C. 
	\section{Experimental results and evaluation}
	\label{seq-results}
	
	Two (2) experiments are carried out to evaluate the {performance of the} proposed algorithm (DPS-NMF). In the first experiment, we compared and evaluated {its performance against} three (3) NMF-based algorithms {sharing the same positivity property}: GNMF (Graph regularized NMF) \cite{cai2010graph}, SymmNMF (Symmetric NMF) \cite{kuang2012symmetric} and SDNMF (Sinkhorn Distance NMF) \cite{qian2016non}. These algorithms can be considered {as belonging to} the same family because they share the same property of preserving data positivity. In the second experiment, the algorithms aforementioned, including ours, were compared to three (3) other algorithms that share the same property of preserving neighborhood structure, namely UMAP (Uniform Manifold Approximation and Projection) \cite{mcinnes2018umapsoftware}, LLE (Locally linear embedding) \cite{roweis2000nonlinear} and SE (Spectral Embedding) \cite{belkin2002laplacian}. All algorithms are evaluated on six (6) real world datasets from different applications of Machine learning. In order to assess the efficiency of each algorithm, the reduced data they generated are used for clustering (K-means) and classification (SVM and KNN). Since NMF is not deterministic, the results may differ from one run to the other. Thus, each NMF-based is executed five (5) times and its average (\textbf{\textit{mean}}) and {maximum} (\textbf{\textit{max}}) performance are reported (for fairness, other algorithms are applied in the same way as well). K-means clustering performance is assessed using \textbf{\textit{NMI}} (Normalized Mutual Information) \cite{vinh2010information} while SVM and KNN classification performance is assessed using the accuracy score (\textbf{\textit{Acc}}).  
	
	\subsection{Dataset description}
	\label{sq-datasets}
	Six (6) real-life datasets are used to evaluated the performance of the six (6) algorithms mentioned above. Two of them are downloaded from the UC Irvine Machine Learning Repository\footnote[1]{\url{https://archive.ics.uci.edu/ml/index.php}}, namely: Breast Cancer Wisconsin (Diagnostic) dataset, and  Wine dataset. The third dataset is the Optical Recognition of Handwritten Digits dataset, which can be found in the {SciKit-learn} library\footnote[2]{\scriptsize\url{https://scikit-learn.org/stable/modules/generated/sklearn.datasets.load_digits.html}}. The fourth an fifth dataset are the Indian Pines (IP) and Pavia University scene which can be downloaded from this link\footnote[3]{\scriptsize \url{http://www.ehu.eus/ccwintco/index.php/Hyperspectral\_Remote\_Sensing\_Scenes}}. They are remote sensed hyper-spectral images. IP has 200 bands after removing 24 water bands. It consists of sixteen classes, seven of which are removed because they contain too few samples (less than 390 pixels) compared to the largest ones. Twenty percent ($20\%$) of the pixels are randomly sampled from each class. Pavia has 103 bands and consists of nine classes. Twenty percent ($20\%$) of the pixels are randomly sampled from each class. The sixth dataset is the Columbia Object Image Library (COIL20)\footnote[4]{\url{https://www.cs.columbia.edu/CAVE/software/softlib/coil-20.php}}. It consists of 1440 128$\times$128 grayscale images (20 objects with 72 poses each). In our experiment, the images are resized to 32$\times$32 pixels lading to 1024 features each.
	\subsection{Parameter setting}
	\label{seq-param}
	
\begin{figure}[!htp]
		\centering
		\begin{tabular}{ccc}
			\includegraphics[scale=0.225]{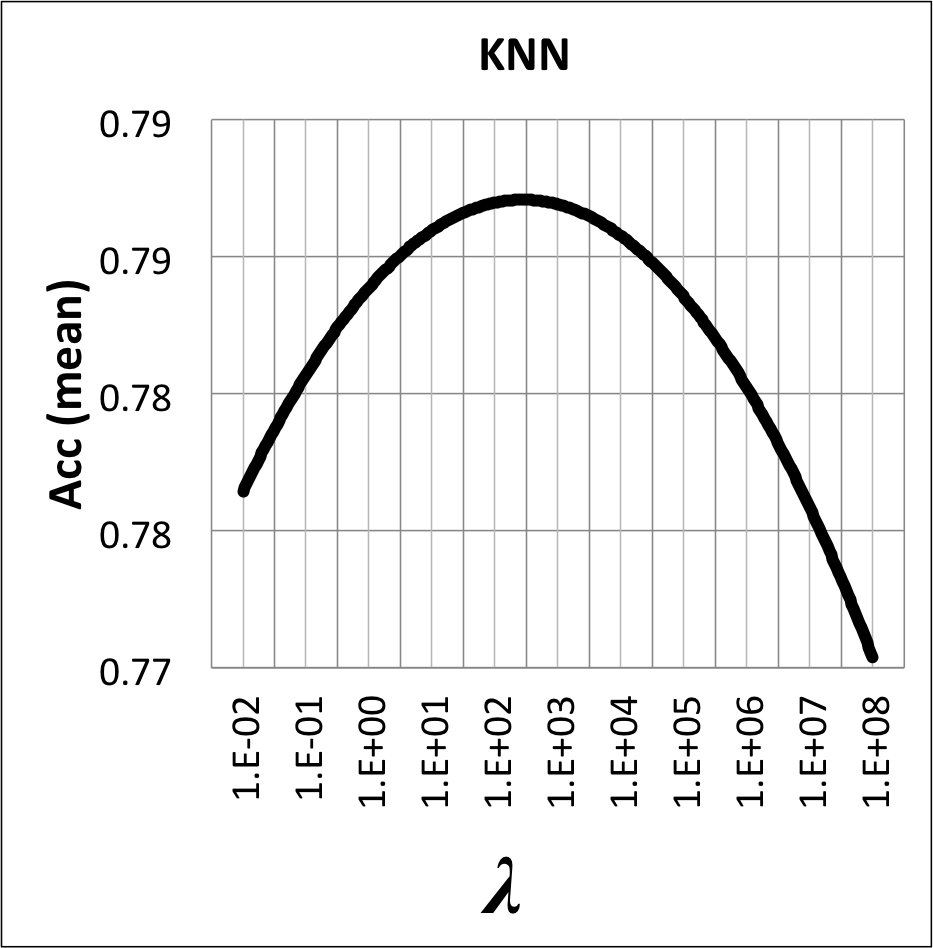}&
			\includegraphics[scale=0.225]{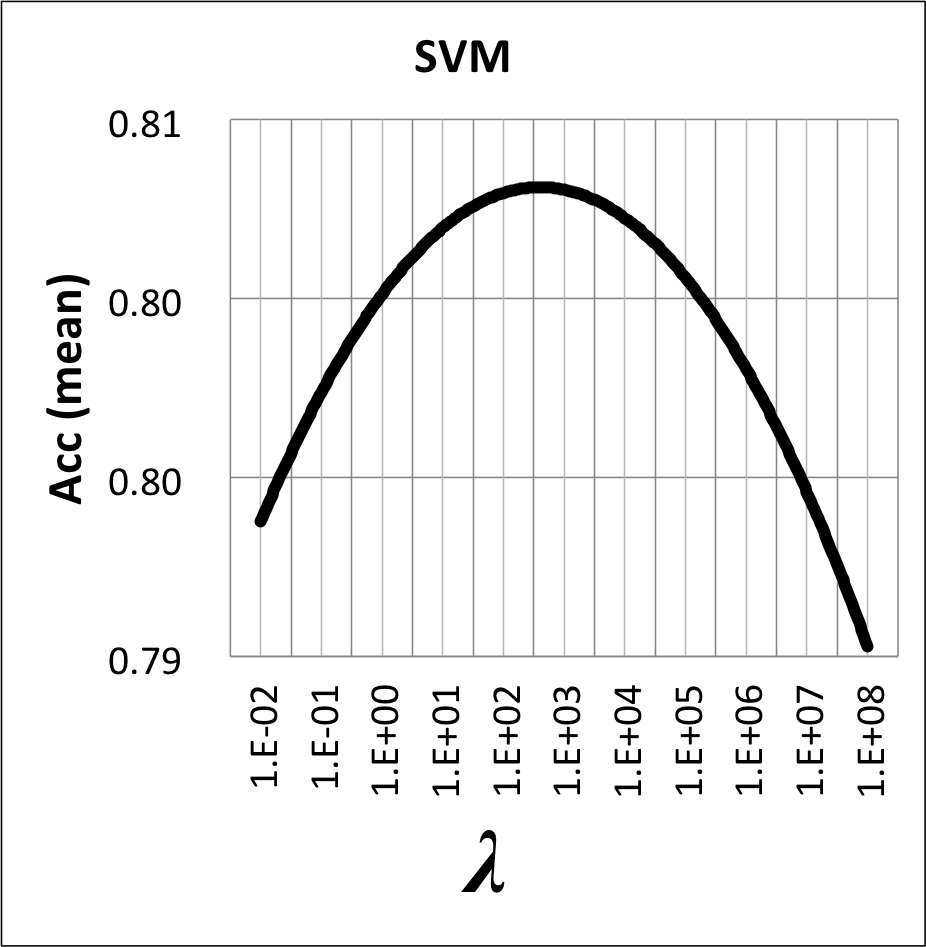}&
			\includegraphics[scale=0.225]{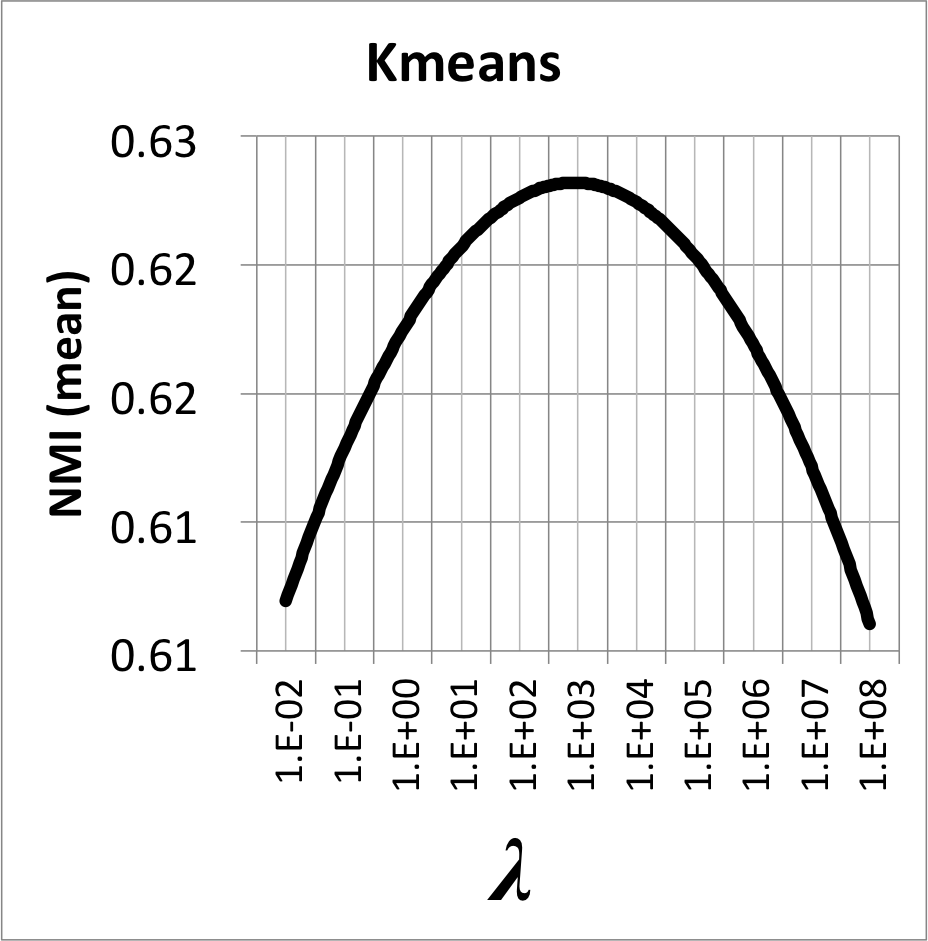}
		\end{tabular}
		\caption{Best $\lambda$ parameter estimation. Example for rank $r=3$.}
		\label{fig-bestL}
	\end{figure}

	There is one free-parameter to set, which is $\lambda $ used to update the coefficient matrix $ \textbf{H} $ in Eq. (\ref{eq-ourh}). In this work, the parameter setting is performed experimentally. To find the best parameter setting for all experiments, we ran the proposed algorithm with different values of $ \lambda $ on all the datasets described in Section \ref{sq-datasets}, then {we} used the corresponding {low-ranked} datasets for clustering (K-means) and also for classification (SVM and K-NN). K-means, SVM and KNN are trained and tested using 5-fold cross-validation process. The clustering performance is assessed based on the \textbf{\textit{NMI}} metric, and the classification performance is assessed based on the accuracy score (\textbf{\textit{Acc}}). The Python implementation of clustering, classification and performance algorithms can be found in the SciKit Python library \cite{scikit-learn}.  We have used the following values of $\lambda$: $[10^{-2},10^{-1},1,10,10^{2},10^{3},10^{4},10^{5},10^{6},10^{7},,10^{8} ]$. We computed the average \textbf{\textit{NMI}} and \textbf{\textit{Acc}} over all the datasets. The evolution of the algorithm's performance versus $ \lambda $ is indicated by the trend lines shown in Figure \ref{fig-bestL} (example with rank $ r = 3 $). Experiments carried out have shown that clustering and classification can achieve their best performance when $\lambda$ is around $10^{2}$ and $10^{3}$ at least for the datasets considered. The other parameters not directly linked to the proposed model are: i) the value $ k $ of the KNN algorithm, which is set to 3; and ii) the $ kernel $ and $ C $ of the SVM classifier, which are respectively set to $ rbf $ and $ 1000 $. 
	
	\subsection{Experimental results}
\begin{figure}[!htp]
		\centering
		\begin{tabular}{ccc}
			\includegraphics[scale=0.165]{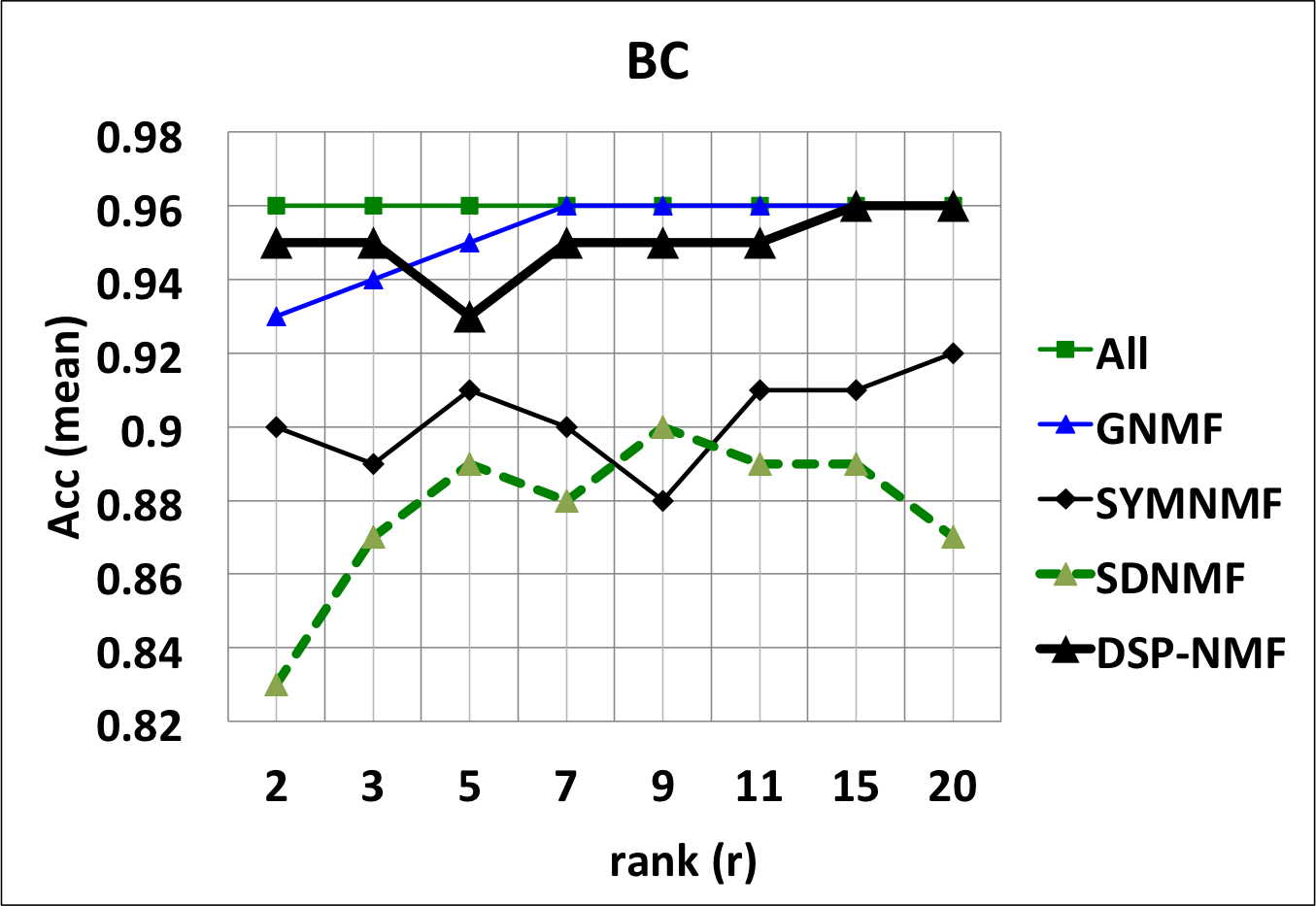}&\hspace{-3ex}
			\includegraphics[scale=0.165]{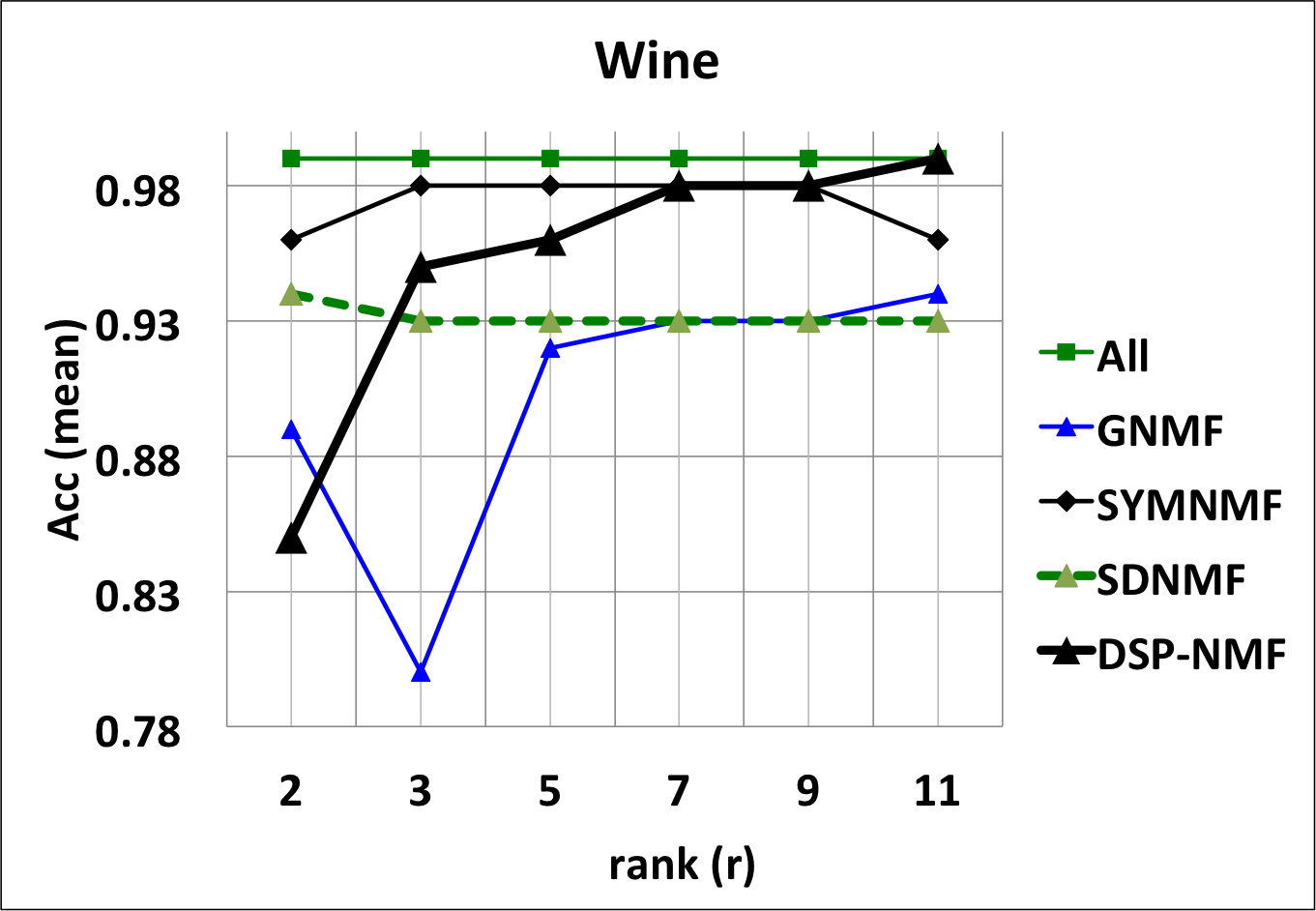}&\hspace{-3ex}
			\includegraphics[scale=0.165]{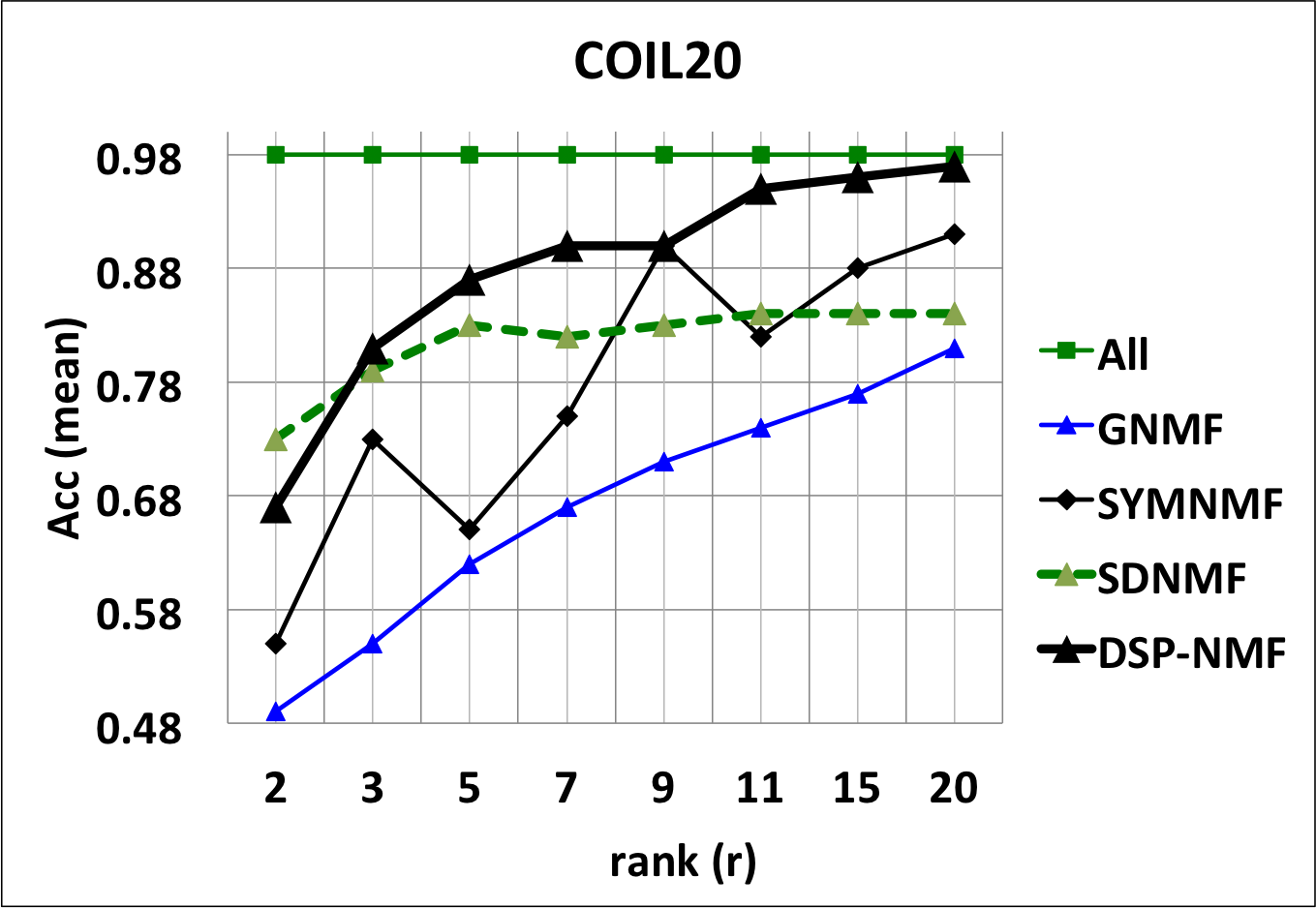}\\
			\includegraphics[scale=0.165]{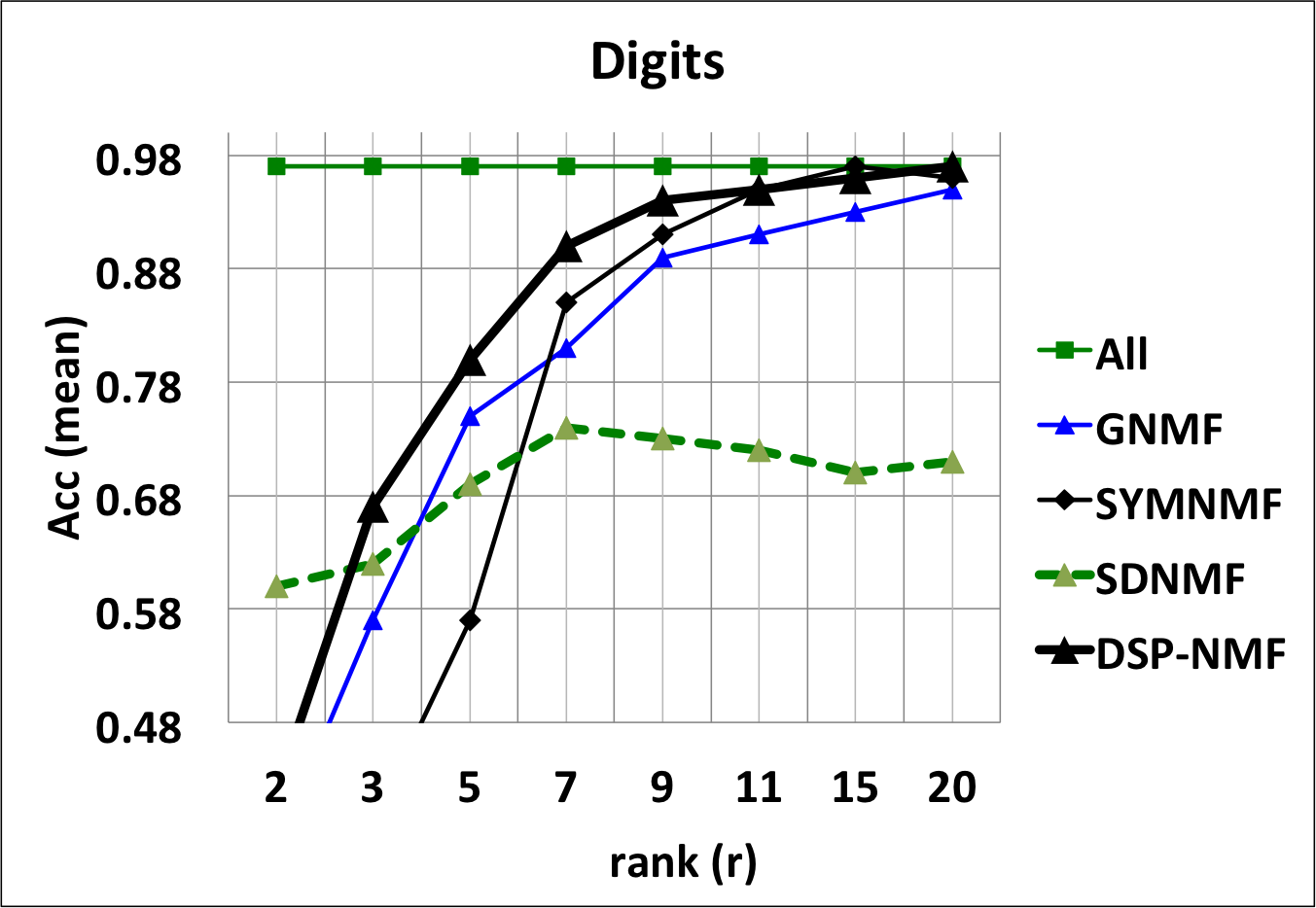}&\hspace{-3ex}
			\includegraphics[scale=0.165]{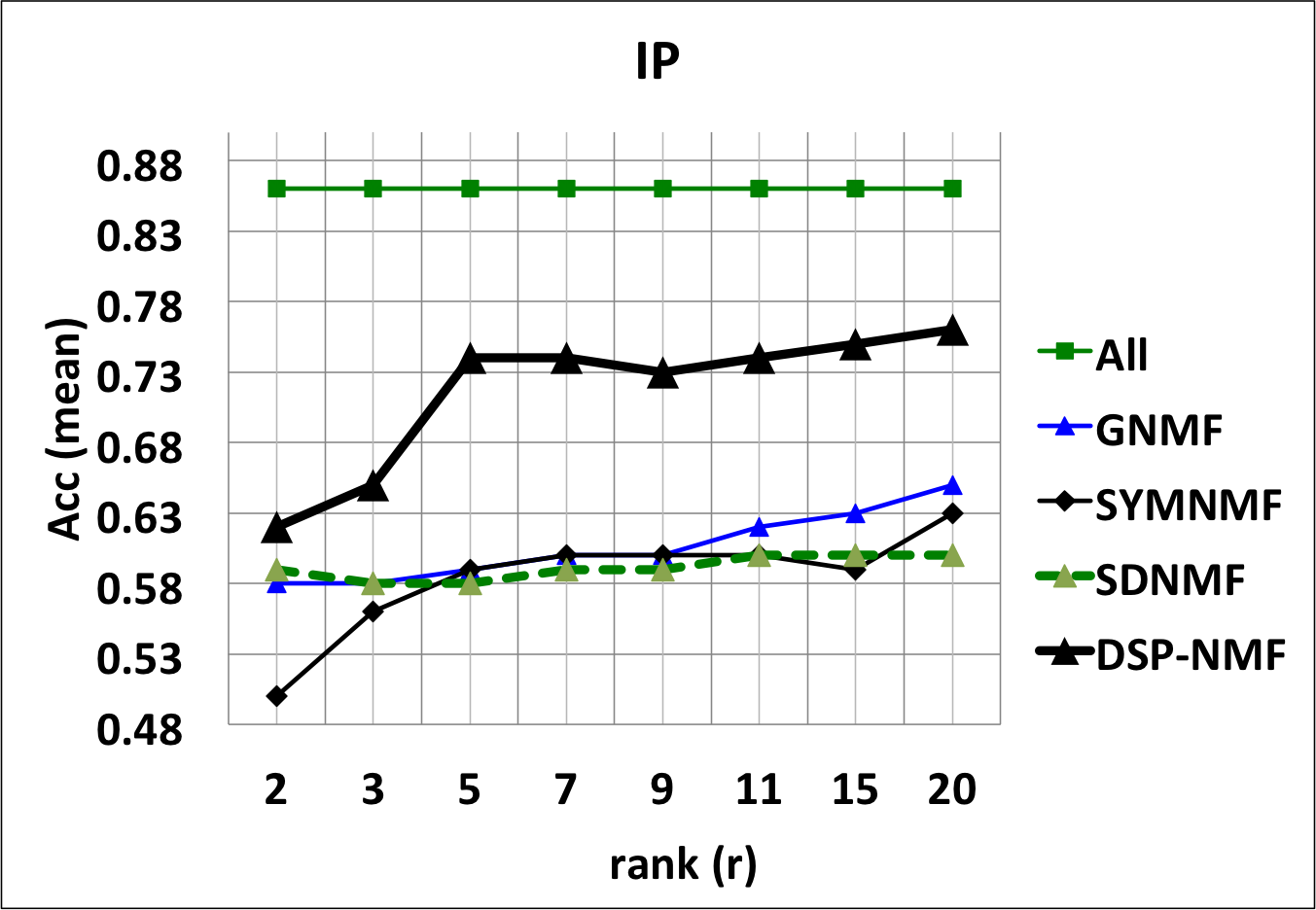}&\hspace{-3ex}
			\includegraphics[scale=0.165]{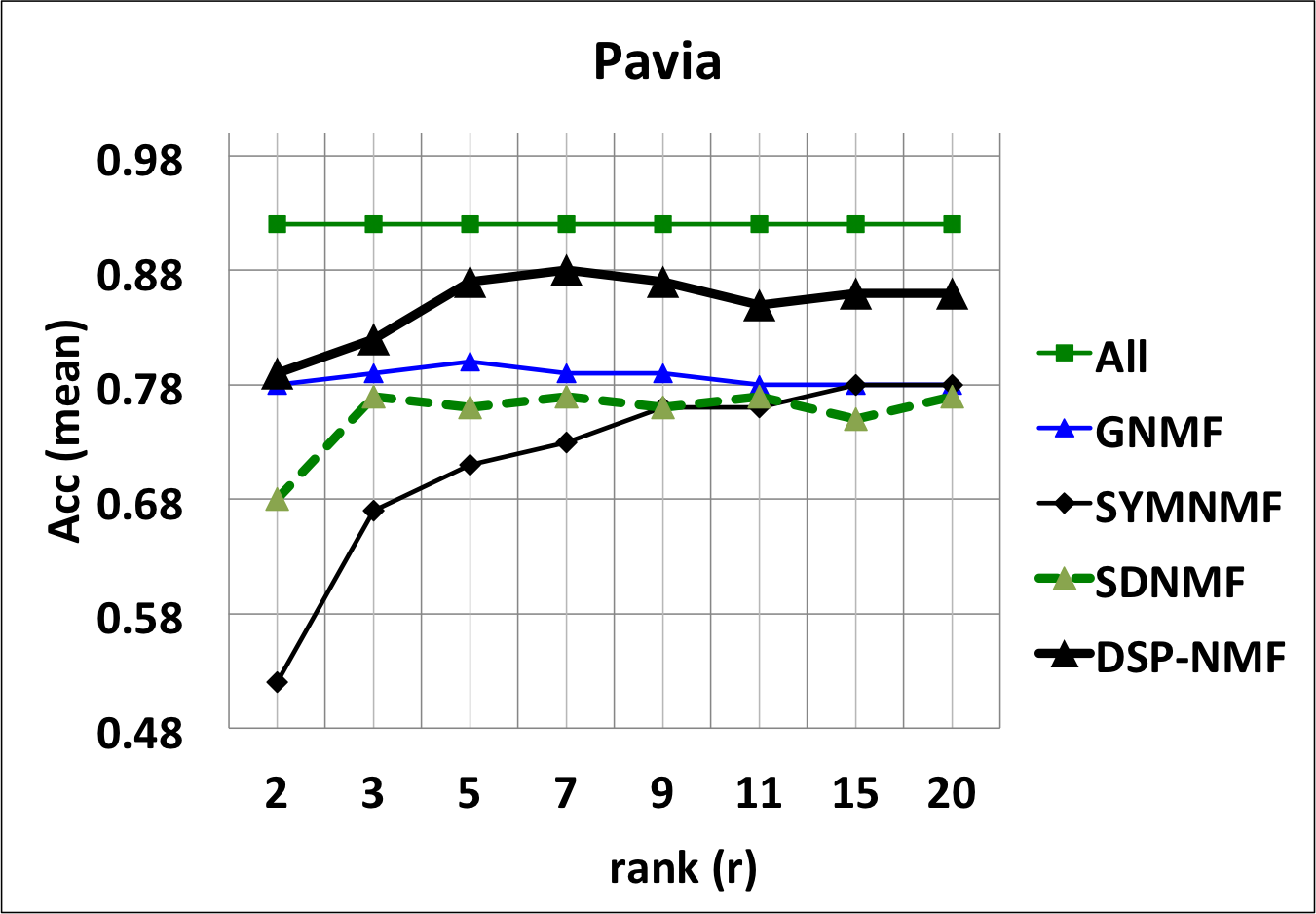}
		\end{tabular}
		\caption{Average SVM accuracy (Acc) of different NMF-based dimension reduction algorithms.}
		\label{fig-resultssvm-2}
	\end{figure}

\begin{figure}[!htp]
		\centering
		\begin{tabular}{ccc}
			\includegraphics[scale=0.165]{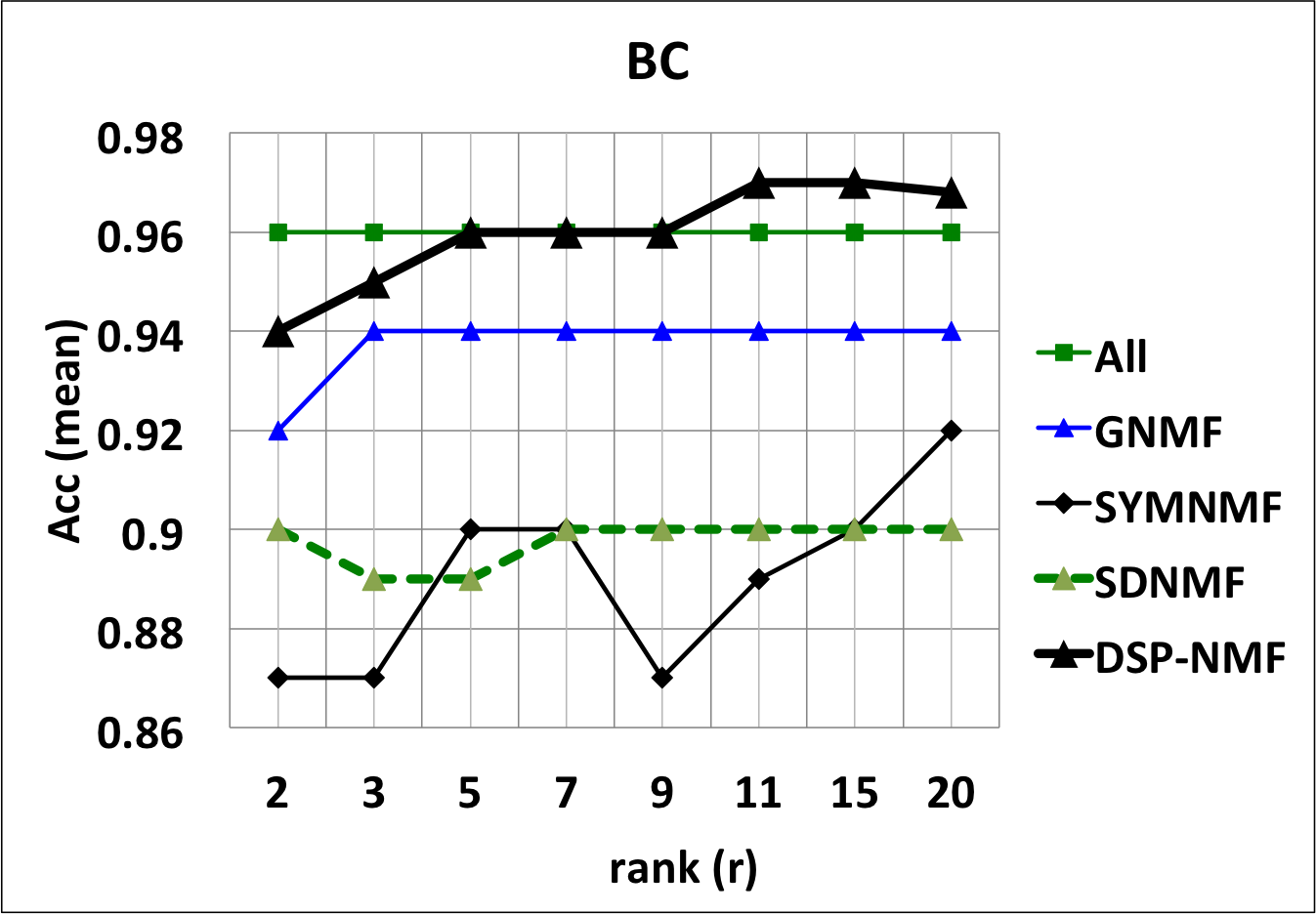}&\hspace{-3ex}
			\includegraphics[scale=0.165]{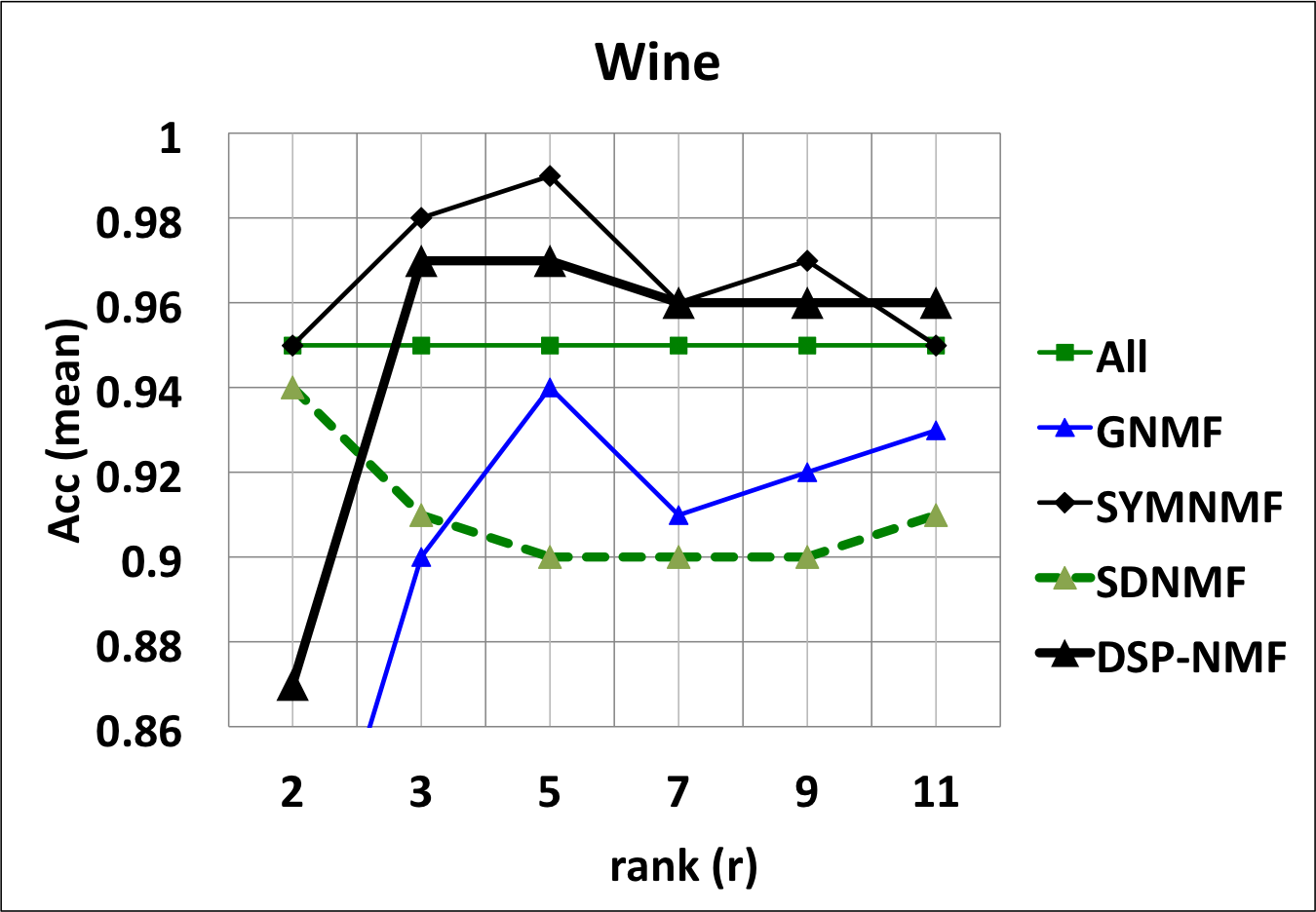}&\hspace{-3ex}
			\includegraphics[scale=0.165]{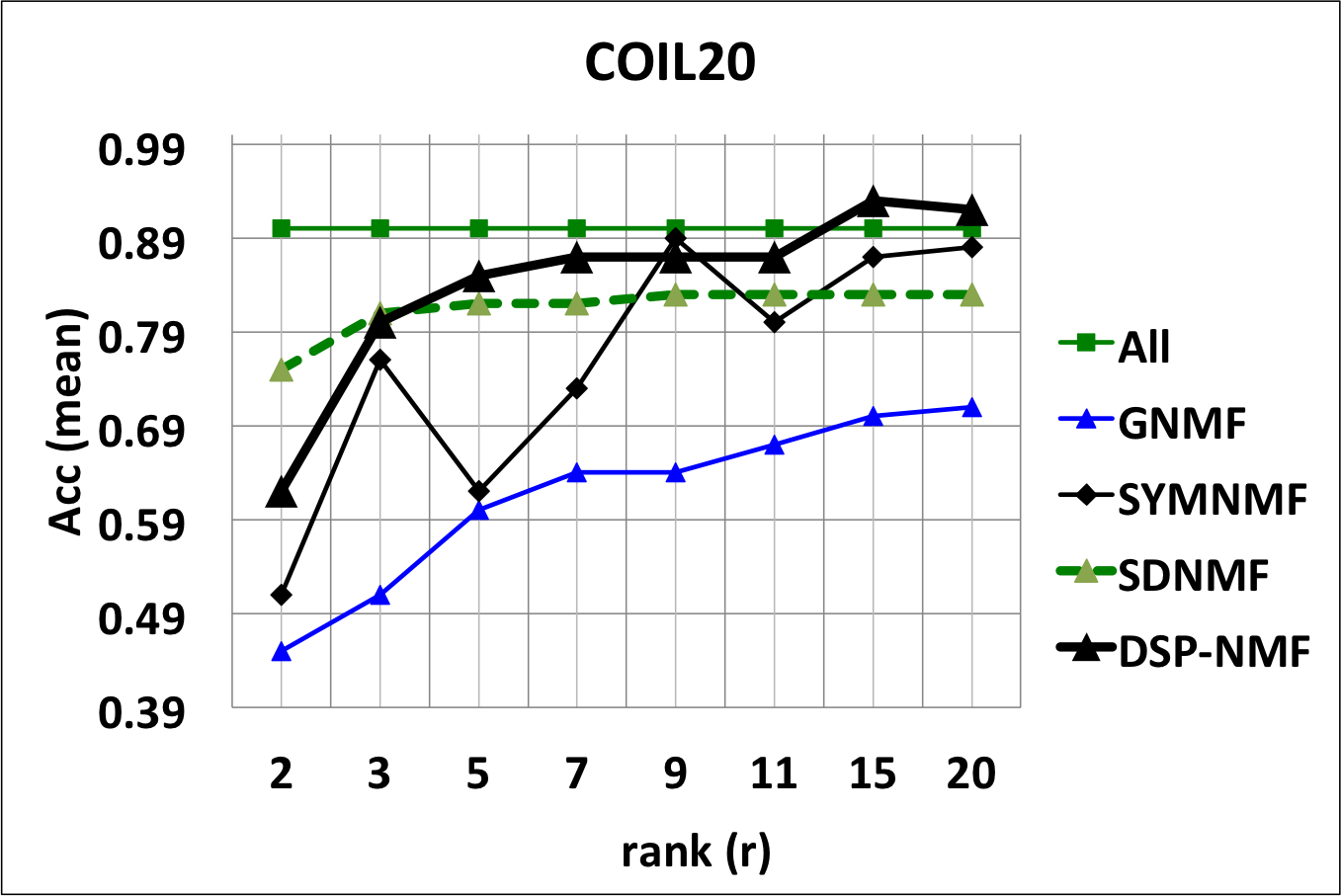}\\
			\includegraphics[scale=0.165]{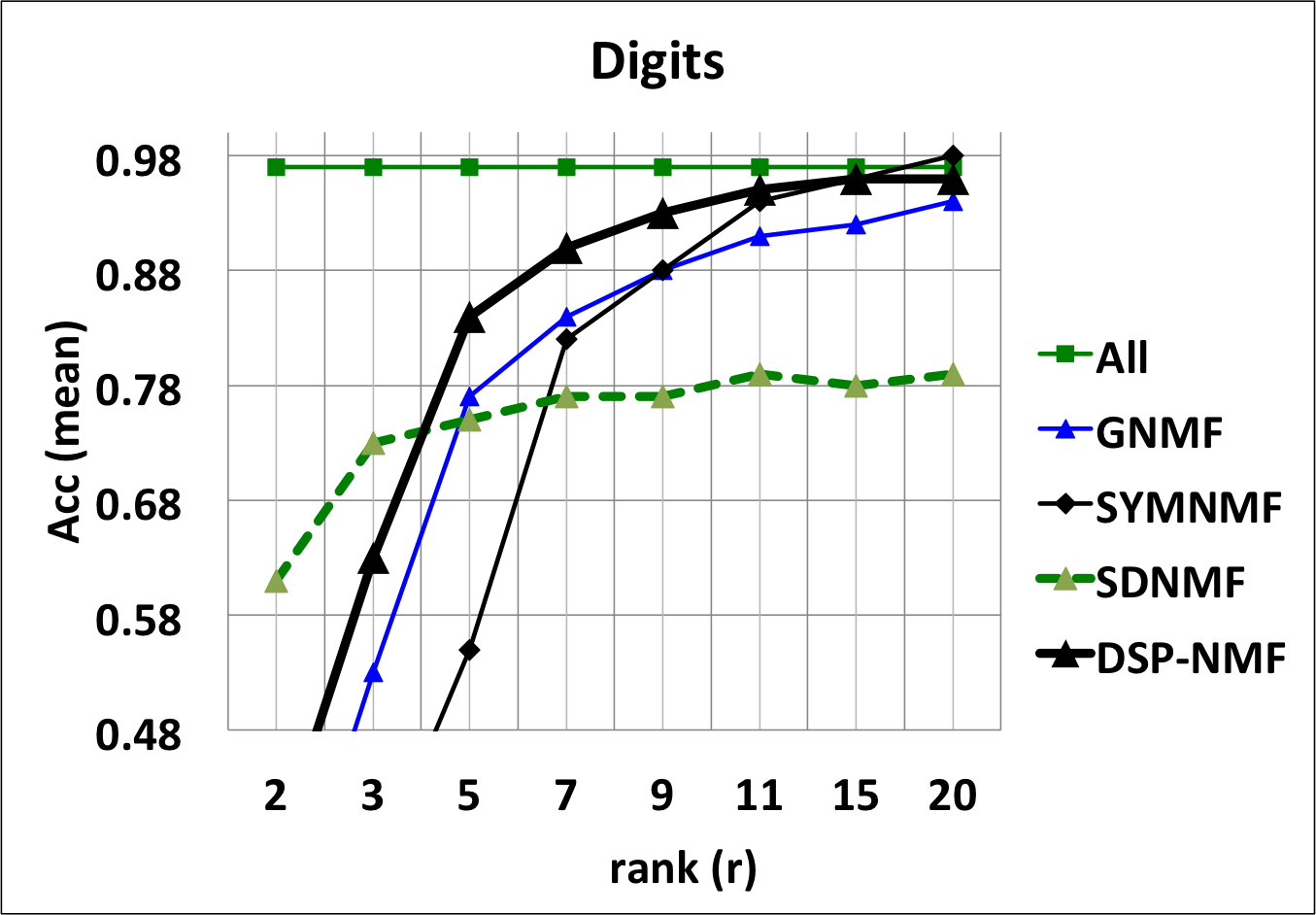}&\hspace{-3ex}
			\includegraphics[scale=0.165]{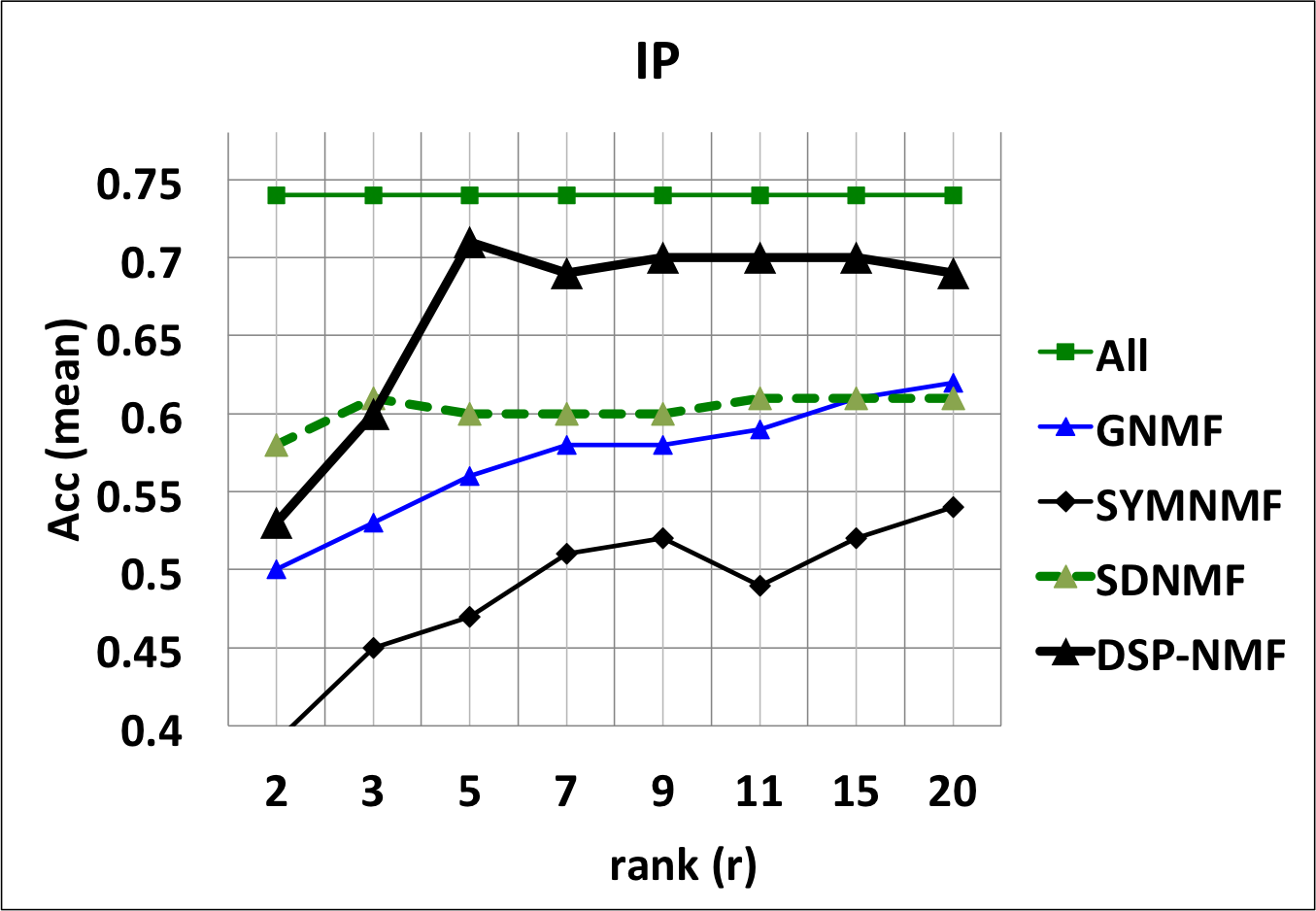}&\hspace{-3ex}
			\includegraphics[scale=0.165]{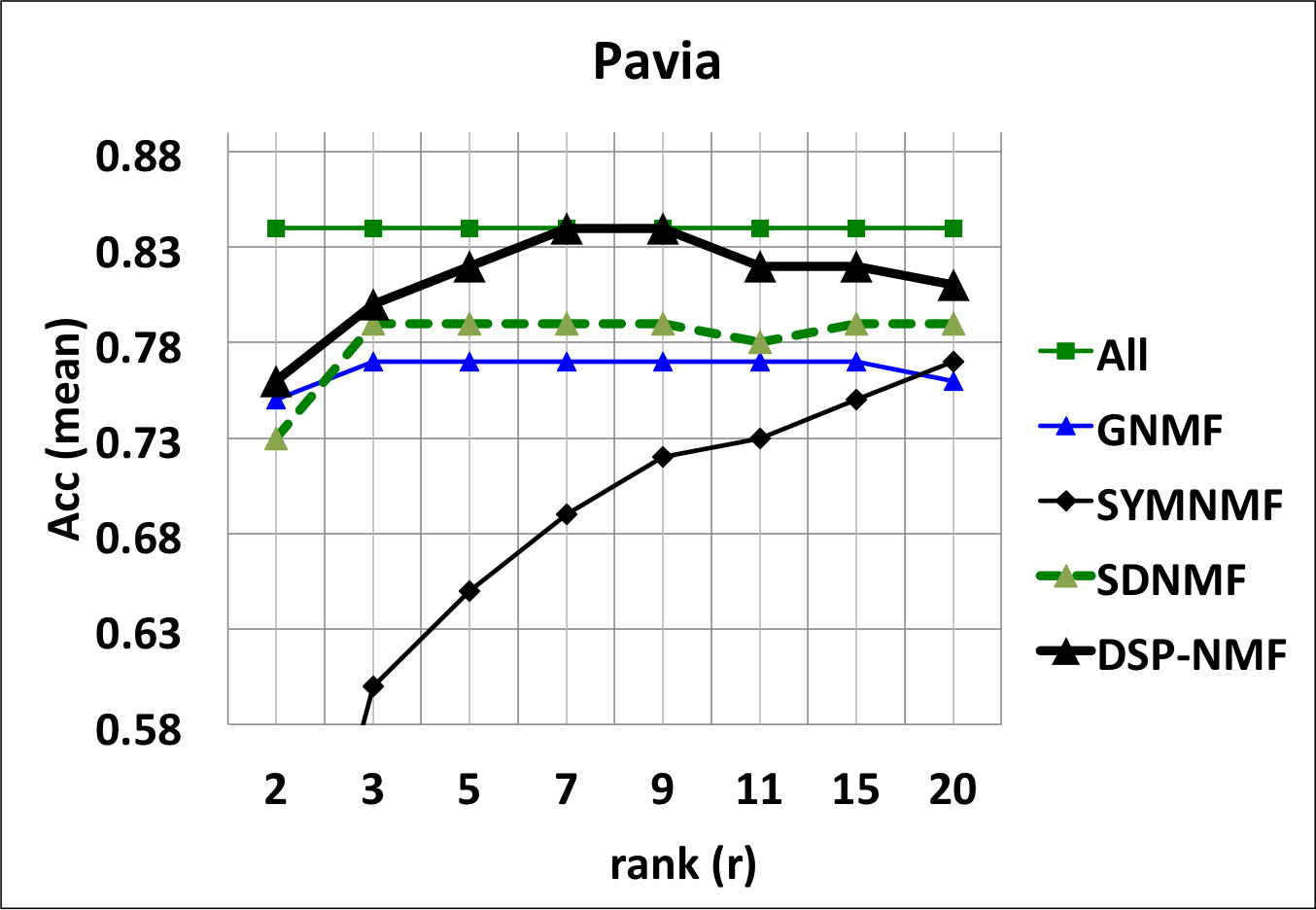}
		\end{tabular}
		\caption{Average KNN accuracy (Acc) of different NMF-based dimension reduction algorithms.}
		\label{fig-resultsknn-2}
	\end{figure}

\begin{figure}[!htp]
		\centering
		\begin{tabular}{ccc}
			\includegraphics[scale=0.165]{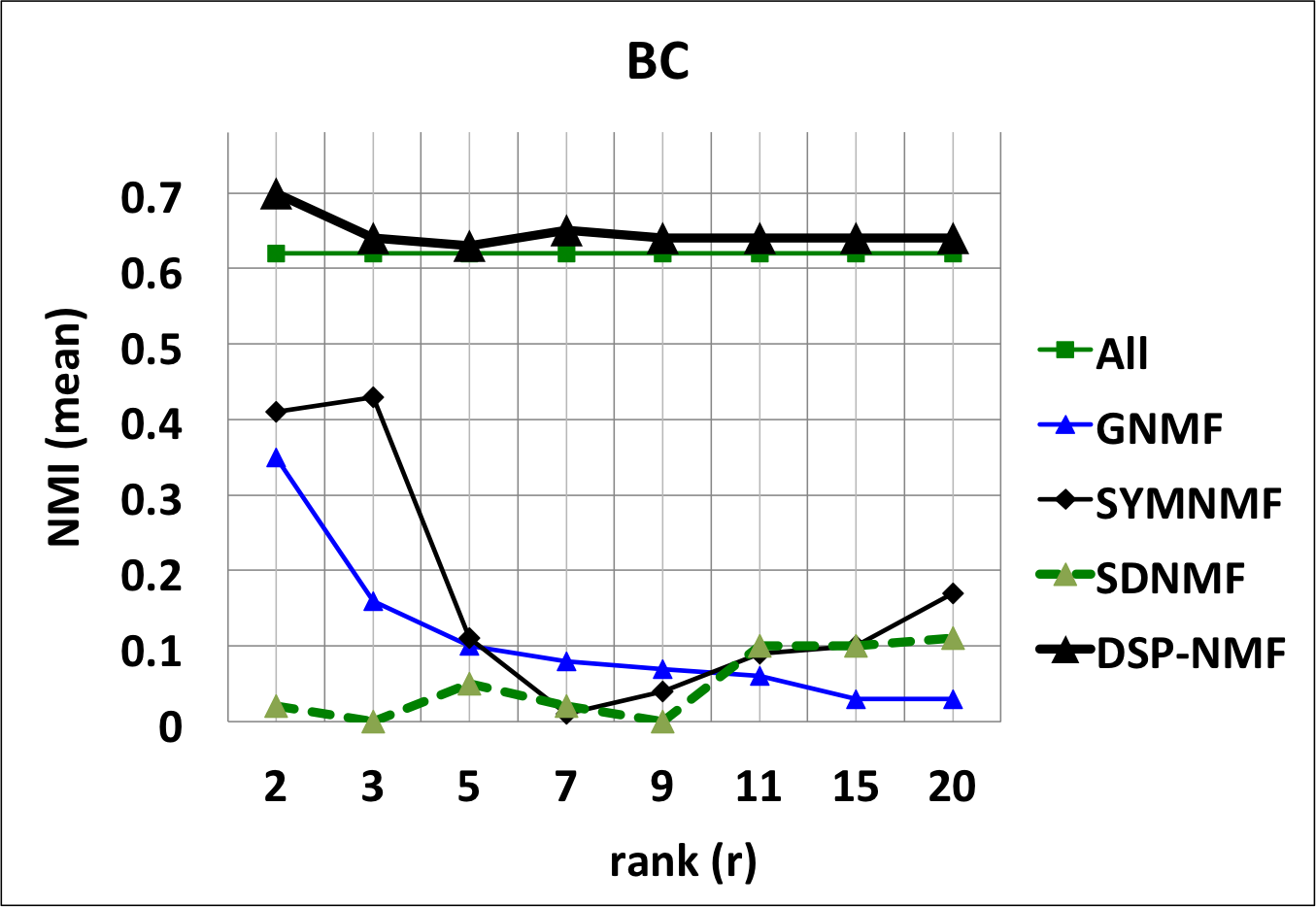}&\hspace{-3ex}
			\includegraphics[scale=0.165]{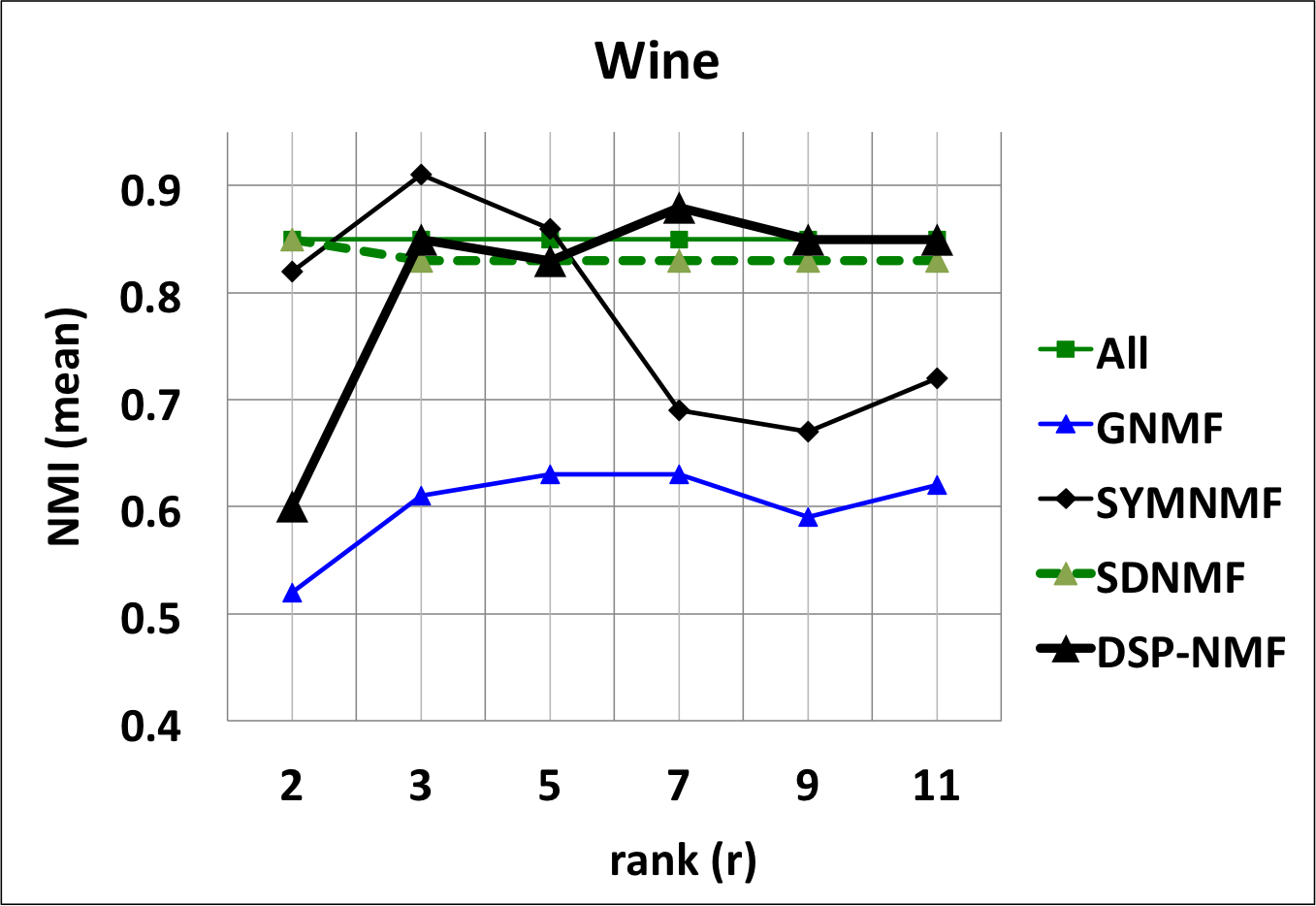}&\hspace{-3ex}
			\includegraphics[scale=0.165]{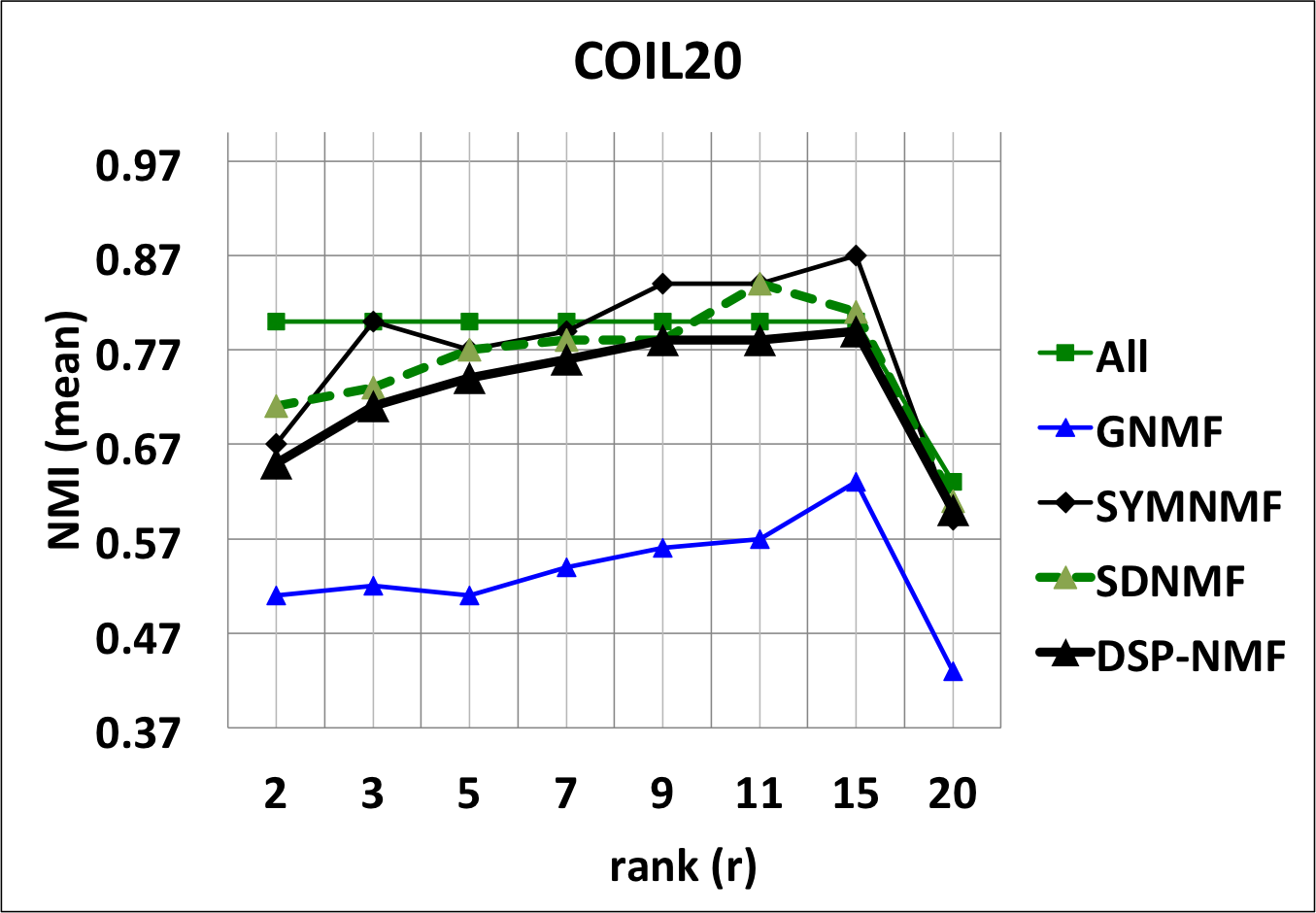}\\
			\includegraphics[scale=0.165]{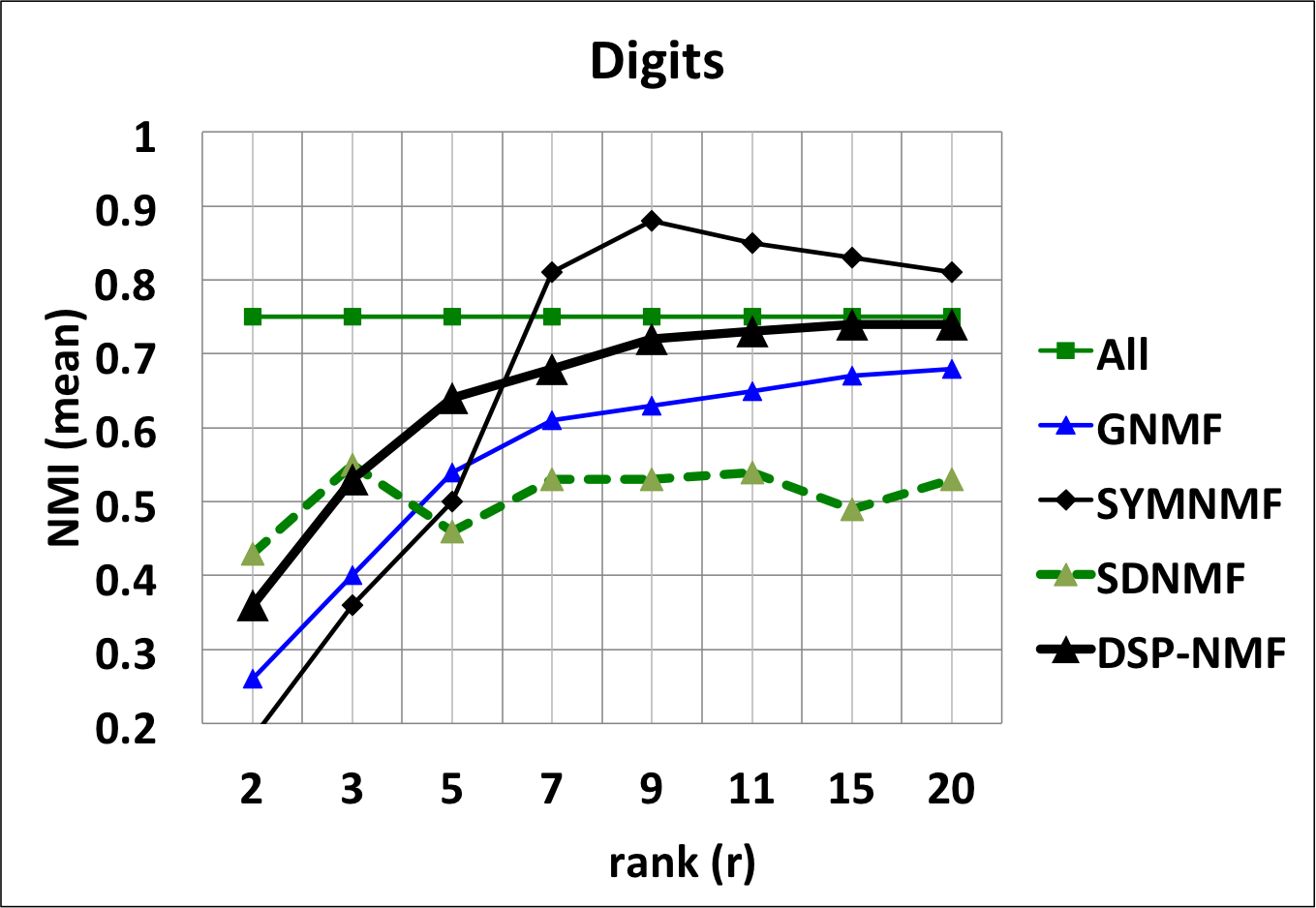}&\hspace{-3ex}
			\includegraphics[scale=0.165]{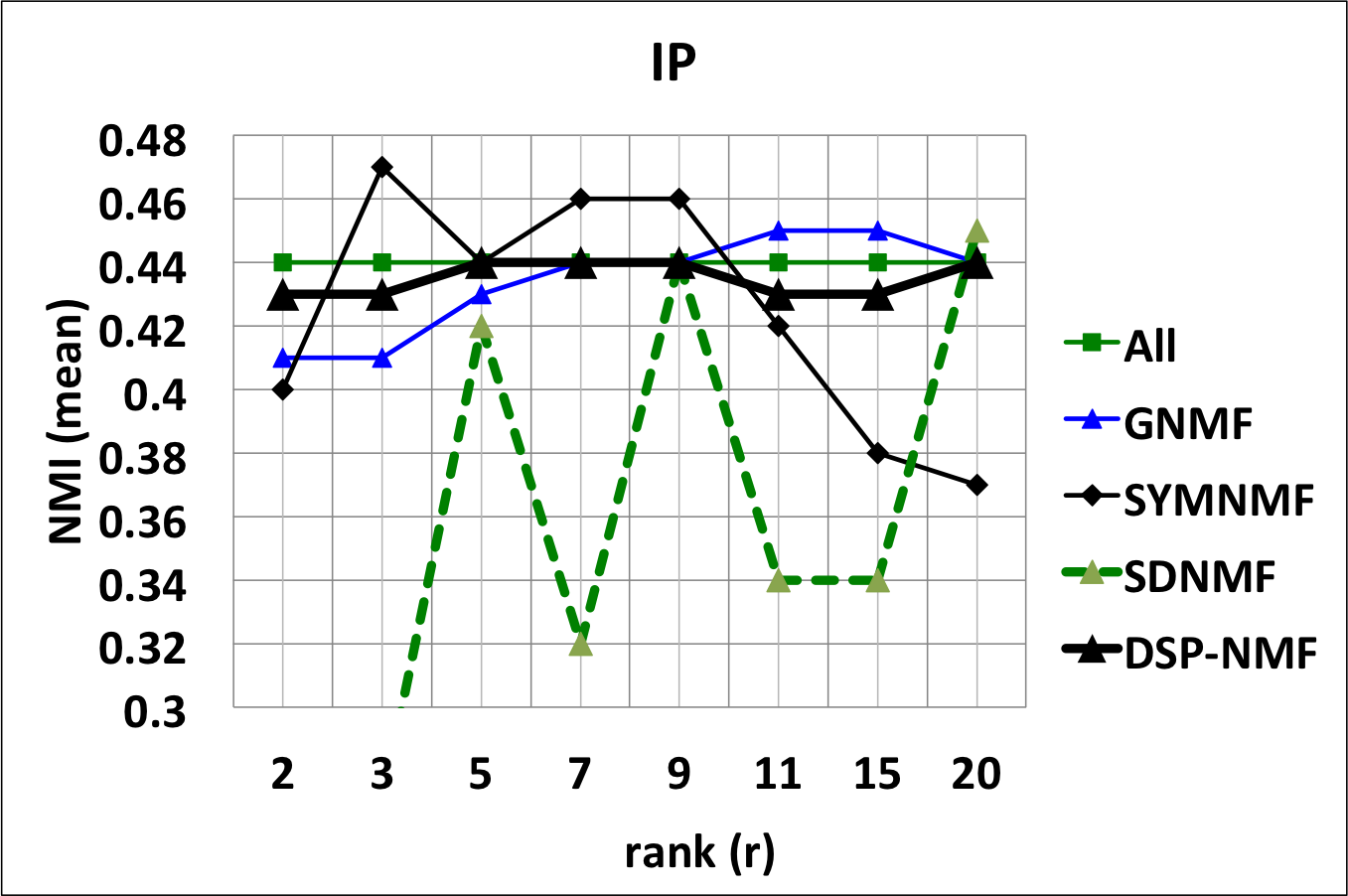}&\hspace{-3ex}
			\includegraphics[scale=0.165]{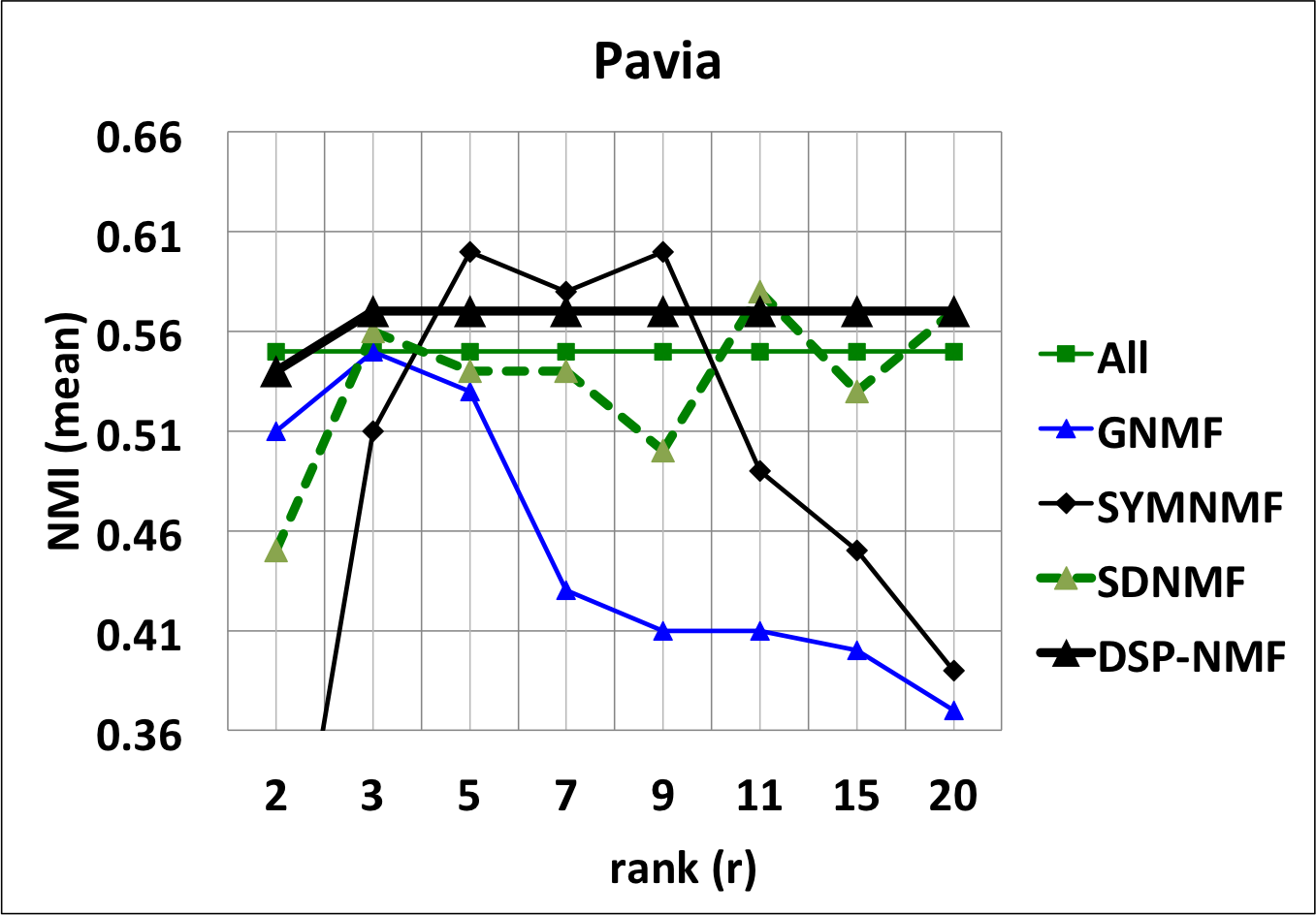}
		\end{tabular}
		\caption{Average K-means clustering performance (NMI) of different NMF-based dimension reduction algorithms.}
		\label{fig-resultskmeans-2}
	\end{figure}

	The parameter $ \lambda$ is set to $ 10^3 $ for all experiments. As mentioned in Section \ref{seq-results}, two comparison experiments were carried out. In the first experiment, we compared our algorithm, DSP-NMF, with GNMF, SymmNMF and SDNMF as they share the same property of preserving data positivity (i.e., NMF-based). In the second experiment, we compared all the NMF-based algorithms to the manifold-based algorithms, UMAP, LLE and SSE, because they share the same property of preserving the neighborhood structure (local invariance). Figure \ref{fig-resultssvm-2}, \ref{fig-resultsknn-2} and \ref{fig-resultskmeans-2} show the results of the first experiment in terms of \textbf{\textit{Acc}} (for SVM and KNN) and \textbf{\textit{NMI}} (for K-means) for all the six datasets after performing feature reduction to ranks $r=[2, 3, 5, 7, 9, 11, 15,20]$. A special case for the Wine dataset where the maximum value of $ r $ is 11 because this dataset has only 13 features. The performance of classification and clustering using the original datasets (indicated by `'\textbf{All}'' in the legend of each figure) is reported as well. From the figures, the following observations can be drawn: 
	\begin{itemize}
		\item[i)] In general, the performance of classification and clustering increases as the data lower rank (i.e., $r$) increases. However some algorithms show the contrary in clustering experiment (see Figure \ref{fig-resultskmeans-2}).
		\item[ii)] No single NMF-based algorithm is the best for all datasets. 
		\item[iii)] For some datasets, the SymmNMF and DSP-NMF algorithms may achieve classification/clustering performance better or at least equal to that obtained when using all features (no feature reduction is applied). However this statement is less valid for SVM classifier (see Figure \ref{fig-resultssvm-2}). 
		\item[iv)] For some algorithms such as GNMF, SymmNMF and DSP-NMF, classification/clustering performance is very poor for small values of $r$ ($\leq3$), but increases considerably when it exceeds 3 (for instance). In contrast, SDNMF does not change much with the change of $ r $.
		\item[v)] Remote sensing datasets (IP and Pavia) seem to be the most challenging (see Figs. \ref{fig-resultsknn-2}, \ref{fig-resultssvm-2} \ref{fig-resultskmeans-2}), while the Digits dataset is the easiest one. However for smaller values of $r$ (e.g., 2 to 5) the BC dataset is the easiest.
	\end{itemize}
	
	Overall, the performance curves shown in Figures \ref{fig-resultssvm-2}, \ref{fig-resultsknn-2} and \ref{fig-resultskmeans-2} indicate that the proposed algorithm, DSP-NMF, is the most effective in generating relevant lower rank data at least for the datasets considered. In some cases, e.g., BC, IP and Pavia datasets, it is far better than other algorithms.

\begin{figure}[!htp]
		\centering
		\begin{tabular}{ccc}
			\includegraphics[scale=0.165]{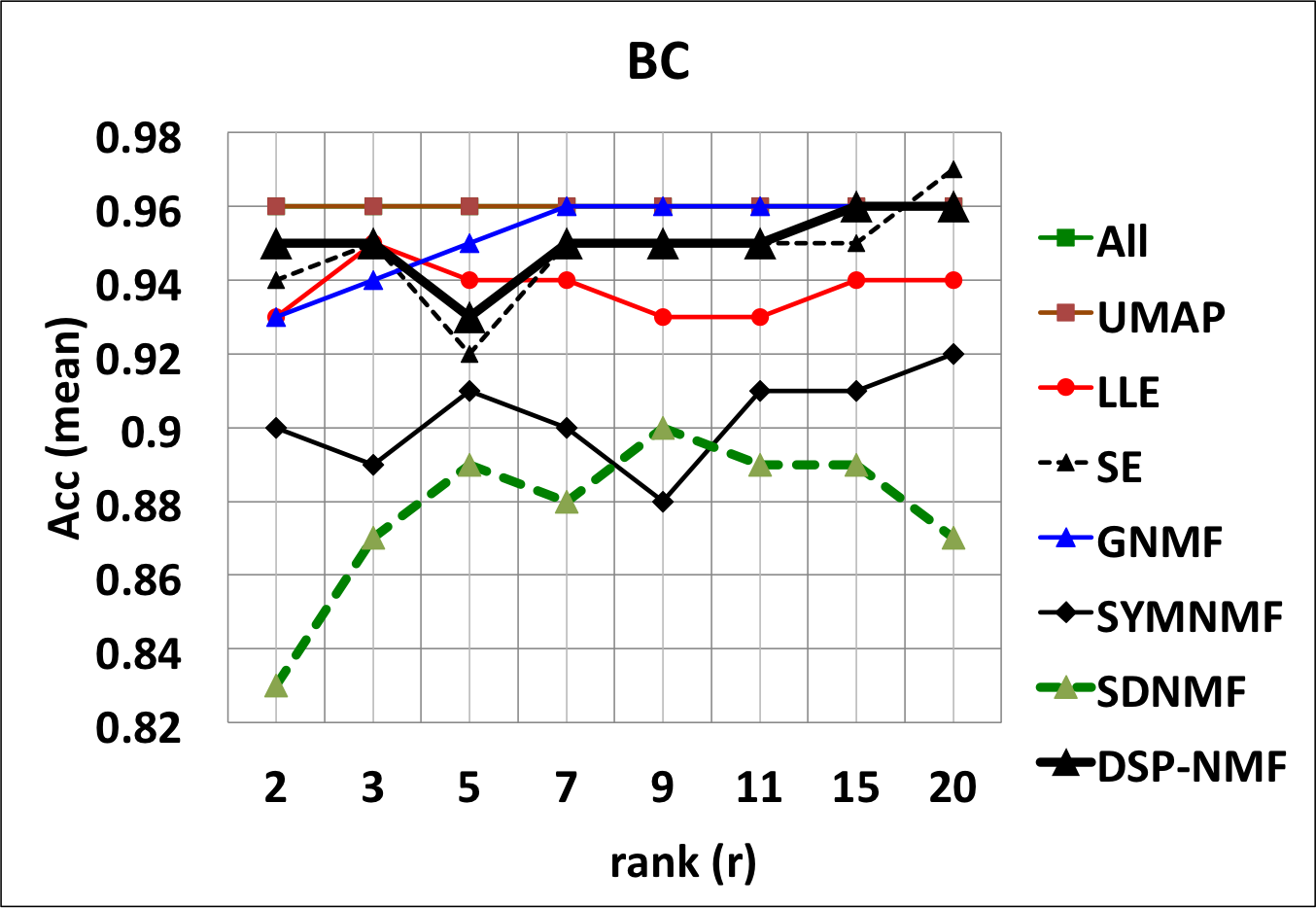}&\hspace{-3ex}
			\includegraphics[scale=0.165]{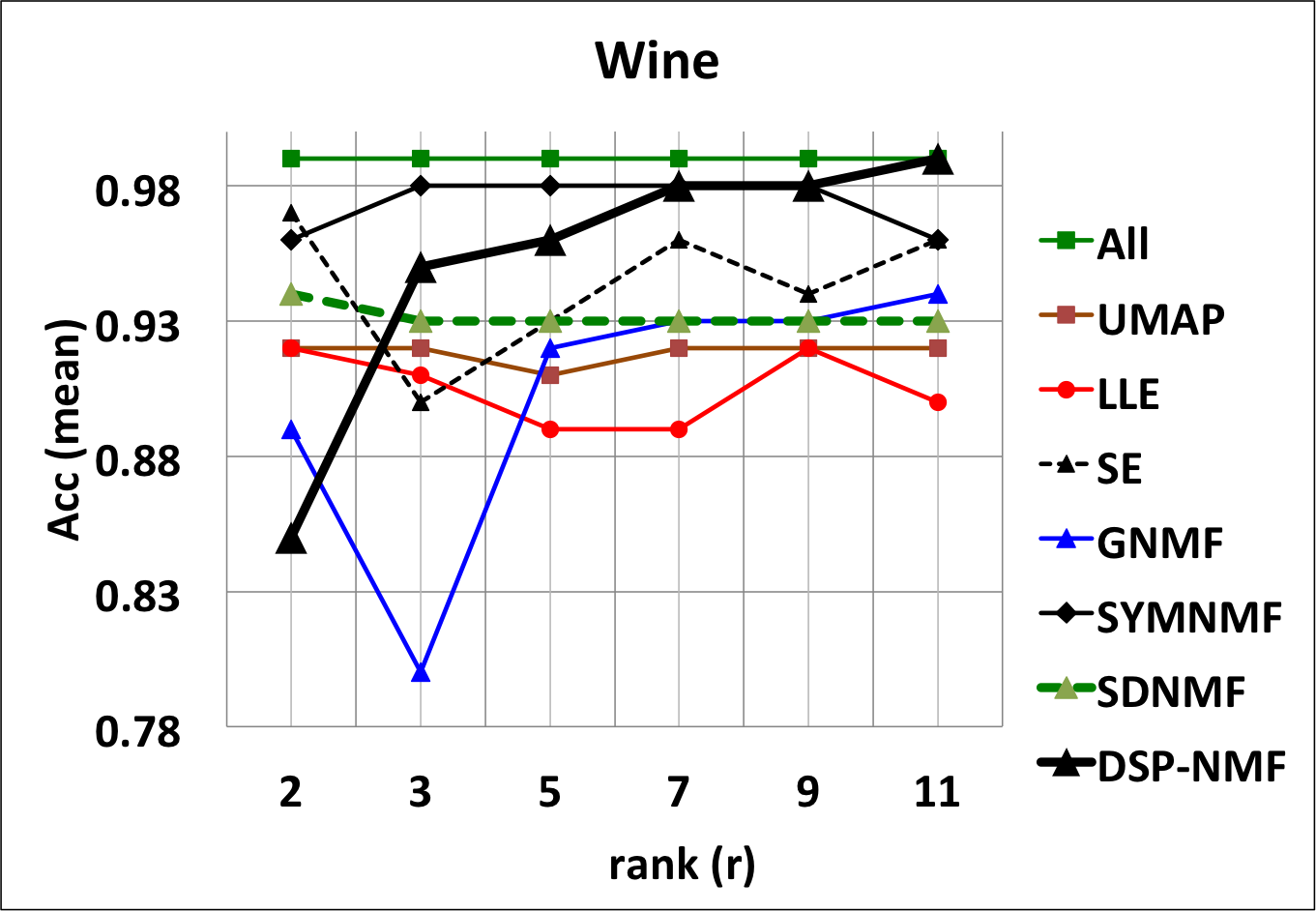}&\hspace{-3ex}
			\includegraphics[scale=0.165]{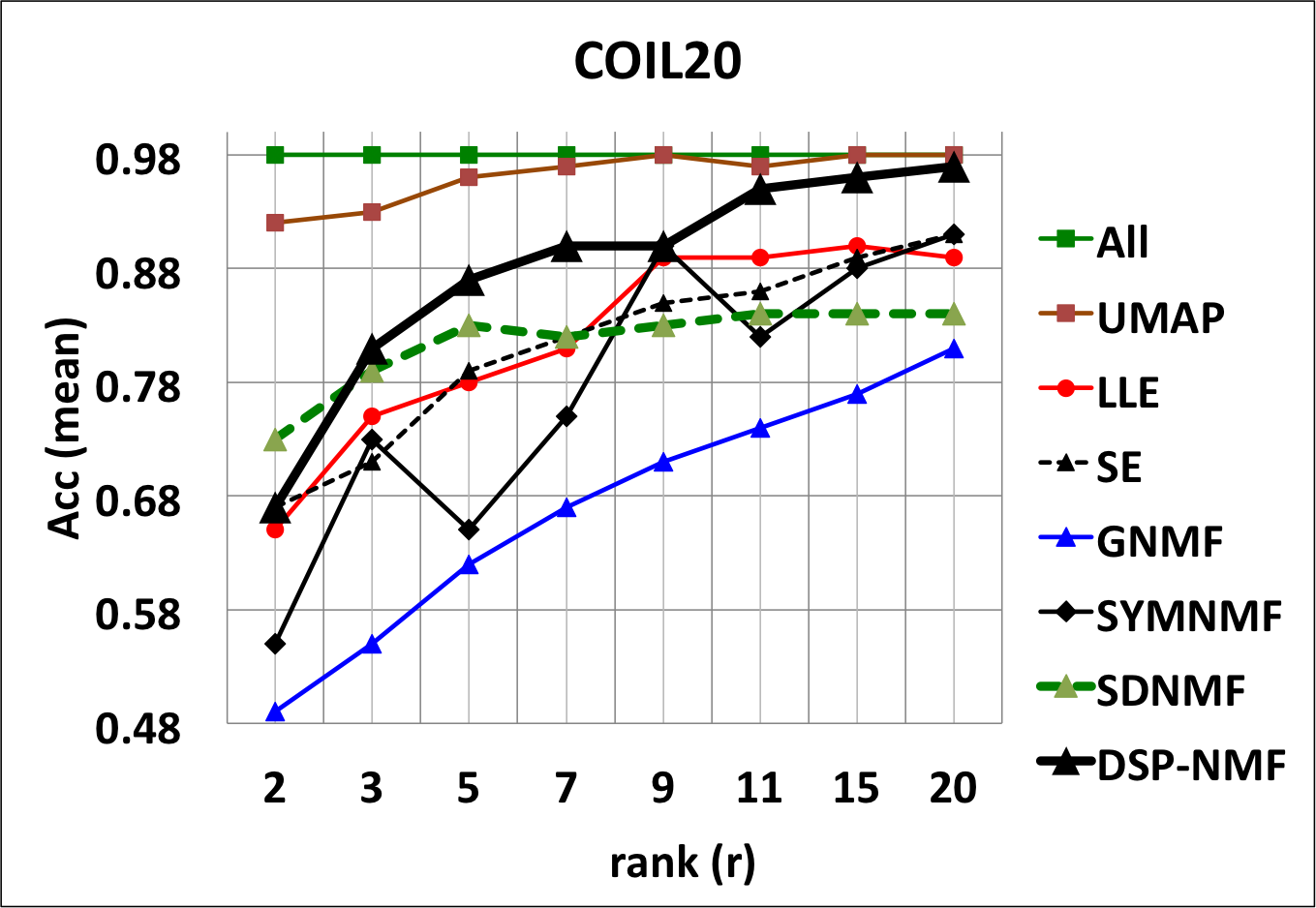}\\
			\includegraphics[scale=0.165]{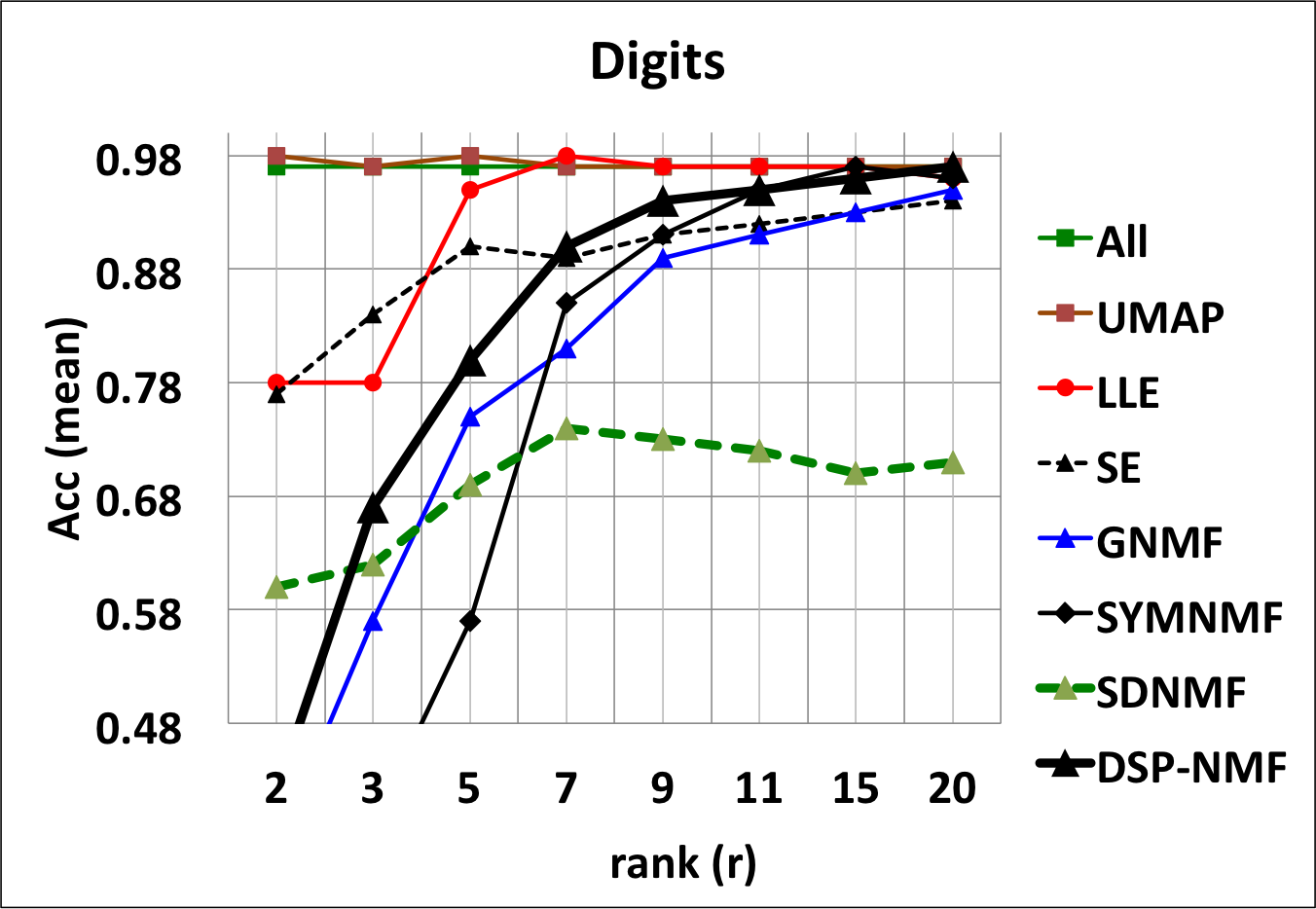}&\hspace{-3ex}
			\includegraphics[scale=0.165]{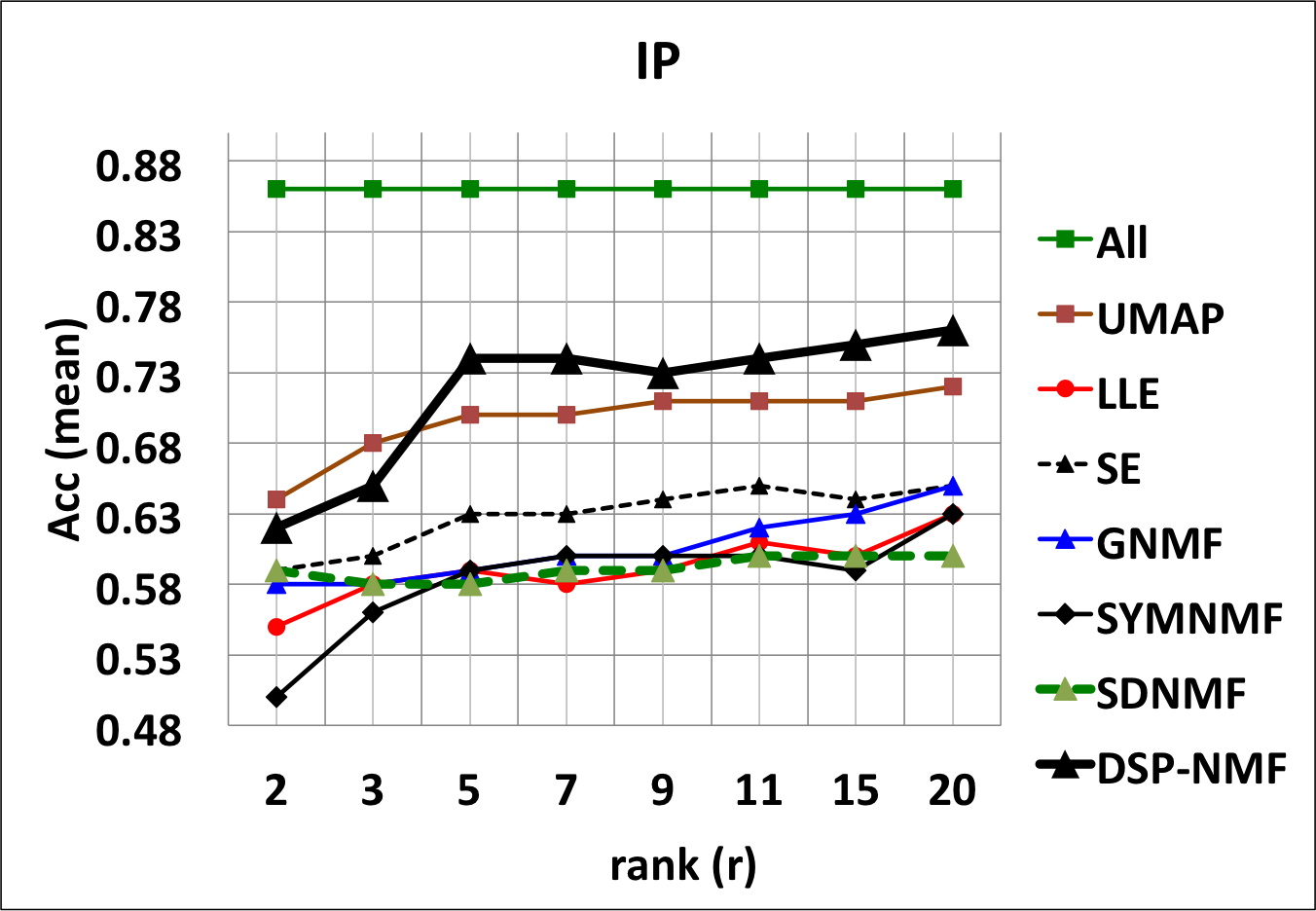}&\hspace{-3ex}
			\includegraphics[scale=0.165]{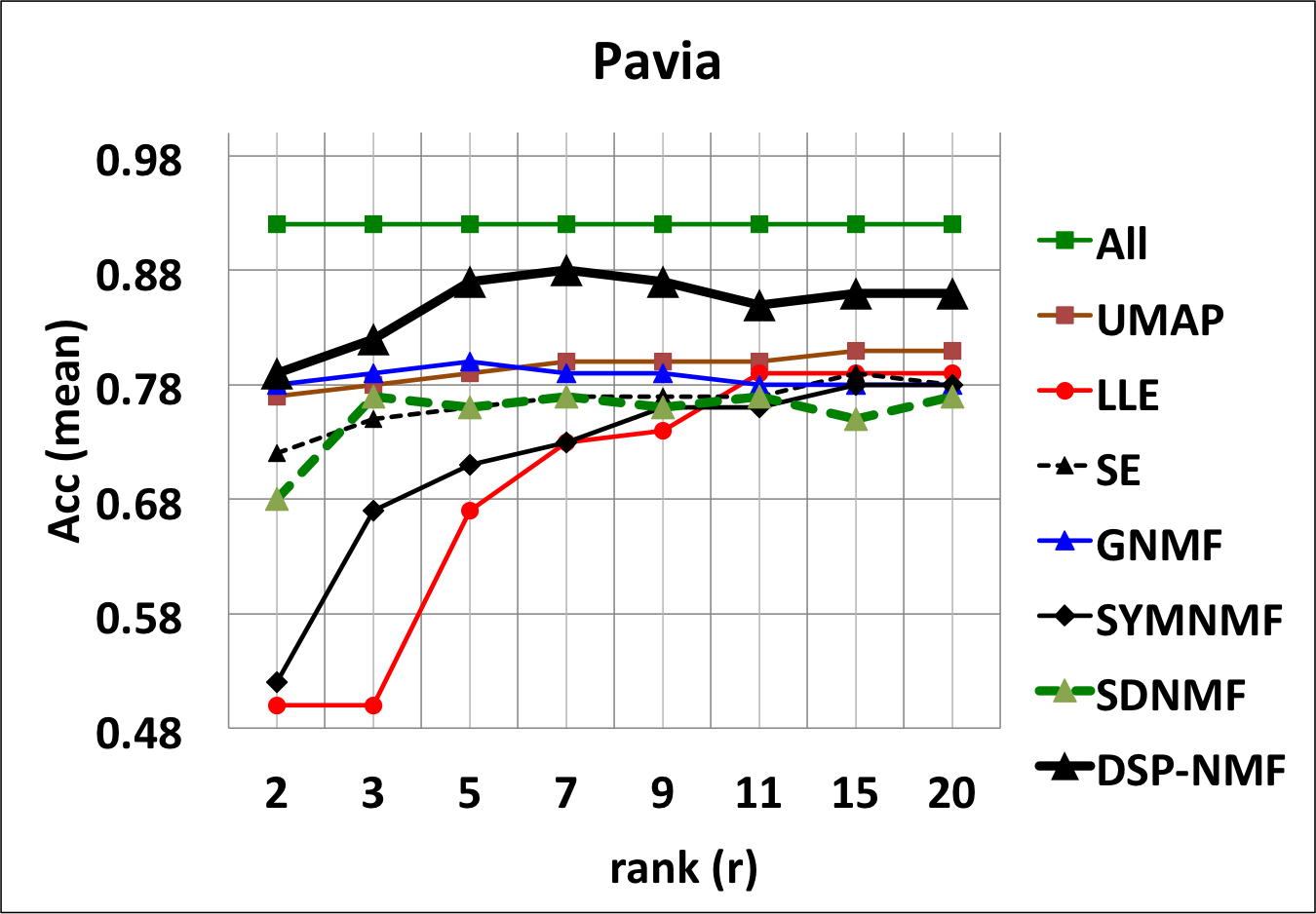}
		\end{tabular}
		\caption{Average SVM accuracy (Acc) of different NMF-based  and manifold-based dimension reduction algorithms.}
		\label{fig-resultssvm-1}
	\end{figure}

\begin{figure}[!htp]
		\centering
		\begin{tabular}{ccc}
			\includegraphics[scale=0.165]{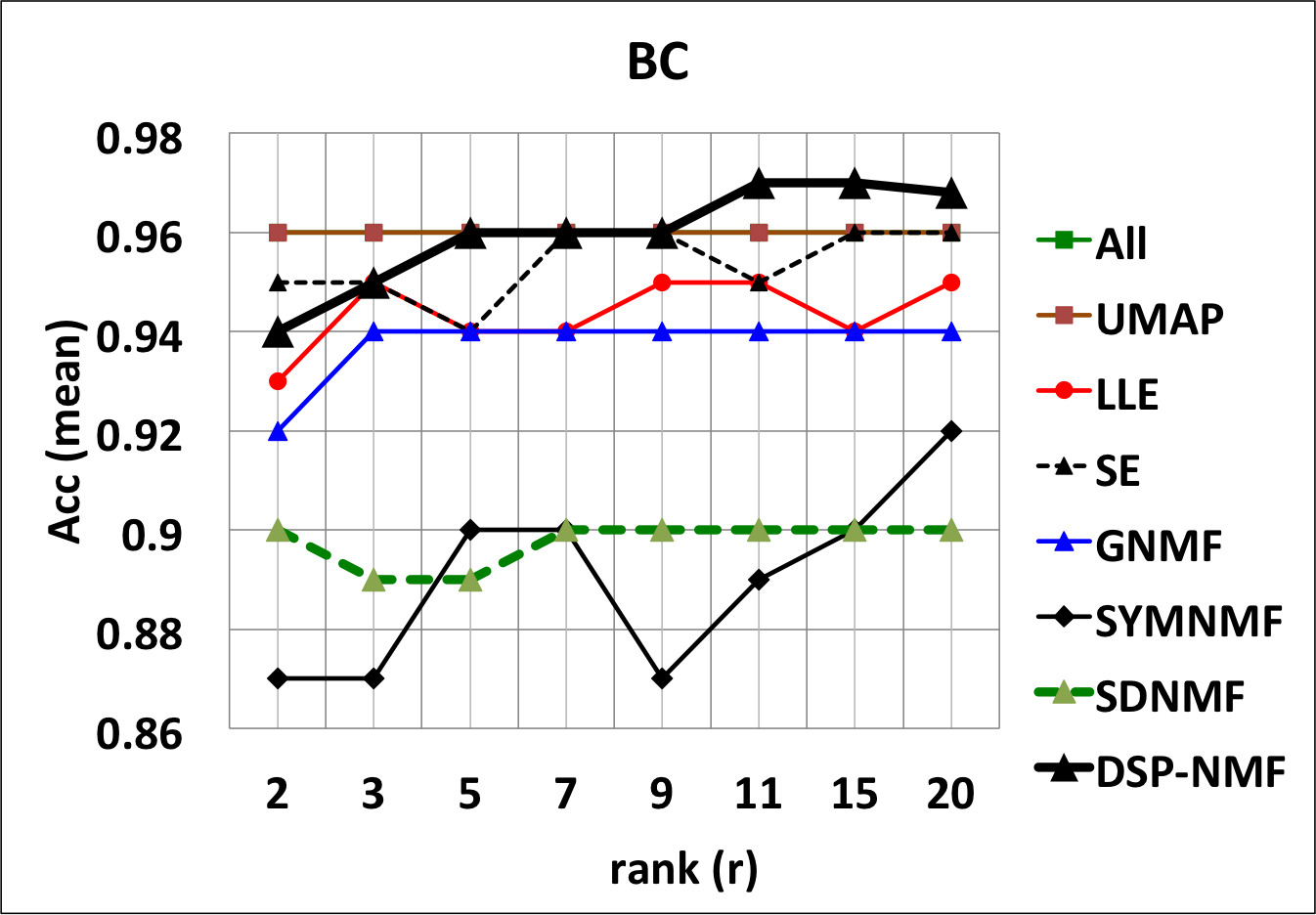}&\hspace{-3ex}
			\includegraphics[scale=0.165]{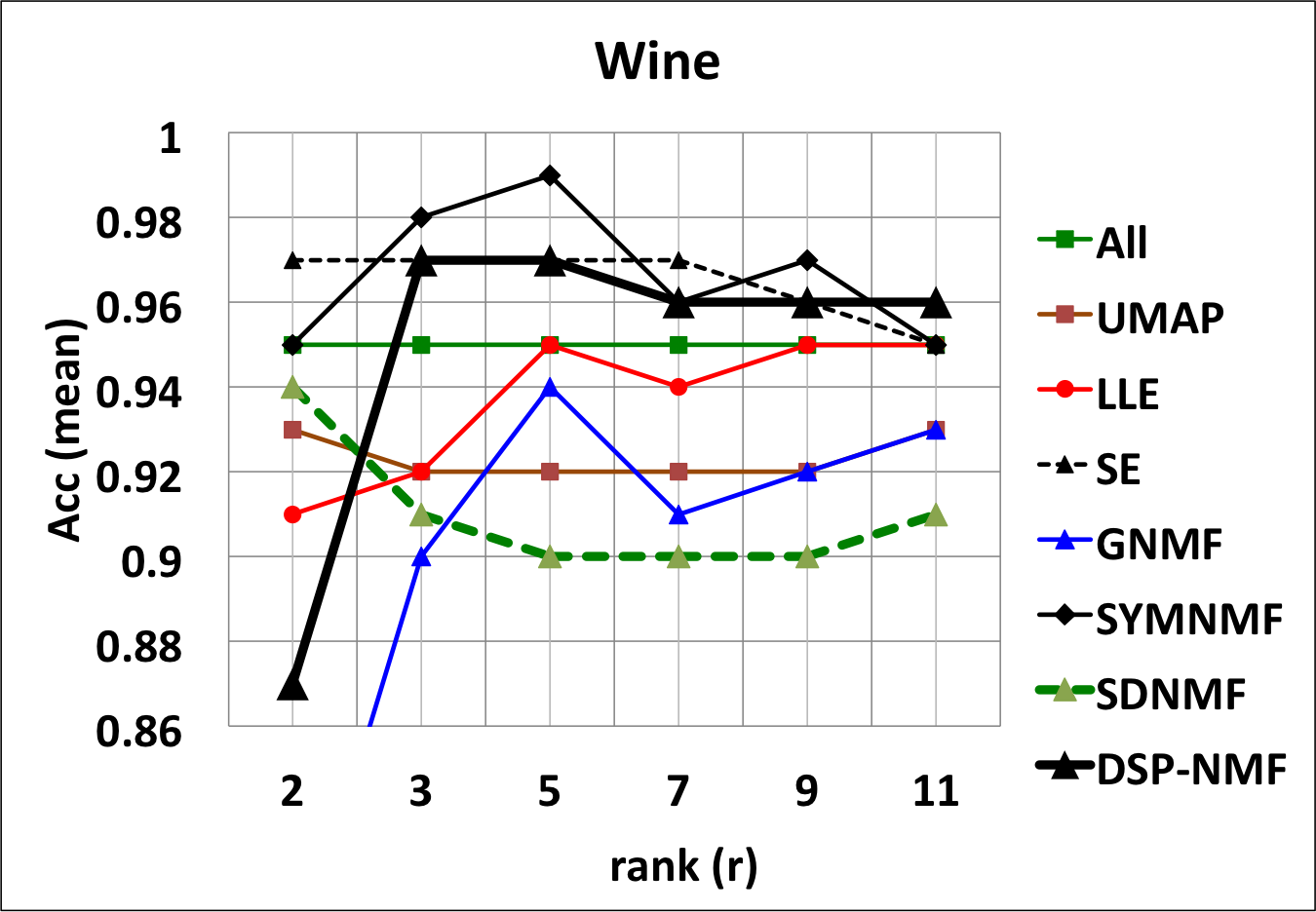}&\hspace{-3ex}
			\includegraphics[scale=0.165]{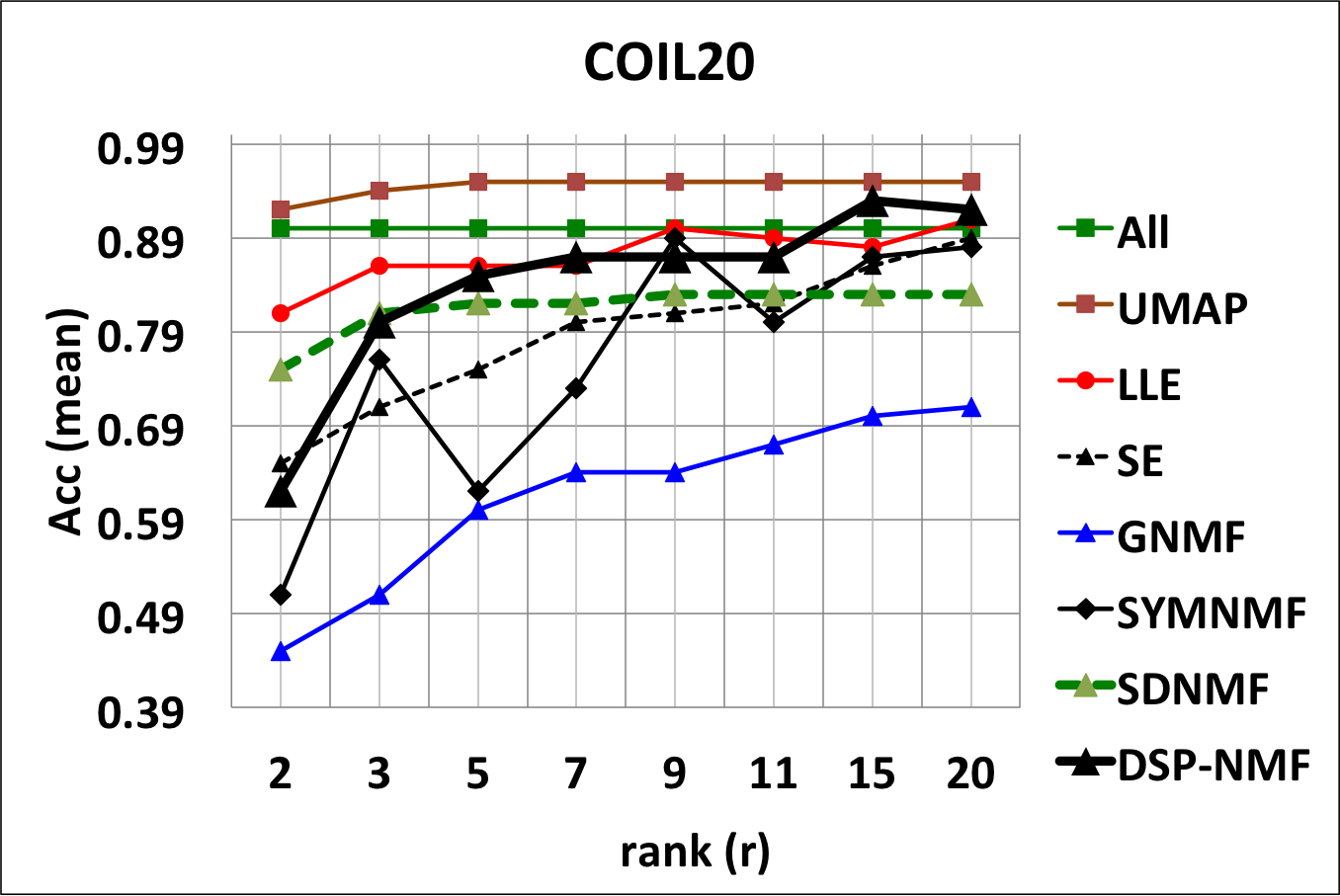}\\
			\includegraphics[scale=0.165]{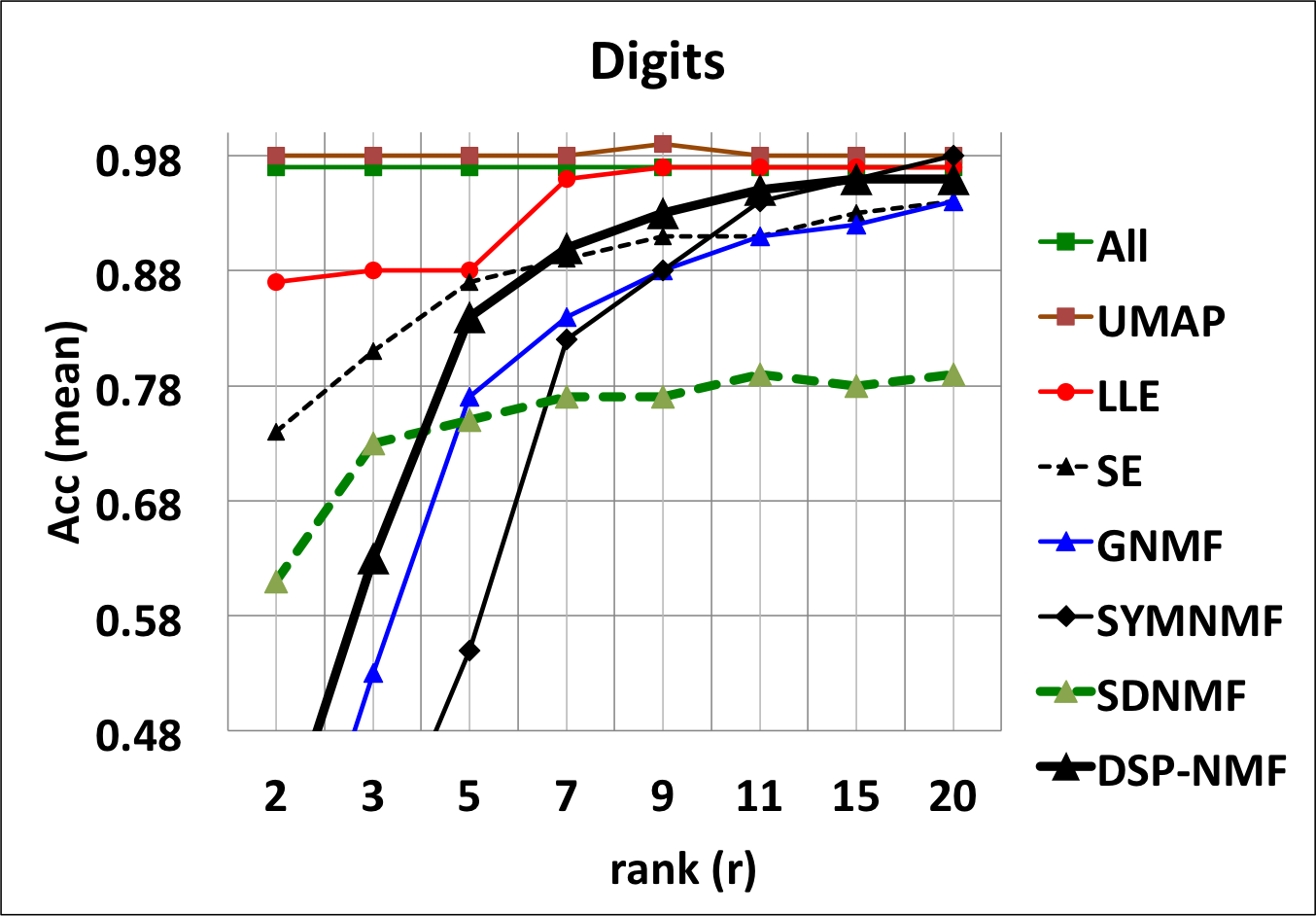}&\hspace{-3ex}
			\includegraphics[scale=0.165]{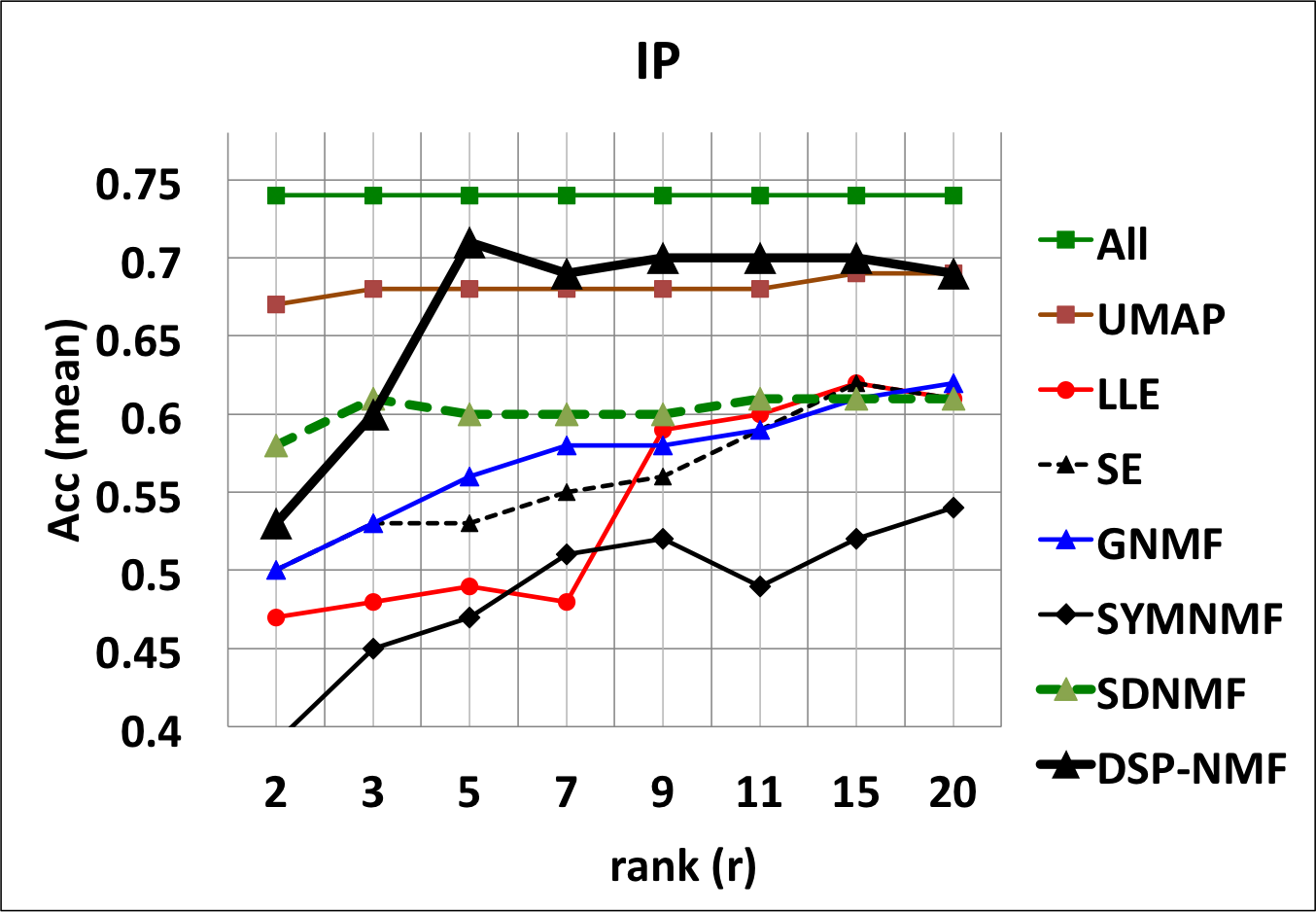}&\hspace{-3ex}
			\includegraphics[scale=0.165]{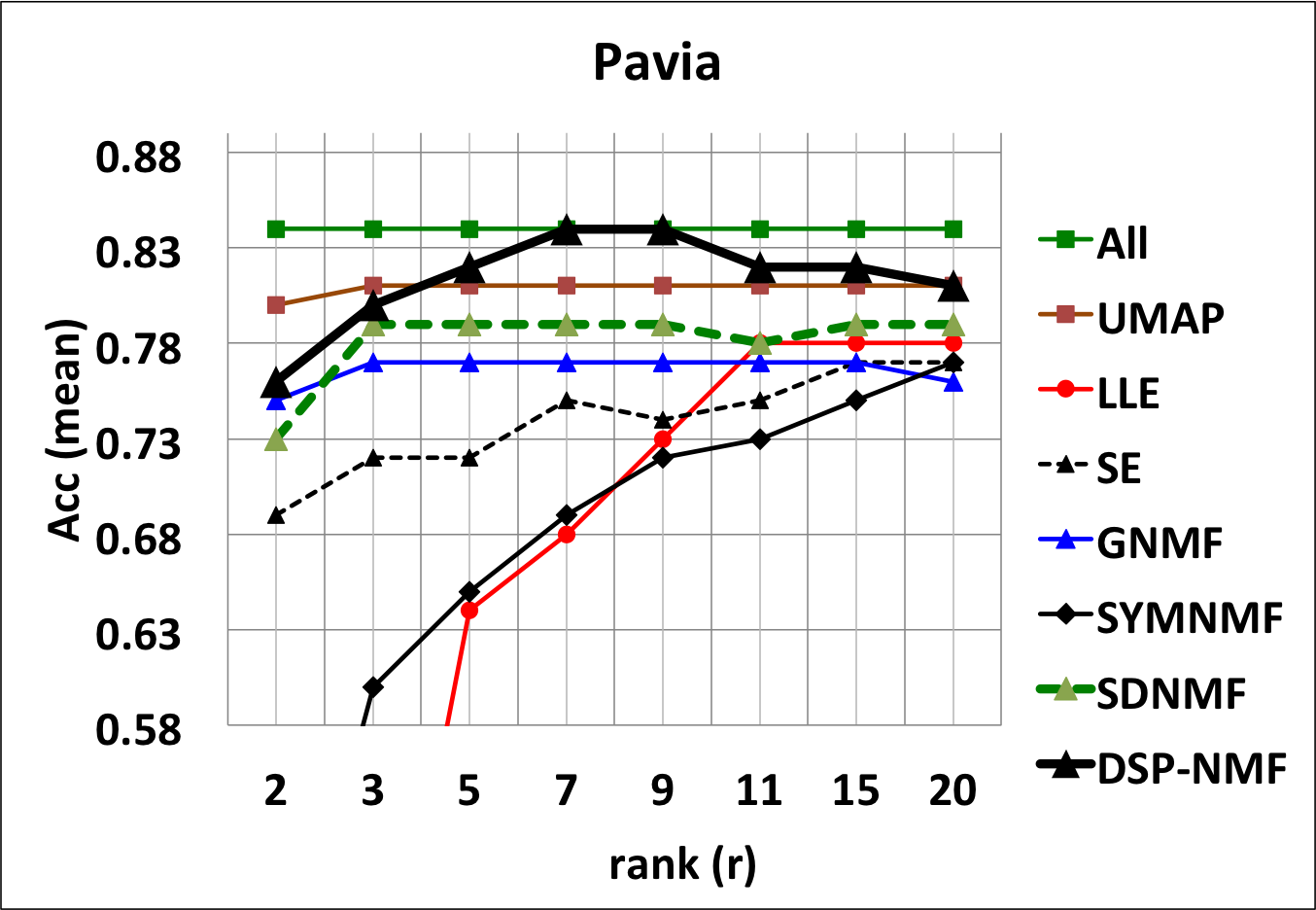}
		\end{tabular}
		\caption{Average KNN accuracy (Acc) of different NMF-based  and manifold-based dimension reduction algorithms.}
		\label{fig-resultsknn-1}
	\end{figure}

\begin{figure}[!htp]
		\centering
		\begin{tabular}{ccc}
			\includegraphics[scale=0.165]{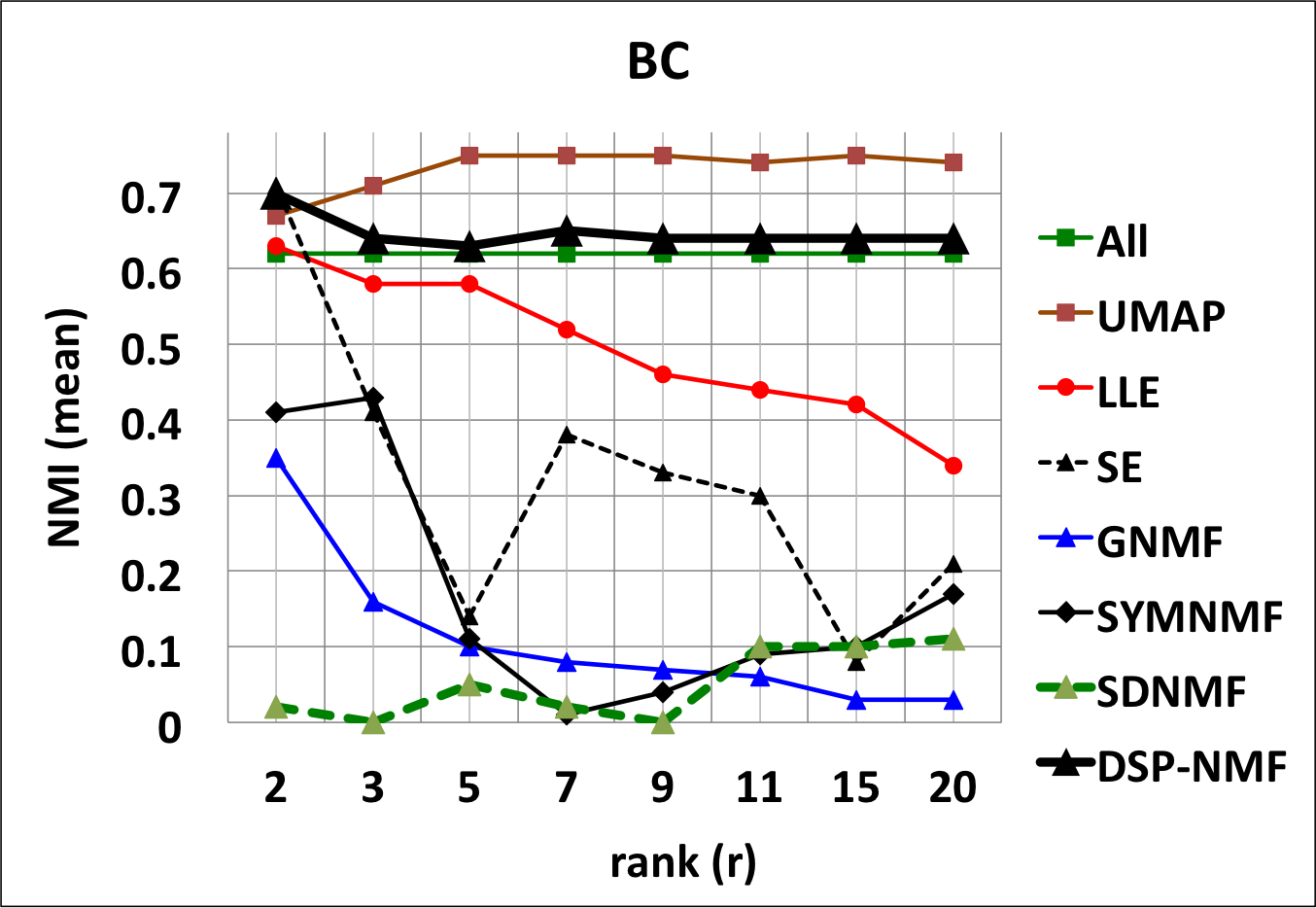}&\hspace{-3ex}
			\includegraphics[scale=0.165]{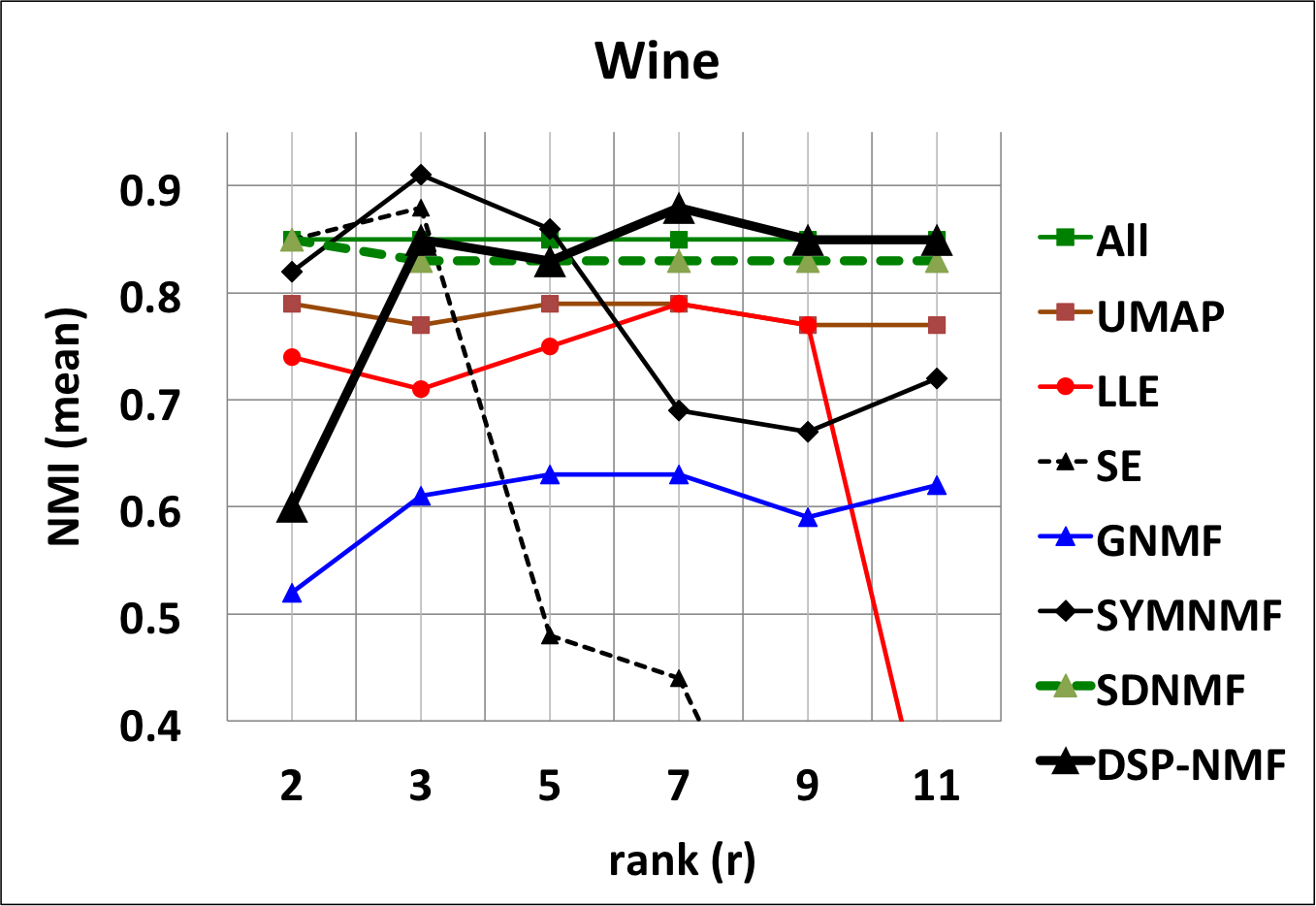}&\hspace{-3ex}
			\includegraphics[scale=0.165]{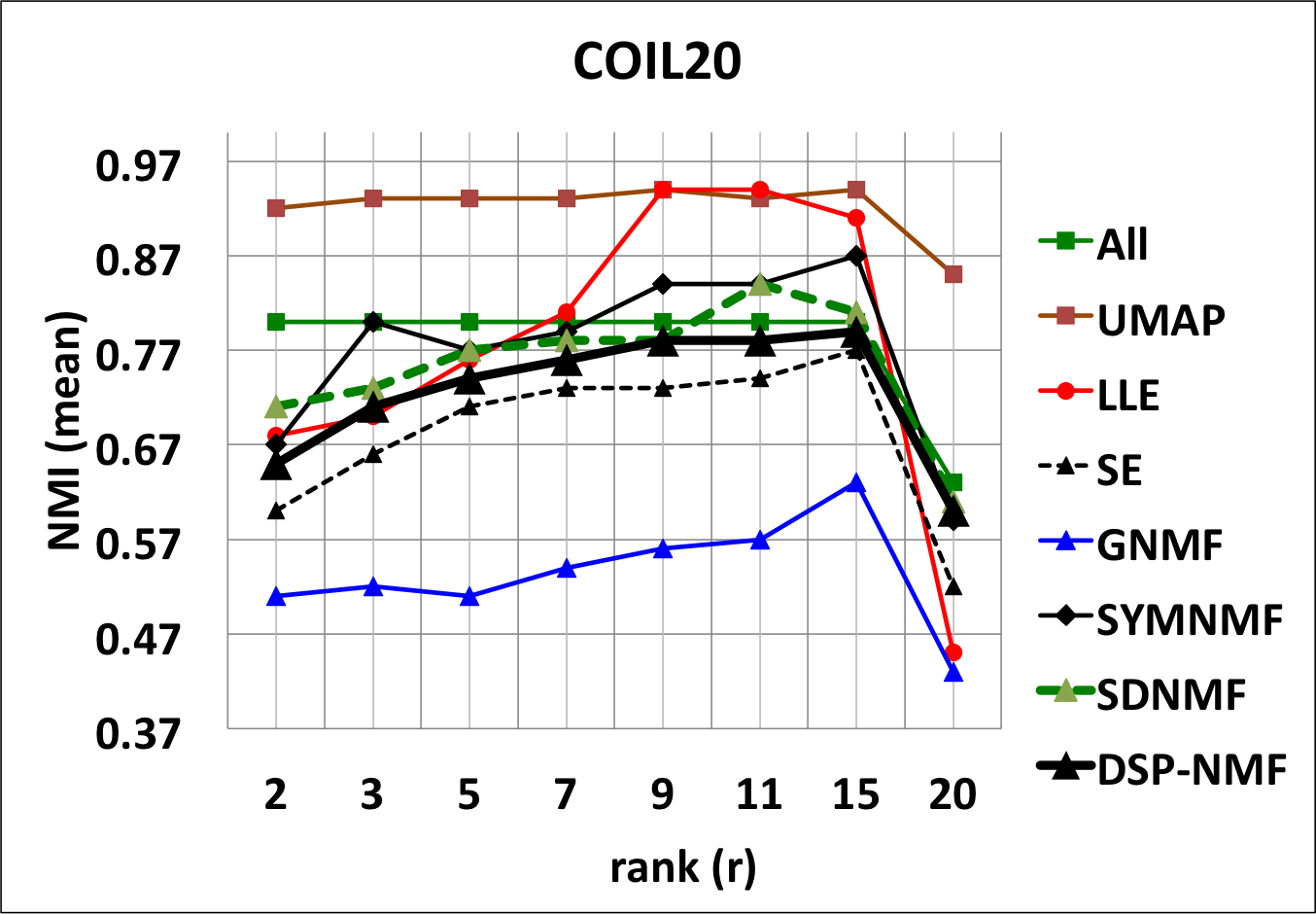}\\
			\includegraphics[scale=0.165]{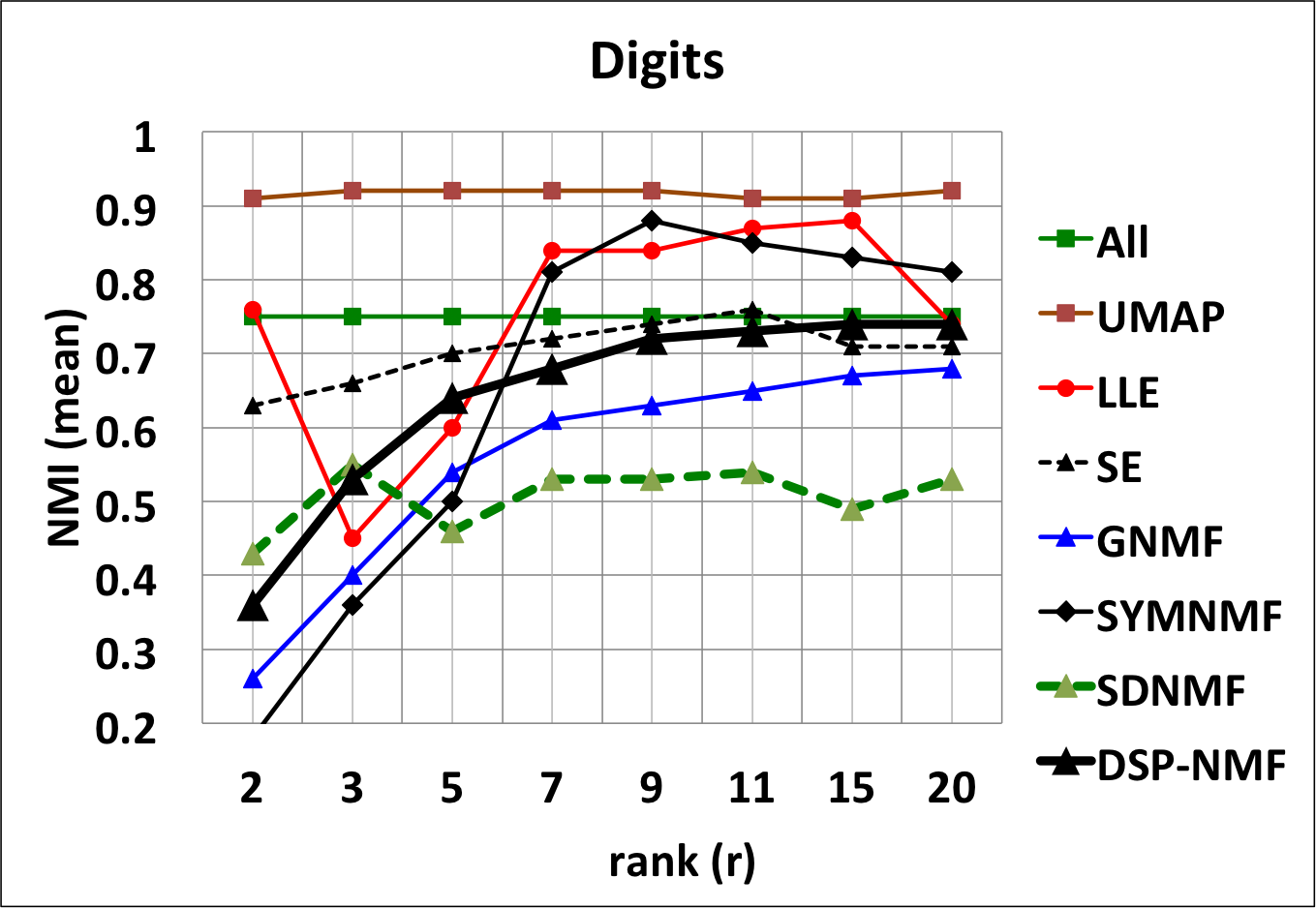}&\hspace{-3ex}
			\includegraphics[scale=0.165]{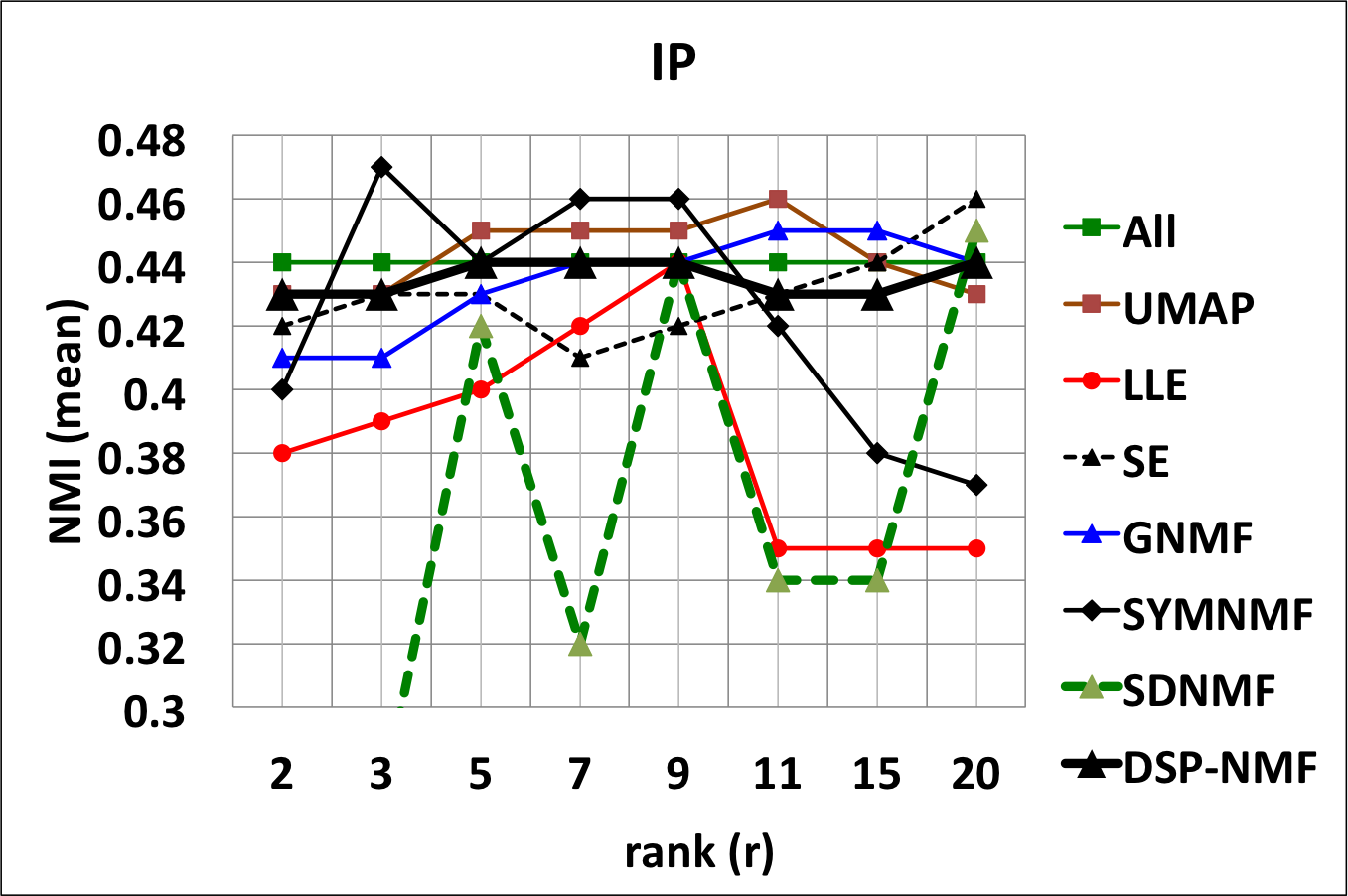}&\hspace{-3ex}
			\includegraphics[scale=0.165]{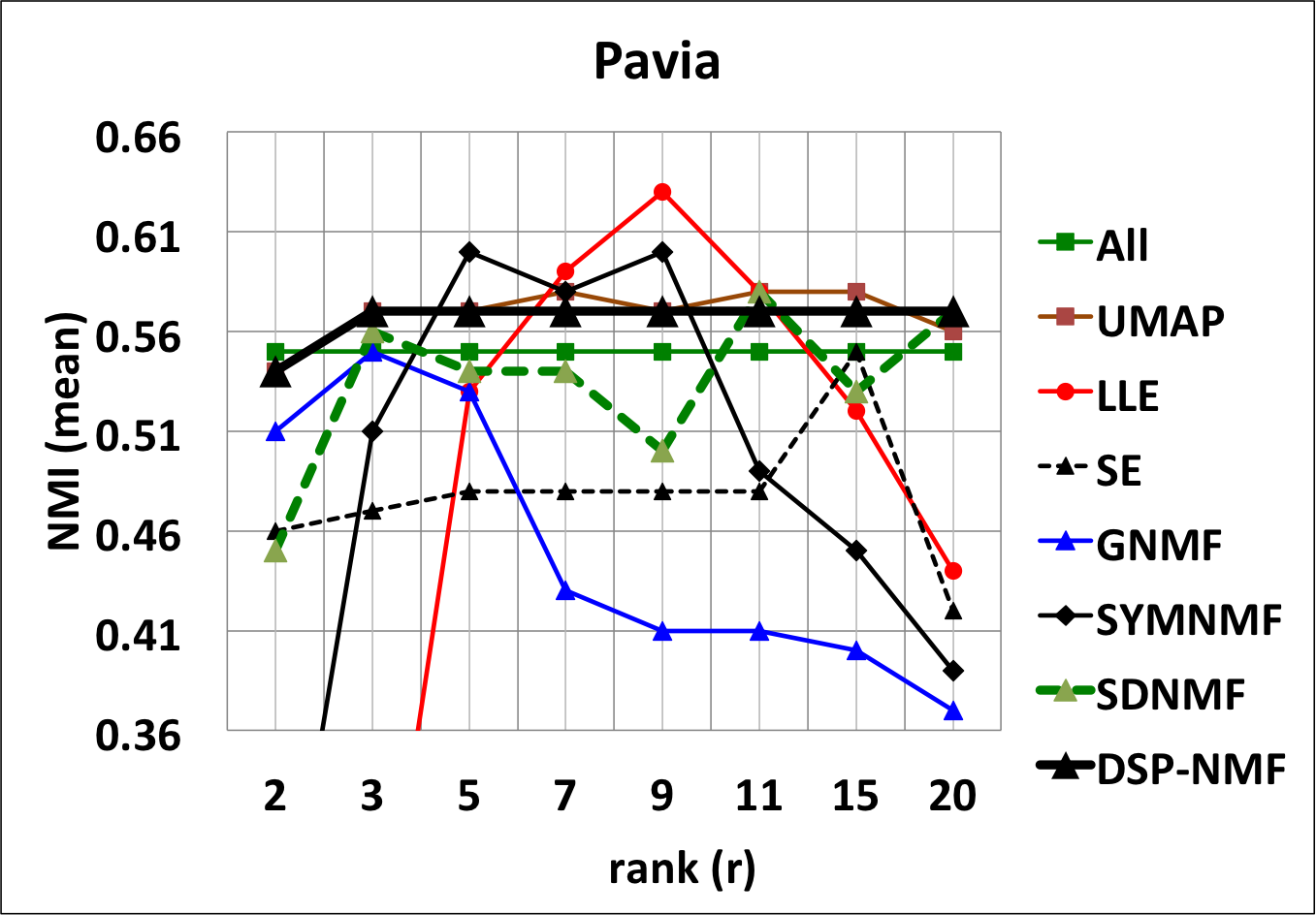}
		\end{tabular}
		\caption{Average K-means clustering performance (NMI) of different NMF-based  and manifold-based dimension reduction algorithms.}
		\label{fig-resultskmeans-1}
	\end{figure}

	In the second experiment, we have compared the NMF-based algorithms against the manifold-based algorithms. The performance curves of classification and  clustering using the reduced data generated  by all the algorithms are shown in Figures \ref{fig-resultssvm-1}, \ref{fig-resultsknn-1} and \ref{fig-resultskmeans-1}. Roughly, the same above mentioned observations were drawn for this experiment as well. On top of that, the UMAP algorithm outperforms all {the others} in clustering in almost all datasets. It {also performs} better with smaller data ranks in both clustering and classification experiments. However, for higher data ranks  DSP-NMF is the best on the majority of the datasets in particular in classification. LLE and SymmNMF are the most competing algorithms to UMAP and DSP-NMF. More objective analysis will be given in the next paragraph.
	%-----------------------
	%-----------------------
	
	In order to objectively assess the effect of different feature reduction algorithms on classification and clustering, we reported, in Tables \ref{tab-svm}, \ref{tab-knn} and \ref{tab-K-means}, their average performance for all the ranks $r=[2, 3, 5, 7, 9, 11, 15,20]$. The performance of classification and clustering using the original datasets (no feature reduction is done) is reported in the rows indicated by `\textbf{All}'. The columns \textbf{mean} and \textbf{max} indicate the mean and max performance of each algorithm computed over 5 runs as explained above in Section \ref{seq-results}. The columns \textbf{Avgm} and \textbf{Avgx} show the average of the \textbf{mean} and \textbf{max} values calculated over all datasets, respectively. They represent the overall performance of each algorithm. Regarding the NMF-based algorithms, based on \textbf{Avgm} and \textbf{Avgx} our algorithm outperforms all other algorithms in classification and in clustering. Based on the \textbf{mean} performance, our algorithm provides also the best results in classification and in clustering for almost all the datasets. In some counter-example cases, other algorithms are better, e.g., {[SymmNMF and KNN algorithms based on the Wine dataset and \textbf{mean} and \textbf{max} measures], [SymmNMF and KNN algorithms based on Digits dataset and \textbf{max} measure], [SymmNMF and SVM algorithms based on the Wine dataset and \textbf{mean} measures]}. Based on the \textbf{max} performance, {SymmNMF} is the best in clustering.
	
	\begin{table}[!htp]\tiny
		\centering
		\caption{SVM performance (\textbf{\textit{Acc}}) of the different algorithms.}
		\label{tab-svm}
		\begin{tabular}{p{0.35cm}p{0.25cm}p{0.25cm}|p{0.25cm}p{0.25cm}|p{0.25cm}p{0.25cm}|p{0.25cm}p{0.25cm}|p{0.25cm}p{0.25cm}|p{0.25cm}p{0.25cm}|p{0.25cm}p{0.25cm}}
			\cline{2-15}
			%	& \multicolumn{14}{c|}{\textbf{SVM (\textit{Acc})}}                                                                                                                                                                                                                                                                    \\ \cline{2-15} 
			& \multicolumn{2}{c|}{\textbf{BC}} & \multicolumn{2}{c|}{\textbf{Wine}} & \multicolumn{2}{c|}{\textbf{COL20}} & \multicolumn{2}{c|}{\textbf{Digits}} & \multicolumn{2}{c|}{\textbf{IP}} & \multicolumn{2}{c|}{\textbf{Pavia}} &                                    &                                    \\ \hline
			\multicolumn{1}{|l|}{\textbf{Algos.}}  & \textbf{mn}   & \textbf{mx}   & \textbf{mn}    & \textbf{mx}    & \textbf{mn}     & \textbf{mx}    & \textbf{mn}     & \textbf{mx}     & \textbf{mn}   & \textbf{mx}   & \textbf{mn}     & \textbf{mx}    & \multicolumn{1}{c|}{\textbf{Avgm}} & \multicolumn{1}{c|}{\textbf{Avgx}} \\ \hline
			\multicolumn{1}{|l|}{\textbf{All}}     & 0.96            & 0.96           & 0.99             & 0.99            & 0.98              & 0.98            & 0.97              & 0.97             & 0.98            & 0.99           & 0.92              & 0.92            & \multicolumn{1}{c|}{0.97}          & \multicolumn{1}{c|}{0.97}          \\ \hline
			\multicolumn{1}{|l|}{\textbf{UMAP}}    & 0.96            & 0.96           & 0.92             & 0.92            & 0.96              & 0.98            & 0.97              & 0.98             & 0.96            & 0.98           & 0.8               & 0.81            & \multicolumn{1}{c|}{0.93}          & \multicolumn{1}{c|}{0.94}          \\ \hline
			\multicolumn{1}{|l|}{\textbf{LLE}}     & 0.94            & 0.95           & 0.91             & 0.92            & 0.82              & 0.9             & 0.92              & 0.98             & 0.92            & 0.98           & 0.69              & 0.79            & \multicolumn{1}{c|}{0.87}          & \multicolumn{1}{c|}{0.92}          \\ \hline
			\multicolumn{1}{|l|}{\textbf{SE}}      & 0.95            & 0.97           & 0.94             & 0.97            & 0.81              & 0.91            & 0.89              & 0.94             & 0.92            & 0.97           & 0.76              & 0.79            & \multicolumn{1}{c|}{0.88}          & \multicolumn{1}{c|}{0.93}          \\ \hline \hline
			\multicolumn{1}{|l|}{\textbf{GNMF}}    & 0.95            & 0.96           & 0.9              & 0.94            & 0.67              & 0.81            & 0.77              & 0.95             & 0.87            & 0.96           & 0.79              & 0.8             & \multicolumn{1}{c|}{0.83}          & \multicolumn{1}{c|}{0.90}          \\ \hline
			\multicolumn{1}{|l|}{\textbf{SymmNMF}} & 0.9             & 0.92           & 0.97             & 0.98            & 0.77              & 0.91            & 0.73              & 0.97             & 0.89            & 0.98           & 0.71              & 0.78            & \multicolumn{1}{c|}{0.83}          & \multicolumn{1}{c|}{0.92}          \\ \hline
			\multicolumn{1}{|l|}{\textbf{SDNMF}}   & 0.88            & 0.9            & 0.93             & 0.94            & 0.82              & 0.84            & 0.69              & 0.74             & 0.84            & 0.94           & 0.75              & 0.77            & \multicolumn{1}{c|}{0.82}          & \multicolumn{1}{c|}{0.86}          \\ \hline
			\multicolumn{1}{|l|}{\textbf{DSP-NMF}} & 0.95            & 0.96           & 0.95             & 0.99            & 0.88              & 0.97            & 0.83              & 0.97             & 0.94            & 0.99           & 0.85              & 0.88            & \multicolumn{1}{c|}{0.90}          & \multicolumn{1}{c|}{0.96}          \\ \hline
		\end{tabular}
	\end{table}
	
	\begin{table}[!htp]\tiny
		\centering
		\caption{KNN performance (\textbf{\textit{Acc}}) of the different algorithms.}
		\label{tab-knn}
		\begin{tabular}{p{0.35cm}p{0.25cm}p{0.25cm}|p{0.25cm}p{0.25cm}|p{0.25cm}p{0.25cm}|p{0.25cm}p{0.25cm}|p{0.25cm}p{0.25cm}|p{0.25cm}p{0.25cm}|p{0.25cm}p{0.25cm}}
			\cline{2-15}
			%	& \multicolumn{14}{c|}{\textbf{KNN (\textit{Acc})}}                                                                                                                                                                                                                                                                   \\ \cline{2-15} 
			& \multicolumn{2}{c|}{\textbf{BC}} & \multicolumn{2}{c|}{\textbf{Wine}} & \multicolumn{2}{c|}{\textbf{COL20}} & \multicolumn{2}{c|}{\textbf{Digits}} & \multicolumn{2}{c|}{\textbf{IP}} & \multicolumn{2}{c|}{\textbf{Pavia}} & \textbf{}                          & \textbf{}                         \\ \hline
			\multicolumn{1}{|l|}{\textbf{Algos.}}  & \textbf{mean}   & \textbf{max}   & \textbf{mean}    & \textbf{max}    & \textbf{mean}     & \textbf{max}    & \textbf{mean}     & \textbf{max}     & \textbf{mean}   & \textbf{max}   & \textbf{mean}     & \textbf{max}    & \multicolumn{1}{c|}{\textbf{Avgm}} & \multicolumn{1}{c|}{\textbf{Avgx}} \\ \hline
			\multicolumn{1}{|l|}{\textbf{All}}     & 0.96            & 0.96           & 0.95             & 0.95            & 0.9               & 0.9             & 0.97              & 0.97             & 0.74            & 0.74           & 0.84              & 0.84            & \multicolumn{1}{c|}{0.89}          & \multicolumn{1}{c|}{0.89}         \\ \hline
			\multicolumn{1}{|l|}{\textbf{UMAP}}    & 0.96            & 0.96           & 0.92             & 0.93            & 0.95              & 0.95            & 0.98              & 0.99             & 0.68            & 0.69           & 0.81              & 0.81            & \multicolumn{1}{c|}{0.88}          & \multicolumn{1}{c|}{0.89}         \\ \hline
			\multicolumn{1}{|l|}{\textbf{LLE}}     & 0.94            & 0.95           & 0.94             & 0.95            & 0.87              & 0.91            & 0.93              & 0.97             & 0.54            & 0.62           & 0.64              & 0.78            & \multicolumn{1}{c|}{0.81}          & \multicolumn{1}{c|}{0.86}         \\ \hline
			\multicolumn{1}{|l|}{\textbf{SE}}      & 0.95            & 0.96           & 0.97             & 0.97            & 0.79              & 0.89            & 0.88              & 0.94             & 0.56            & 0.62           & 0.74              & 0.77            & \multicolumn{1}{c|}{0.82}          & \multicolumn{1}{c|}{0.86}         \\ \hline \hline
			\multicolumn{1}{|l|}{\textbf{GNMF}}    & 0.94            & 0.94           & 0.9              & 0.94            & 0.62              & 0.71            & 0.76              & 0.94             & 0.57            & 0.62           & 0.77              & 0.77            & \multicolumn{1}{c|}{0.76}          & \multicolumn{1}{c|}{0.82}         \\ \hline
			\multicolumn{1}{|l|}{\textbf{SymmNMF}} & 0.89            & 0.92           & 0.97             & 0.99            & 0.76              & 0.89            & 0.72              & 0.98             & 0.49            & 0.54           & 0.67              & 0.77            & \multicolumn{1}{c|}{0.75}          & \multicolumn{1}{c|}{0.85}         \\ \hline
			\multicolumn{1}{|l|}{\textbf{SDNMF}}   & 0.9             & 0.9            & 0.91             & 0.94            & 0.82              & 0.83            & 0.75              & 0.79             & 0.6             & 0.61           & 0.78              & 0.79            & \multicolumn{1}{c|}{0.79}          & \multicolumn{1}{c|}{0.81}         \\ \hline
			\multicolumn{1}{|l|}{\textbf{DSP-NMF}} & 0.96            & 0.97           & 0.95             & 0.97            & 0.84              & 0.93            & 0.82              & 0.96             & 0.67            & 0.71           & 0.81              & 0.84            & \multicolumn{1}{c|}{0.84}          & \multicolumn{1}{c|}{0.90}         \\ \hline
		\end{tabular}
	\end{table}

	To summarize the analysis of the results, we computed the number of times each algorithm ranks first (based on \textbf{mean} and \textbf{max} performances) over all the datasets, and report them in Tables \ref{tab-ranking-nmf} and \ref{tab-ranking-all}. 
	
	Table \ref{tab-ranking-nmf}, which concerns the NMF-based algorithms shows clearly that the proposed algorithm outperforms all others. For instance, based on the \textbf{mean} performance, the proposed algorithm is the best in classification and in clustering. However, based on \textbf{max} performance, SymmNMF is the best for K-means clustering and the proposed algorithm is the second best. For SVM classification, the proposed algorithm ranked first on all the datasets. In many cases, GNMF and SDNMF never ranked first. 
	
	\begin{table}[!htp]\tiny
		\centering
		\caption{K-means performance (\textbf{NMI}) of the different algorithms.}
		\label{tab-K-means}
		\begin{tabular}{p{0.35cm}p{0.25cm}p{0.25cm}|p{0.25cm}p{0.25cm}|p{0.25cm}p{0.25cm}|p{0.25cm}p{0.25cm}|p{0.25cm}p{0.25cm}|p{0.25cm}p{0.25cm}|p{0.25cm}p{0.25cm}}
			\cline{2-15}
			%	& \multicolumn{14}{c|}{\textbf{K-means (\textit{NMI})}}                                                                                                                                                                                                                                                           \\ \cline{2-15} 
			& \multicolumn{2}{c|}{\textbf{BC}} & \multicolumn{2}{c|}{\textbf{Wine}} & \multicolumn{2}{c|}{\textbf{COL20}} & \multicolumn{2}{c|}{\textbf{Digits}} & \multicolumn{2}{c|}{\textbf{IP}} & \multicolumn{2}{c|}{\textbf{Pavia}} & \textbf{}                          & \textbf{}                          \\ \hline
			\multicolumn{1}{|l|}{\textbf{Algos.}}  & \textbf{mean}   & \textbf{max}   & \textbf{mean}    & \textbf{max}    & \textbf{mean}     & \textbf{max}    & \textbf{mean}     & \textbf{max}     & \textbf{mean}   & \textbf{max}   & \textbf{mean}     & \textbf{max}    & \multicolumn{1}{c|}{\textbf{Avgm}} & \multicolumn{1}{c|}{\textbf{Avgx}} \\ \hline
			\multicolumn{1}{|l|}{\textbf{All}}     & 0.62            & 0.62           & 0.85             & 0.85            & 0.78              & 0.8             & 0.75              & 0.75             & 0.44            & 0.44           & 0.55              & 0.55            & \multicolumn{1}{c|}{0.67}          & \multicolumn{1}{c|}{0.67}          \\ \hline
			\multicolumn{1}{|l|}{\textbf{UMAP}}    & 0.73            & 0.75           & 0.78             & 0.79            & 0.92              & 0.94            & 0.92              & 0.92             & 0.44            & 0.46           & 0.57              & 0.58            & \multicolumn{1}{c|}{0.73}          & \multicolumn{1}{c|}{0.74}          \\ \hline
			\multicolumn{1}{|l|}{\textbf{LLE}}     & 0.5             & 0.63           & 0.67             & 0.79            & 0.77              & 0.94            & 0.75              & 0.88             & 0.39            & 0.44           & 0.46              & 0.63            & \multicolumn{1}{c|}{0.59}          & \multicolumn{1}{c|}{0.72}          \\ \hline
			\multicolumn{1}{|l|}{\textbf{SE}}      & 0.32            & 0.71           & 0.49             & 0.88            & 0.68              & 0.77            & 0.7               & 0.76             & 0.43            & 0.46           & 0.48              & 0.55            & \multicolumn{1}{c|}{0.52}          & \multicolumn{1}{c|}{0.69}          \\ \hline \hline
			\multicolumn{1}{|l|}{\textbf{GNMF}}    & 0.11            & 0.35           & 0.6              & 0.63            & 0.53              & 0.63            & 0.56              & 0.68             & 0.43            & 0.45           & 0.45              & 0.55            & \multicolumn{1}{c|}{0.45}          & \multicolumn{1}{c|}{0.55}          \\ \hline
			\multicolumn{1}{|l|}{\textbf{SymmNMF}} & 0.17            & 0.43           & 0.78             & 0.91            & 0.77              & 0.87            & 0.65              & 0.88             & 0.43            & 0.47           & 0.48              & 0.6             & \multicolumn{1}{c|}{0.55}          & \multicolumn{1}{c|}{0.69}          \\ \hline
			\multicolumn{1}{|l|}{\textbf{SDNMF}}   & 0.05            & 0.11           & 0.83             & 0.85            & 0.75              & 0.84            & 0.51              & 0.55             & 0.36            & 0.45           & 0.53              & 0.58            & \multicolumn{1}{c|}{0.51}          & \multicolumn{1}{c|}{0.56}          \\ \hline
			\multicolumn{1}{|l|}{\textbf{DSP-NMF}} & 0.65            & 0.7            & 0.81             & 0.88            & 0.73              & 0.79            & 0.64              & 0.74             & 0.44            & 0.44           & 0.57              & 0.57            & \multicolumn{1}{c|}{0.64}          & \multicolumn{1}{c|}{0.69}          \\ \hline
		\end{tabular}
		
	\end{table}
	
	Table \ref{tab-ranking-all} reports the same statistics for all the algorithms considered. Without a competitor, the UMAP algorithm is the best based on both \textbf{mean} and \textbf{max} performances. The proposed algorithm is the second best, except for K-means clustering, where LLE and SymmNMF are the second best algorithms based on \textbf{max} performance. SSE and GNMF perform the least. Each ranked first only once, as shown in Table \ref{tab-ranking-all}.
	
	\begin{table}[!htp]\footnotesize
		\centering
		\caption{Occurrence of ranking first (NMF-based algorithms). }
		\label{tab-ranking-nmf}
		\begin{tabular}{l|c|c|c||c|c|c|}
			\cline{2-7}
			& \multicolumn{3}{c|}{\textbf{Mean performance}} & \multicolumn{3}{c|}{\textbf{Max performance}} \\ \cline{2-7} 
			& \textbf{KNN}  & \textbf{SVM} & \textbf{K-means} & \textbf{KNN} & \textbf{SVM} & \textbf{K-means} \\ \hline
			\multicolumn{1}{|l|}{\textbf{GNMF}}    & 0             & 1            & 0               & 0            & 1            & 0               \\ \hline
			\multicolumn{1}{|l|}{\textbf{SymmNMF}}  & 1             & 1            & 2               & 2            & 1            & \textbf{5}               \\ \hline
			\multicolumn{1}{|l|}{\textbf{SDNMF}}   & 0             & 0            & 1               & 0            & 0            & 0               \\ \hline
			\multicolumn{1}{|l|}{\textbf{DSP-NMF}} & \textbf{5}             & \textbf{5}            & \textbf{3}               & \textbf{4}            & \textbf{6}           & 1               \\ \hline
		\end{tabular}
		
	\end{table}

	\begin{table}[!htp]\footnotesize
		\centering
		\caption{Occurrence of ranking first (All algorithms). }
		\label{tab-ranking-all}
		\begin{tabular}{l|c|c|c||c|c|c|}
			\cline{2-7}
			& \multicolumn{3}{c|}{\textbf{Mean performance}} & \multicolumn{3}{c|}{\textbf{Max performance}} \\ \cline{2-7} 
			& \textbf{KNN}  & \textbf{SVM} & \textbf{K-means} & \textbf{KNN} & \textbf{SVM} & \textbf{K-means} \\ \hline
			\multicolumn{1}{|l|}{\textbf{UMAP}}    & \textbf{4}    & \textbf{4}            & \textbf{5}               & 2            & \textbf{3}            & \textbf{3}               \\ \hline
			\multicolumn{1}{|l|}{\textbf{LLE}}     & 0             & 0            & 0               & 0            & 0            & 2               \\ \hline
			\multicolumn{1}{|l|}{\textbf{SE}}      & 1             & 0            & 0               & 0            & 0            & 0               \\ \hline
			\multicolumn{1}{|l|}{\textbf{GNMF}}    & 0             & 0            & 0               & 0            & 1            & 0               \\ \hline
			\multicolumn{1}{|l|}{\textbf{SymmNMF}}  & 1             & 1            & 0               & 1            & 0            & 2               \\ \hline
			\multicolumn{1}{|l|}{\textbf{SDNMF}}   & 0             & 0            & 1               & 0            & 0            & 0               \\ \hline
			\multicolumn{1}{|l|}{\textbf{DSP-NMF}} & 2             & 1            & 2               & \textbf{3}            & 1            & 0               \\ \hline
		\end{tabular}
		
	\end{table}
	
	\subsection{Convergence study}
	\label{seq-convergence}
\begin{figure}[!htp]
		\centering
		\begin{tabular}{ccc}
			\includegraphics[scale=0.17]{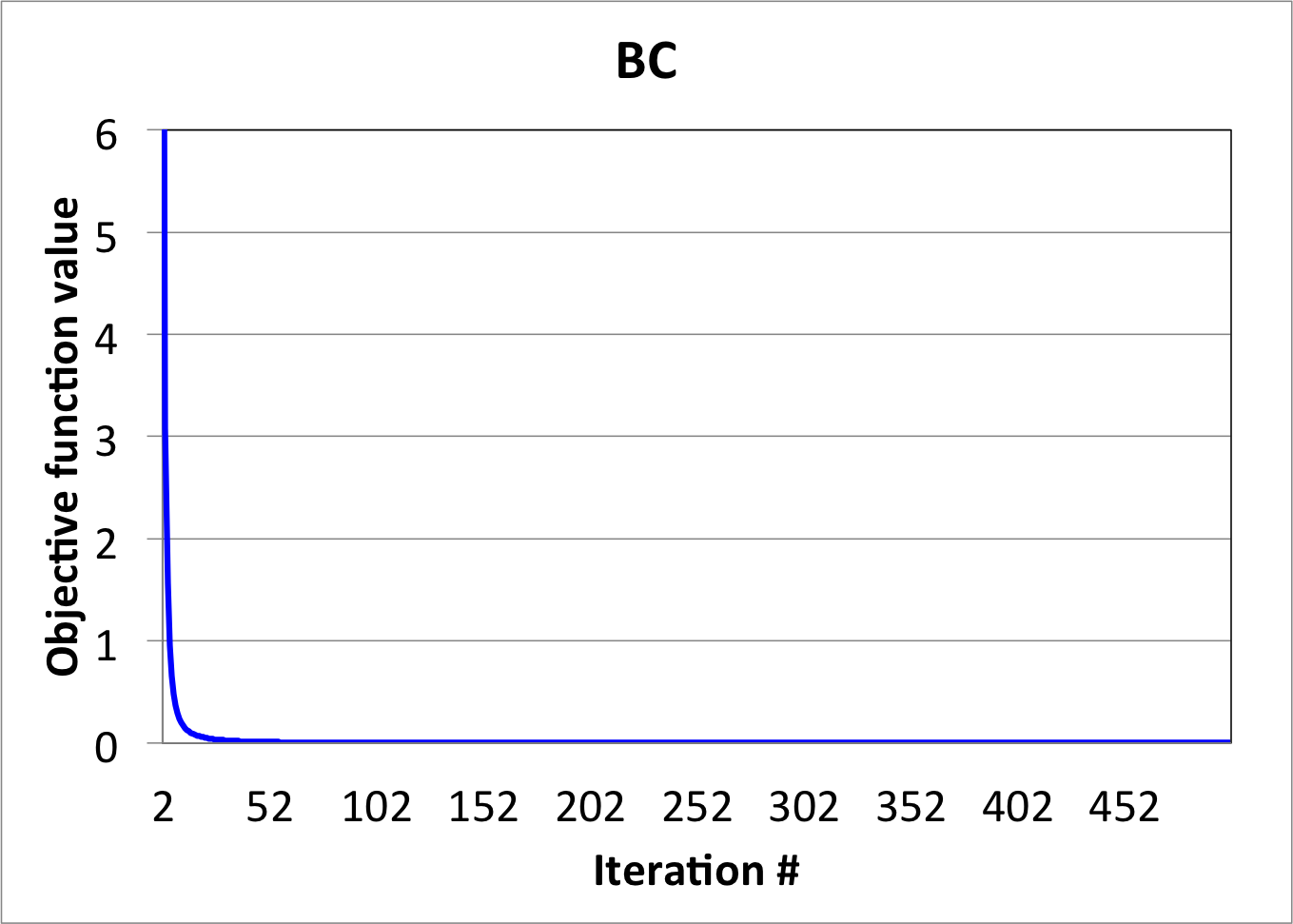}&\hspace{-3ex}
			\includegraphics[scale=0.17]{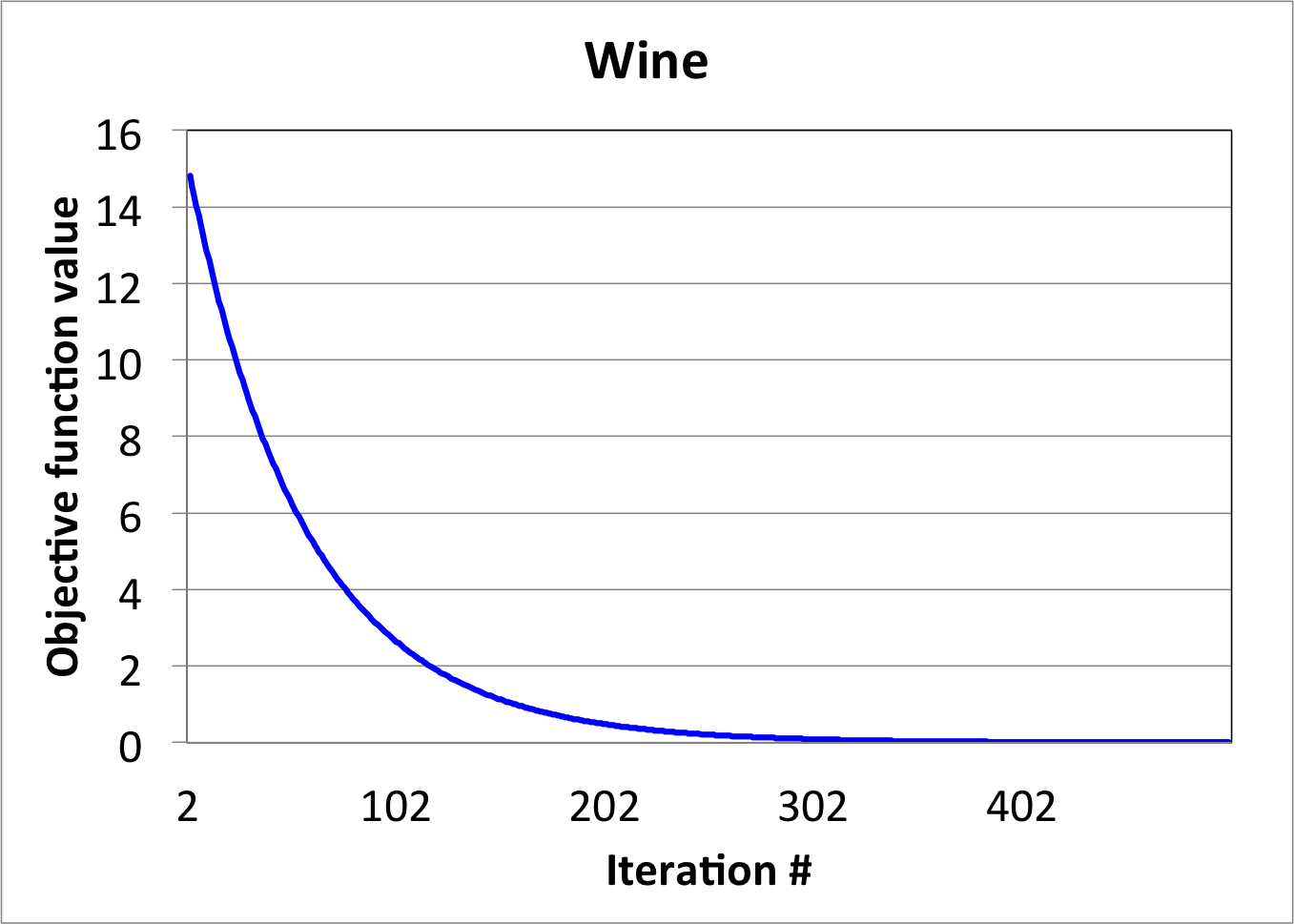}&\hspace{-3ex}
			\includegraphics[scale=0.17]{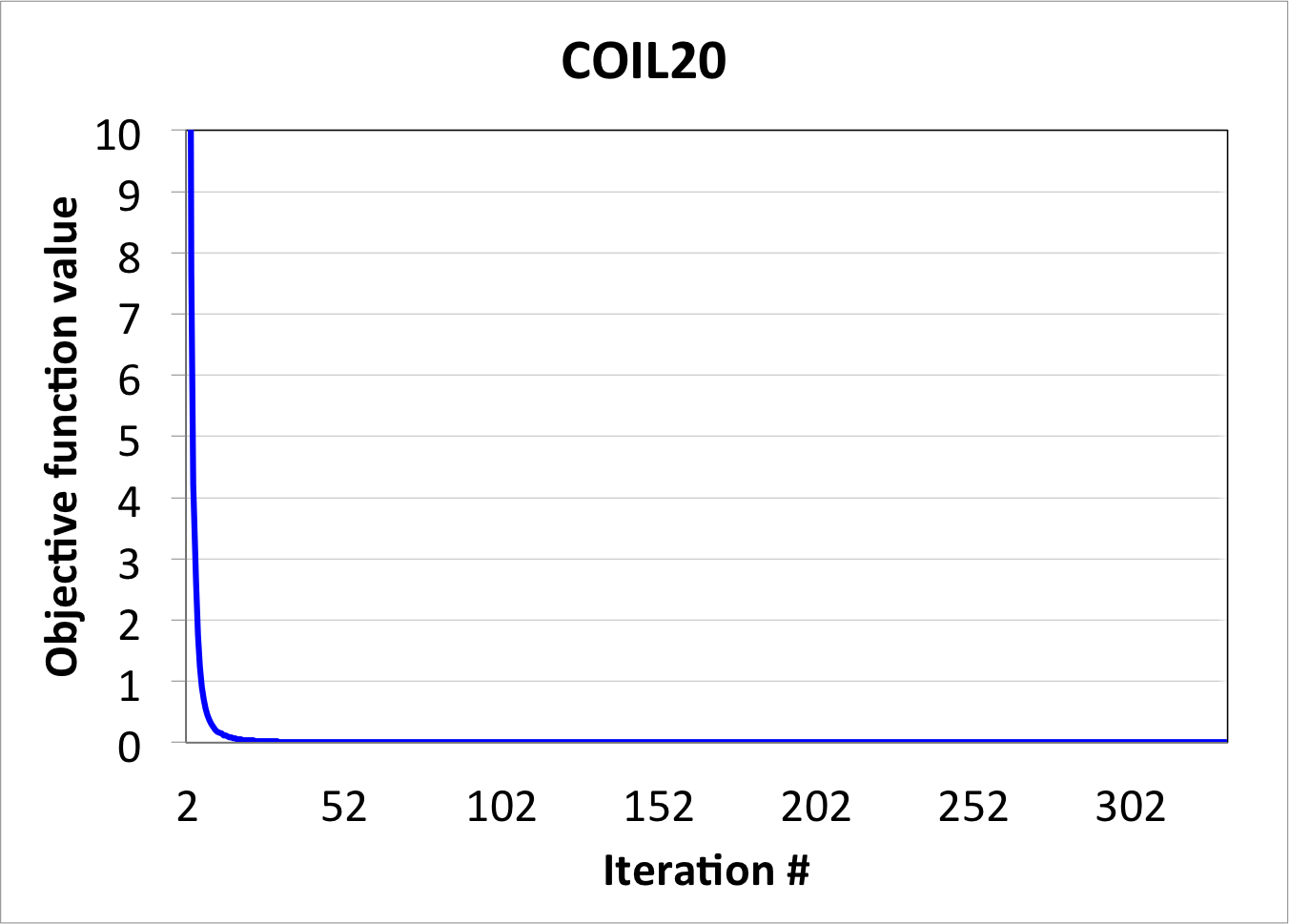}\\
			\includegraphics[scale=0.17]{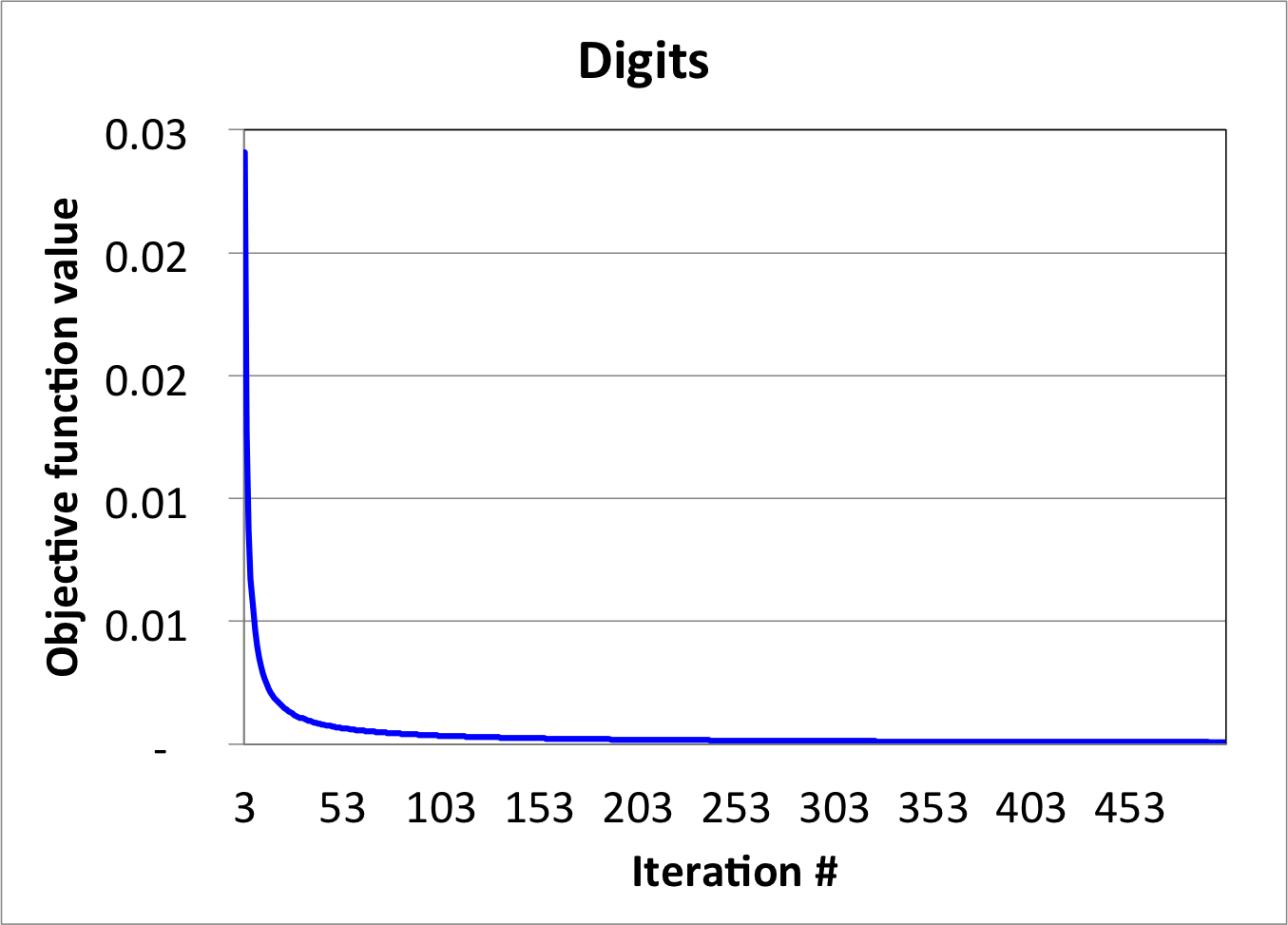}&\hspace{-3ex}
			\includegraphics[scale=0.17]{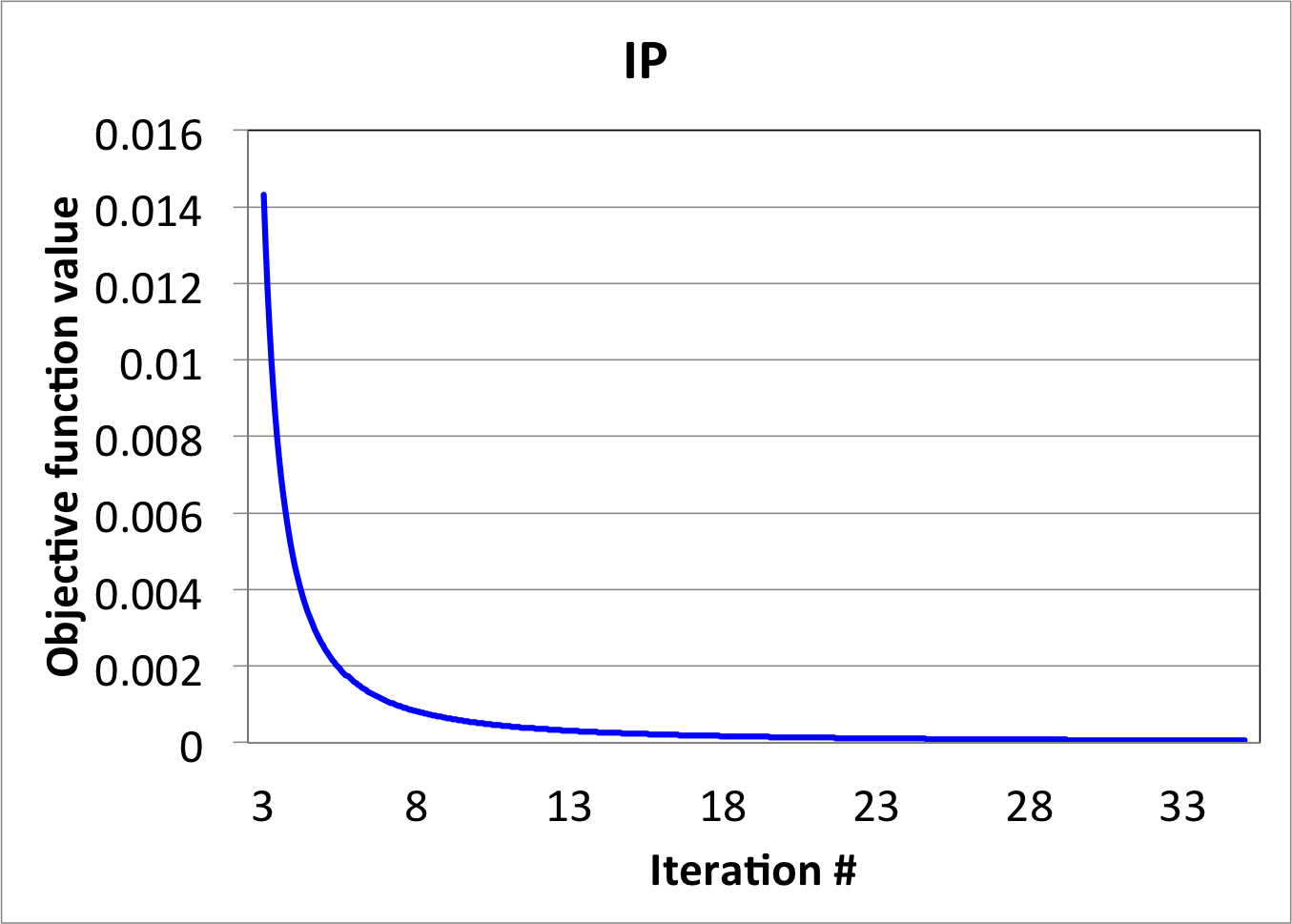}&\hspace{-3ex}
			\includegraphics[scale=0.17]{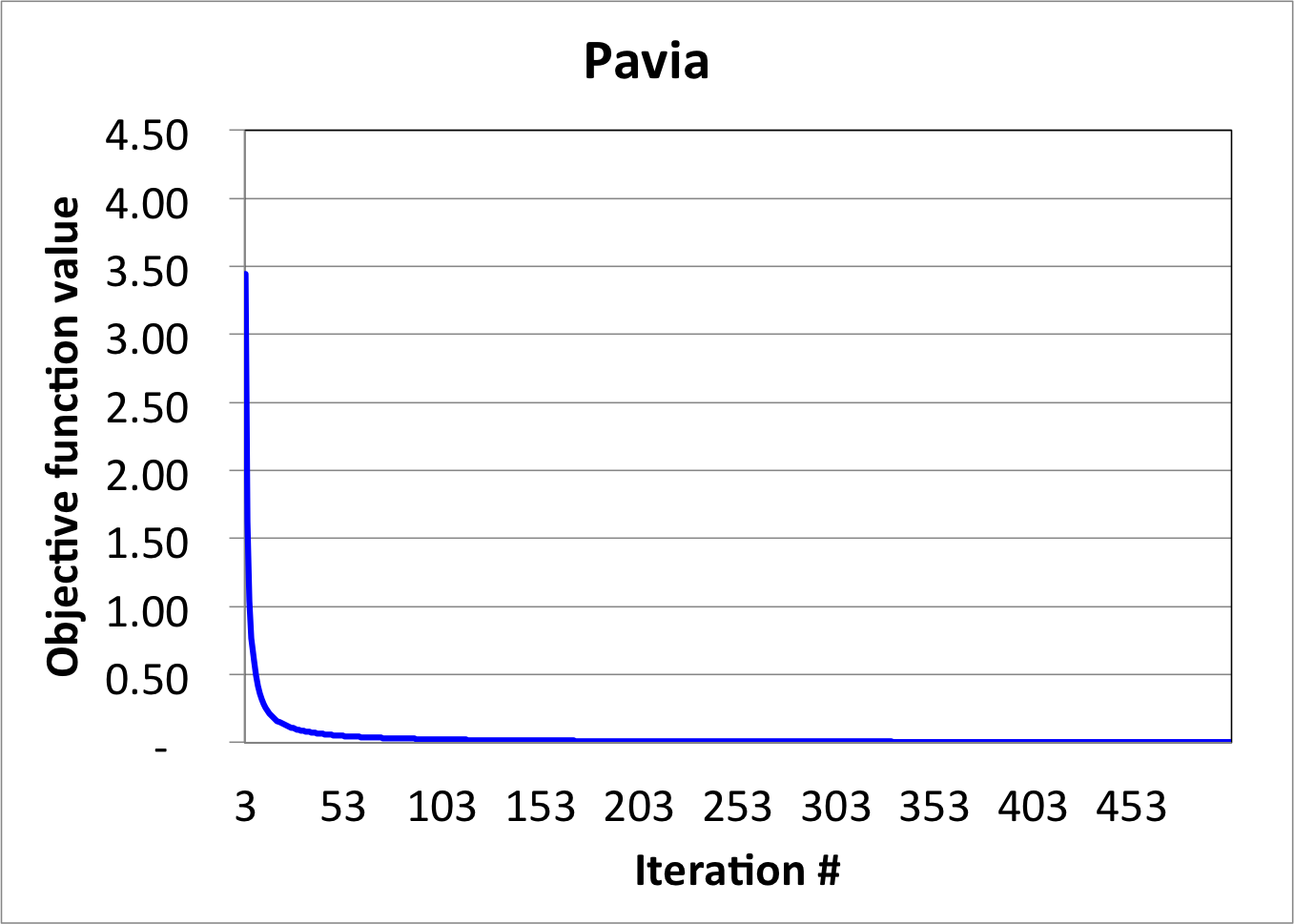}
		\end{tabular}
		\caption{Convergence curve of DPS-NMF.}
		\label{fig-convergence}
	\end{figure}

	In the appendix B below, we have proved the convergence of the iterative updating rules of Eqs. (\ref{eq-ourh}) and (\ref{eq-ourw}). Figure \ref{fig-convergence} illustrates the convergence curves of DPS-NMF on all the datasets considered. In each figure, the y-axis indicates the values of the objective function, and x-axis indicates the number of iterations required for convergence. {The plot in the figure show} that the proposed algorithm converges after at most 100 iterations. For some datasets, it converges in fewer iterations.
	
	\section{Conclusion and discussion}
	\label{seq-conclusion}
	In this paper, we have proposed a new NMF-based algorithm for data representation and dimension reduction. The derived optimization model makes the update of the coefficient matrix parameterizable, and therefore by adjusting this parameter, the quality of lower rank data can be improved. The quality of lower rank data is assessed by its impact on classification and clustering performance. The penalty term added to the NMF cost leads to the diagonalization of the similarity matrix of the bases (i.e., $\textbf{W}^\top \textbf{W}=\lambda \textbf{I}$). The diagnalization of $\textbf{W}^\top \textbf{W}$ favors the decorrelation of the bases. This can have a positive impact on the discrimination of the data points when they are mapped to the reduced space. The {experiments we conducted on various datasets show the superiority of the proposed algorithm in particular compared to those belonging to same family, i.e. NMF-based}. 
	Furthermore, parameterizing the update of the coefficient matrix makes it possible {to convert the algorithm in to a supervised NMF} where the $\lambda $ parameter {is} learned based on a training set and used during the test phase. {This means} that the matrix $ \textbf {W} $ optimized during the training process can be used to reduce feature of test samples. With regard to the cost, the complexity of the multiplicative update algorithm proposed by Lee and Seung \cite{lee2001algorithms}, has been studied by several researchers, i.e., \cite{lin2007projected}, and found that the terms in Eqs. (\ref{eq-h0}) and (\ref{eq-i0}) cost $\mathcal{O}(mnk)$ and the algorithm costs $T \times \mathcal{O}(mnk)$, where $T$ is the number of iterations. Our model includes two additional terms to update $\textbf{H}$, which are $\textbf{HX}^\top \textbf{X}$ and $\textbf{HH}^\top \textbf{H}$ (Eq. (\ref{eq-ourh})) compared to the first update in \cite{lee2001algorithms}. In our work, $\textbf{HX}^\top \textbf{X}$ should be computed by $(\textbf{H}\textbf{X}^\top)\textbf{X}$, which costs $\mathcal{O}(knm)$ and not by $\textbf{H}(\textbf{X}^\top\textbf{X})$, which costs $\mathcal{O}(kn^3)$ since $k,m<<n$. Similarly, $\textbf{H}\textbf{H}^\top\textbf{H}$ should be computed by $(\textbf{H}\textbf{H}^\top)\textbf{H}$, which costs $\mathcal{O}(k^2n)$ and not by $\textbf{H}(\textbf{H}^\top\textbf{H})$, which costs $\mathcal{O}(kn^3)$, since $k<<n$. Therefore, the additional costs is $\mathcal{O}(knm)+\mathcal{O}(k^2n)=\mathcal{O}(knm)$, since $k<<m$. {Consequently},  the complexity of the proposed algorithm can be estimated by is $T \times \mathcal{O}(mnk)$, where $T$ is the number of iterations of the algorithm. Like NMF and other NMF-based algorithms, the proposed method has the advantage of convergence to fixed point; however, the speed of convergence depends on the initialization of \textbf{W} and \textbf{H}. The main disadvantage is that the fixed point can be a local minimum.
	
	In future work, we will try to extend the proposed algorithm to supervised NMF by automatically learning the parameter $ \lambda $.

	%\section*{Acknowledgements}
	%The author would like to thank the SQU Internal Grant (IG/SCI/COMP/19/02) for their financial support.

	\section*{Declarations}
	
	\begin{itemize}
		\item[] \textbf{Funding} This work was supported by SQU Internal Grant (IG/SCI/COMP/22/04).
		\item[] \textbf{Conflict of interest} The authors declare that they have no conflict of interest.
		\item[] \textbf{Ethics approval} Not applicable. 
		\item[] \textbf{Consent to participate} Not applicable. 
		\item[] \textbf{Consent for publication} Not applicable. 
		\item[] \textbf{Availability of data and materials} Not applicable. 
		\item[] \textbf{Code availability} The code will be available on GitHub after publication.
		\item[] \textbf{Authors' contributions} Design and methodology: RH, AA, RA; Experiments: RH; Proofreading and criticism: MC.
	\end{itemize}
	
	\begin{appendices}
		
		\section*{Appendix A (Update rules derivation)}
		\label{seq-appendixA}
		\begin{equation}
		\min_{\textbf{W},\textbf{H}} \quad \|\textbf{X}-\textbf{WH}\|^2_F+\|\textbf{X}^\top\textbf{X}-\lambda \textbf{H}^\top\textbf{H}\|_F^2 \quad \text{s.t}\quad \textbf{W},\textbf{H}\geq 0 
		\label{eq-tag1}
		\end{equation}
		
		By introducing two variable matrices $\Lambda_h$ and $\Lambda_w$ of the same dimension as $\textbf{W}^\top$ and $\textbf{H}^\top$ respectively, the model above can be transformed in the following Lagrangian form:
		
		\begin{align}
		J & = \min_{\textbf{W},\textbf{H}}\underbrace{\|\textbf{X}-\textbf{WH}\|^2_F}_{J_1}+\underbrace{\|\textbf{X}^\top \textbf{X}-\lambda \textbf{H}^\top \textbf{H}\|^2_F}_{J_2} - (\underbrace{Tr(\Lambda_w \textbf{W}^\top)+Tr(\Lambda_h \textbf{H}^\top)}_{J_3} 
		\end{align}
		where $Tr(\Lambda_w \textbf{W}^\top) = \sum_{ij}\textcolor{black} {(\Lambda_w)_{ij}}\textbf{W}_{ij}$ and $T(\Lambda_h \textbf{H}^\top) = \sum_{ij}\textcolor{black} {(\Lambda_h)_{ij}}\textbf{H}_{ij}$.

		\begin{align}
		J_1&=Tr\Big [\big (\textbf{X}-\textbf{WH}\big)^\top \big(\textbf{X}-\textbf{WH}\big)\Big]\notag \\
		&=Tr(\textbf{X}^\top \textbf{X})-2Tr(\textbf{X}^\top \textbf{WH})+Tr(\textbf{H}^\top \textbf{W}^\top \textbf{WH}) \notag
		\end{align}
		
		\begin{align}
		J_2 &=Tr\Big[\big(\textbf{X}^\top \textbf{X}-\lambda \textbf{H}^\top \textbf{H}\big)^\top \big(\textbf{X}^\top \textbf{X}-\lambda \textbf{H}^\top \textbf{H} \big)\Big]\notag \\
		&=Tr(\textbf{X}^\top \textbf{X} \textbf{X}^\top \textbf{X})-2\lambda Tr(\textbf{X}^\top \textbf{XH}^\top \textbf{H})+\lambda^2Tr(\textbf{H}^\top \textbf{HH}^\top \textbf{H}) \notag
		\end{align}
		
		\begin{align}
		J_3 &= Tr(\Lambda_w \textbf{W}^\top)+Tr(\Lambda_h \textbf{H}^\top) \notag
		\end{align}
		Let us solve:
		\begin{align}
		\frac{\partial J}{\partial \textbf{H}}&=\frac{\partial J_1}{\partial \textbf{H}}+\frac{\partial J_2}{\partial \textbf{H}}+\frac{\partial J_3}{\partial \textbf{H}} = 0 \notag
		\end{align}
		and
		\begin{align}
		\frac{\partial J}{\partial \textbf{W}}&=\frac{\partial J_1}{\partial \textbf{W}}+\frac{\partial J_2}{\partial \textbf{W}}+\frac{\partial J_3}{\partial \textbf{W}} =0 \notag
		\end{align}
		with
		\begin{align}
		\frac{\partial J_1}{\partial \textbf{H}} & = -2\textbf{W}^\top X+2\textbf{W}^\top \textbf{WH} \notag \\ 
		\frac{\partial J_2}{\partial \textbf{H}} & = -4\lambda \textbf{HX}^\top \textbf{X} + 4\lambda^2\textbf{HH}^\top \textbf{H} \notag\\ 
		\frac{\partial J_3}{\partial \textbf{H}} & = \Lambda_h \notag
		\end{align}
		and
		\begin{align}
		\frac{\partial J_1}{\partial \textbf{W}} &=-2\textbf{XH}^\top+2\textbf{WHH}^\top  \notag\\ 
		\frac{\partial J_2}{\partial \textbf{W}} &= 0  \notag\\ 
		\frac{\partial J_3}{\partial \textbf{W}} &= \Lambda_w \notag
		\end{align}
		
		Therefore 
		\begin{align}
		\frac{\partial J}{\partial \textbf{H}} &= -2\textbf{W}^\top X+2\textbf{W}^\top \textbf{WH} -4 \lambda \textbf{HX}^\top \textbf{X} + 4\lambda^2\textbf{HH}^\top \textbf{H} - \Lambda_h  \\
		\frac{\partial J}{\partial \textbf{W}} &= -2\textbf{XH}^\top+2\textbf{WHH}^\top - \Lambda_w 
		\label{eq-14}
		\end{align}
		
		By setting  $\frac{\partial J}{\partial \textbf{H}}=0$ and $\frac{\partial J}{\partial \textbf{W}}=0$, and by using the KKT conditions, $(\Lambda_h)_{ij}\textbf{H}_{ij} = 0$ and $(\Lambda_w)_{ij}\textbf{W}_{ij} = 0$, we get the following equations for $\textbf{H}_{ij}$ and $\textbf{W}_{ij}$:
		
		\begin{align}
		-(2\textbf{W}^\top\textbf{X}+ 4\lambda \textbf{HX}^\top \textbf{X})_{ij}\textbf{H}_{ij} + (2\textbf{W}^\top \textbf{WH} +4\lambda^2\textbf{HH}^\top \textbf{H}  )_{ij}\textbf{H}_{ij} = 0, \notag\\
		\therefore (\textbf{W}^\top \textbf{X} + \lambda \textbf{HX}^\top \textbf{X})_{ij}\textbf{H}_{ij} = (\textbf{W}^\top \textbf{WH}  + \lambda^2\textbf{HH}^\top \textbf{H} )_{ij}\textbf{H}_{ij},  \notag
		\end{align}
		hence,
		\begin{align}
		\textbf{H}_{ij} = \textbf{H}_{ij}.\frac{\Big (\textbf{W}^\top \textbf{X}+ 2\lambda \textbf{HX}^\top \textbf{X} \Big)_{ij}}{\Big (\textbf{W}^\top \textbf{WH} + 2\lambda^2\textbf{HH}^\top \textbf{H}   \Big)_{ij}}.
		\end{align}
		And also 
		{\footnotesize
			\begin{align}
			-(2\textbf{XH}^\top)_{ij}\textbf{W}_{ij}+(2\textbf{WHH}^\top )_{ij} &=0, \notag\\
			\therefore (\textbf{XH}^\top)_{ij}\textbf{W}_{ij}+(\textbf{WHH}^\top )_{ij},  \notag
			\end{align}
		}
		hence,
		\begin{align}
		\textbf{W}_{ij}= \textbf{W}_{ij}.\frac{ \Big (\textbf{XH}^\top \Big)_{ij} }{ \Big (\textbf{WHH}^\top \Big )_{ij} }. 
		\end{align}

		\section*{Appendix B (Proof of Theorem 1)}
		\label{seq-appendixB}
		
		The proof of the Theorem 1 requires to demonstrate that $\mathcal{O}$ is non-increasing under the updating steps in Eqs. (\ref{eq-ourh}) and (\ref{eq-ourw}). Since the second term of $\mathcal{O}$ is not related to $\textbf{H}$, we have the same update formula for $\textbf{W}$ in DSP-NMF as in the basic NMF. Thus, we can use the convergence proof of NMF tp show that $\mathcal{O}$ is non-increasing under the update step in Eq. (\ref{eq-ourh}). See \cite{lee2001algorithms} for more details. Hence, the only thing we need is to prove that $\mathcal{O}$ is non-increasing under the update step in Eq. (\ref{eq-ourh}). The auxiliary function \cite{lee2001algorithms} is used to prove the convergence. Let first define the auxiliary function.
		
		\begin{definition}
			$G(h, h^\prime)$ is an auxiliary function for $F(h)$ if the following conditions are satisfied, ie.,:\\
			$G(h, h^\prime) \geq F(h)$, \quad     $G(h,h)=F(h)$
			\label{def-def1}
		\end{definition}
		
		%	\begin{lemma}
		If $G$ is an auxiliary function of $F$, then $F$ is non-increasing under the update
		\begin{equation}
		h^{t+1} = \arg\min_h G(h, h^\prime)
		\label{eq-lema1}
		\end{equation}
		%	\end{lemma}
		
		\textit{Proof:}\\
		\begin{equation}
		F(h^{t+1}) \leq G(h^{t+1}, h^t)\leq G(h^t, h^t) = F(h^t)
		\label{eq-proof}
		\end{equation}
		
		Next, we will show that the update rule for $\textbf{H}$ in Eq. (\ref{eq-ourh}) is exactly the update in Eq. (\ref{eq-lema1}) with a proper auxiliary function.
		
		Let
		\begin{equation}
		\mathcal{O} = \|\textbf{X}-\textbf{WH}\|^2_F+\|\textbf{X}^\top\textbf{X}-\lambda \textbf{H}^\top\textbf{H}\|_F^2,
		\label{eq-eq0}
		\end{equation}
		
		then we have
		\begin{equation}
		F^\prime_{ab} = \Bigg ( \frac{\partial \mathcal{O}}{\partial \textbf{H} } \Bigg)_{ab} = \Bigg (-2\textbf{W}^\top \textbf{X}+2\textbf{W}^\top \textbf{WH} -4 \lambda \textbf{HX}^\top \textbf{X} + 4\lambda^2\textbf{HH}^\top \textbf{H }\Bigg )_{ab},
		\label{eq-fprime}
		\end{equation}
		
		and
		\begin{equation}
		F^{\prime \prime}_{ab} = 2(\textbf{W}^\top\textbf{W})_{aa} 
		- 4\lambda(\textbf{X}^\top\textbf{X})_{bb} + 12\lambda^2(\textbf{H}^\top\textbf{H})_{bb}.
		\label{eq-fprime2}
		\end{equation}
		
		Since the update rule is essentially element-wise, it is sufficient to show that each $F^\prime_{ab}$ is non-increasing under the update of Eq. (\ref{eq-ourh}).
		
		%	\begin{lemma}
		\begin{align}
		G(h, h_{ab}^t) &= F_{ab}(h_{ab}^t) + F^\prime_{ab}(h_{ab}^t)(h-h_{ab}^t) \notag\\		
		&+\frac{  \Big (\textbf{WW}^\top\textbf{H}\Big )_{ab}+2\lambda^2\Big (\textbf{HH}^\top \textbf{H }\Big )_{ab}   }{h_{ab}^t}(h-h_{ab}^t)^2
		\label{eq-lema2}
		\end{align}
		%	\end{lemma}
		is an auxiliary function for $F_{ab}$.
		
		\textit{Proof:} It is obvious that $G(h,h) = F_{ab}(h)$. We need only to show that $G(h,h_{ab}^t) \geq F_{ab}(h)$. First we get the Taylor series expansion of $F_{ab}(h)$ as follows:
		\begin{align}
		F_{ab}(h) &= F_{ab}(h_{ab}^t) + F^\prime_{ab}(h_{ab}^t)(h-h_{ab}^t) \notag \\
		&+\Bigg ( 2(\textbf{W}^\top\textbf{W})_{aa} 
		- 4\lambda(\textbf{X}^\top\textbf{X})_{bb} + 12\lambda^2(\textbf{H}^\top\textbf{H})_{bb} \Bigg )(h-h_{ab}^t)^2,
		\label{eq-f}
		\end{align}  
		
		and compare it with Eq. (\ref{eq-lema2}) to find that $G(h, h_{ab}^t) \geq F_{ab}(h)$ is equivalent to
		\begin{equation}
		\frac{  \Big (\textbf{W}^\top\textbf{W}\textbf{H}\Big )_{ab}+2\lambda^2\Big (\textbf{HH}^\top \textbf{H }\Big )_{ab}   }{h_{ab}^t} \geq 2(\textbf{W}^\top\textbf{W})_{aa} 
		- 4\lambda(\textbf{X}^\top\textbf{X})_{bb} + 12\lambda^2(\textbf{H}^\top\textbf{H})_{bb}.
		\label{eq-compare}
		\end{equation}
		We have 
		
		\begin{equation}
		\Big (\textbf{W}^\top\textbf{W}\textbf{H}\Big )_{ab} = \sum_{l=1}^k\Big (\textbf{W}^\top\textbf{W} \Big)_{al}h_{lb}^t \geq 2\Big (\textbf{W}^\top\textbf{W} \Big)_{aa}h_{ab}^t
		\label{eq-comapre1}
		\end{equation}
		and
		\begin{align}
		\lambda^2\Big (\textbf{HH}^\top \textbf{H }\Big )_{ab} &= \sum_{j=1}^n \lambda^2 h^t_{aj}\Big (\textbf{H}^\top \textbf{H }\Big )_{jb} \notag \\
		& \geq 12\lambda^2 h^t_{ab}\Big (\textbf{H}^\top \textbf{H }\Big )_{bb}\notag \\
		& \geq 12\lambda^2 h^t_{ab}\Big (\textbf{H}^\top \textbf{H }\Big )_{bb} - 4\lambda(\textbf{X}^\top\textbf{X})_{bb} 
		\label{eq-compare2}
		\end{align}
		
		Therefore, Eq. (\ref{eq-compare}) holds and $G(h, h_{ab}^t) \geq F_{ab}(h)$. Now, we can demonstrate the convergence of Theorem 1.
		
		\section*{Proof of Theorem 1:} Substituting $G(h, h^t_{ab})$ in Eq. (\ref{eq-lema1}) by Eq. (\ref{eq-lema2}) results in the following update rule:
		{\footnotesize
			\begin{align}
			h_{ab}^{t+1} &= h_{ab}^t - h_{ab}^t\frac{F_{ab}^\prime(h_{ab}^t)}{2\Big ( \textbf{W}^\top \textbf{WH}\Big)_{ab}+4\lambda^2\Big ( \textbf{HH}^\top\textbf{H}\Big)_{ab} }\notag \\
			&=h_{ab}^t\Bigg [ \frac{2\Big ( \textbf{W}^\top \textbf{WH}\Big)_{ab}+4\lambda^2\Big ( \textbf{HH}^\top\textbf{H}\Big)_{ab} -  \Big (-2\textbf{W}^\top \textbf{X}+2\textbf{W}^\top \textbf{WH} -4 \lambda \textbf{HX}^\top \textbf{X} + 4\lambda^2\textbf{HH}^\top \textbf{H }\Big )_{ab}     }{ 2\Big ( \textbf{W}^\top \textbf{WH}\Big)_{ab}+4\lambda^2\Big ( \textbf{HH}^\top\textbf{H}\Big)_{ab}  } \Bigg ] \notag \\
			&=h_{ab}^t\frac{\Bigg (\textbf{W}^\top\textbf{X}+2\lambda\textbf{HX}^\top\textbf{X}\Bigg )_{ab}}{ \Big ( \textbf{W}^\top \textbf{WH}\Big)_{ab}+2\lambda^2\Big ( \textbf{HH}^\top\textbf{H}\Big)_{ab}}
			\end{align}
		}
		
		Since Eq. (\ref{eq-lema2}) is an auxiliary function, and $F_{ab}$ is non-increasing under this update rule.

		\section*{Appendix C (Proof of Theorem 2)}
		\label{seq-appendixC}
		
		%\begin{proof}
		We will follow the same proof process as in \cite{HEDJAM2021107814}. One side inverse matrices have  the following properties  \cite{rao1972generalized}:
		\begin{enumerate}
			\item[(1)] If the matrix $\textbf{A}$ has dimensions $n\times k$ and rank $\rho(\textbf{A})=k$, then there exists a $k\times n$ matrix $\textbf{A}_L$ called the left inverse of $\textbf{A}$ such that $\textbf{A}_L\textbf{A}=\textbf{I}_k$, where $\textbf{I}_k$ is the $k\times k$ identity matrix.
			\item[(2)] If the matrix $\textbf{A}$ has dimensions $k\times n$ and rank $\rho(\textbf{A})=k$, then there exists an $n\times k$ matrix $\textbf{A}_R$ called the right inverse of $\textbf{A}$ such that $\textbf{A}\textbf{A}_R=\textbf{I}_k$, where $\textbf{I}_k$ is the $k\times k$ identity matrix.
		\end{enumerate}
		Therefore, If $\textbf{H}\in \mathbb{R}^{k\times n}$ has a rank $\rho(\textbf{H})=k$, then its transpose $\textbf{H}^\top$ has a rank $\rho(\textbf{H}^\top)=k$ as well, and by using the properties (1) and (2) above we can conclude that $\textbf{H}^\top$ has a left inverse $\textbf{H}_L$ and $\textbf{H}$ has a right inverse $\textbf{H}_R$. Based on this, the following proof is derived:
		\begin{align}
		\label{eq-relationship0}
		\textbf{X}^\top\textbf{X} \approx\lambda \textbf{H}^\top\textbf{H} &\Leftrightarrow  \textbf{H}^\top\textbf{W}^\top \textbf{W}\textbf{H} \approx \lambda \textbf{H}^\top\textbf{H}, \quad \quad (\text{since } \textbf{X}=\textbf{WH}) \notag\\
		&\Leftrightarrow \textbf{H}^\top\textbf{W}^\top\textbf{WH} \approx  \textbf{H}^\top\lambda\textbf{I}_k\textbf{H}, \notag \\
		&\Leftrightarrow \textbf{H}_L\textbf{H}^\top\textbf{W}^\top\textbf{WH} \approx  \textbf{H}_L\textbf{H}^\top\lambda\textbf{I}_k\textbf{H}, \notag \\
		&\Leftrightarrow \textbf{W}^\top\textbf{WH} \approx  \lambda\textbf{I}_k\textbf{H}, \notag \\
		&\Leftrightarrow \textbf{W}^\top\textbf{WH}\textbf{H}_R \approx  \lambda\textbf{I}_k\textbf{H}\textbf{H}_R, \notag \\
		&\Leftrightarrow \textbf{W}^\top\textbf{W} \approx  \lambda\textbf{I}_k, \notag \\
		\end{align}
		%\end{proof}

		
		
		
	\end{appendices}

\bibliographystyle{unsrtnat}
\bibliography{references}  %%% Uncomment this line and comment out the ``thebibliography'' section below to use the external .bib file (using bibtex) .





% \end{thebibliography}

\end{document}